\documentclass[11pt]{article}
\usepackage{enumerate,amsmath,hyperref,amssymb,latexsym,amsfonts,amsthm,amscd,amsxtra,natbib}
\usepackage{xypic}
\usepackage[mathscr]{eucal}
\usepackage{epsfig}

\theoremstyle{plain}

\newtheorem{theorem}{Theorem}[section]

\theoremstyle{definition}
\newtheorem{definition}{Definition}[section]
\newtheorem{example}{Example}[section]
\newtheorem{remark}{Remark}[section]

\numberwithin{equation}{section}

\newcommand{\A}{\mathcal A}
\newcommand{\B}{\mathcal B}
\newcommand{\Ce}{\mathcal C}
\newcommand{\D}{\mathcal D}

\newcommand{\F}{\mathcal F}
\newcommand{\G}{\mathcal G}

\renewcommand{\cL}{\mathcal L}

\newcommand{\cP}{\mathcal P}
\newcommand{\Q}{\mathcal Q}
\newcommand{\R}{\mathcal R}
\newcommand{\T}{\mathcal T}

\newcommand{\lra}{\longrightarrow}
\newcommand{\lla}{\longleftarrow}

\newcommand{\ovsetl}[1]{\overset {#1}{\lra}}
\newcommand{\ovsetll}[1]{\overset {#1}{\lla}}

\newcommand{\what}{\widehat}

\newcommand{\wti}{\widetilde}

\newcommand{\med}{\medbreak}

\newcommand{\bign}{\bigbreak \noindent}

\newcommand{\sfC}{\mathsf{C}}
\newcommand{\sfD}{\mathsf{D}}

\newcommand{\sfK}{\mathsf{K}}

\newcommand{\sfN}{\mathsf{N}}

\newcommand{\sfS}{\mathsf{S}}

\newcommand{\sfa}{\mathsf{a}}

\newcommand{\sfo}{\mathsf{o}}

\newcommand{\Cl}{\mathsf{Cl}}
\newcommand{\Evt}{\mathsf{Evt}}
\newcommand{\Inf}{\mathsf{Inf}}
\newcommand{\Lg}{\mathsf{Lg}}
\newcommand{\Th}{\mathsf{Th}}

\renewcommand{\a}{\alpha}
\newcommand{\be}{\beta}

\newcommand{\vp}{\varphi}

\begin{document}

\title{A mosaic of Chu spaces and Channel Theory with applications to \\ Object Identification and Mereological Complexity 
%%\\\emph{Version 4.2}
}

\author{Chris Fields\\
23 Rue des Lavandi\`{e}res\\
11160 Caunes Minervois, FRANCE\\
fieldsres@gmail.com\\
\\and\\
\\James  F. Glazebrook
\\Department of Mathematics and Computer
Science \\
 Eastern Illinois University,
600  Lincoln Ave.\\ Charleston, IL 61920--3099, USA \\
jfglazebrook@eiu.edu
\\ Adjunct Faculty
\\ Department of Mathematics \\ University of Illinois at
Urbana--Champaign\\ Urbana, IL 61801, USA
 }

\maketitle

\begin{abstract}
Chu Spaces and Channel Theory are well established areas of investigation in the general context of category theory.  We review a
range of examples and applications of these methods in logic and computer science, including Formal Concept Analysis, distributed systems and ontology development.  We then employ these methods to describe human object perception, beginning with the construction of uncategorized object files and proceeding through categorization, individual object identification and the tracking of object identity through time.  We investigate the relationship between abstraction and mereological categorization, particularly as these affect object identity tracking.  This we accomplish in terms of information flow that is semantically structured in terms of local logics, while at the same time this framework also provides an inferential mechanism towards identification and perception. We show how a mereotopology naturally emerges from the representation of classifications by simplicial complexes, and briefly explore the emergence of geometric relations and interactions between objects.
\end{abstract}

\med
%\textbf{Mathematics Subject Classification (2010)}:

\med

\tableofcontents

%%%%%%%%%%%%%%%%%%%%%%%%%%%%%%%%%%%%%%%%%%%%%%%%%%%%%%%%%%%%%%%%%%%%%%%%%%%%%%%%%%%%%%%%%%%%%%%%%%

\section{Introduction}

Category theory provides a language and a range of conceptual tools towards the general study of complexity, originally in a mathematical framework, and later applied extensively to computer science, artificial intelligence, the life sciences, and the study of ontologies (reviewed by \citet{Baianu2006,EV2007,Goguen1,Goguen2,Healy1,Healy2,Poli1,Poli2,Rosen}). This follows a tradition in conceiving of a range of descriptive methods in analytical philosophy as first advocated by F. \citet{Brentano} and then later by E. \citet{Husserl}, and others (surveyed by e.g. \citet{Simons,Smith2003}).

One particular categorical concept is that of a \emph{Chu space}, which entered computer science as a representable model of linear logic originally formulated by \citet{Barr1,Barr2} and \citet{Seely}. An advantage of using Chu spaces is their flexibility in adapting to a wide range of interpretations and applications.  They are more general than topological spaces, and they can be represented in straightforward \emph{object-attribute} rectangular/matrix-like arrays (the rows consisting of object names, and the columns consisting of attribute names; so an $[ij]$-entry simply means that an object $o_i$ has an attribute $a_j$). From the observational perspective, the attributes are taken to provide information about the structural and dynamical configurations of, and between objects.  Following earlier developments of the theory, Chu spaces emerged with importance in areas dealing with machine learning and data mining; these include (but are not limited to) parallel programming algorithms, information retrieval, concurrent computation automata, physical systems, local logics, formal concept analysis (see below), the semantics of observation-measurement problems, decision theory, and ontological engineering \citep{Abramsky1,Allwein,Barr1,Barwise1,BHL,Pratt1,Pratt2,Pratt3,ZS}. Accordingly, one may find a variety of interpretations of Chu-space representations, including the object/attribute criteria used to define informational relationships, depending on the chosen context.

During information processing, the various channels of information assimilation may possess intrinsic qualities that influence the type of inferences they derive from the basic premise that ``X being A carries the information that Y is B'' \citep{Dretske1}.
As a step towards conceptualizing information flow within a logical environment in category-theoretic terms, the basic elements of Chu spaces have been adapted to the concept of \emph{Classifications}, as the latter are expressed in terms of \emph{Tokens} and \emph{Types} \citep{Barwise1,Barwise2,Barwise3,Barwise4}. The resulting framework of \emph{Channel Theory} casts information flow within a logical and distributed systems environment. An \emph{infomorphism}, as a reformulated Chu morphism (in a sense `dual') constitutes a pivotal concept of Channel Theory, by  defining a channel through which the information represented by one classification is re-represented in another.

We have two broad aims in this paper: 1) to assemble the main concepts and tools of Chu spaces and Channel Theory in one place, and briefly review some of their applications, and 2) to apply these concepts and tools to develop category-theoretic descriptions of three interdependent cognitive processes, the construction of object files \citep{Kahneman} and object tokens \citep{Zimmer}, the binding of type and token information in object categorization \citep{Martin07,Keifer12}, and the recognition and categorization of mereologically-complex individuals.  While the notion of ``entry-level'' categories and the extension of such categories both upward and downward in an abstraction (or type) hierarchy has been intensively investigated both experimentally \citep{Clarke15} and theoretically (\citet{sowa06}; see also \S\ref{cognitive} below), the representation of mereological complexity has received far less attention.  Mereological categorization can be functionally dissociated from abstraction-based categorization in humans, e.g. in high-functioning autism where ``weak central coherence'', and hence deficit understanding of mereological complexity may be displayed alongside normal or even superior abstraction ability \citep{Happe06, Booth:16}.  How abstraction-based types and mereological types are related, and how their implementations in humans are related, thus remain to be worked out.  We are also interested in how tokens representing individual, re-identifiable objects, i.e. object tokens as discussed in \S\ref{object-token}, are able to participate both as such and as instances of classified types in both hierarchies simultaneously.  How, for example, can an object token representing a particular dog be both an instance of the entry-level category [dog], as well as more abstract categories such as [mammal] or [animal], while at the same time being represented as both an individual entity with mereological complexity at multiple scales and as a proper component of even more complex entities?  As both abstraction and mereology contribute to the construction of prior probabilities and to the regulation of precision or attention within Bayesian classifiers \citep{Friston2}, the question of how these representations interact -- from an implementation perspective, how they cross-modulate each other -- is crucial to understanding both how mereologically complex objects are identified as individual entities and how identifiable individual entities are recognized as being mereologically complex.

The first part of the paper addresses the initial aim of the tool assembly.  We begin by defining and reviewing some of the basic properties of Chu spaces in \S\ref{chu}.   Although Chu spaces have been traditionally applied to fields such as those listed above, they also have a number of other significant applications of interest here.  How Chu spaces can be implemented within Formal Concept Analysis and Domain Theory (e.g. to represent information systems and approximable concepts following \citet{HZ,KHZ,Scott1982,ZS}) is reviewed in \S\ref{FCA}. In \S\ref{topology} we discuss representations of spaces (and representations by spaces), spatial coarse-graining and finite sampling of information \citep{GP1,Sorkin1}; we then review the representation of sampled information by simplicial complexes constructed ``above'' the sampled space in \S\ref{simplicial}.  The following two sections, \S\ref{channel-I} and \S\ref{channel-II} establish a similar working account of Channel Theory.  We survey a number of motivating examples and applications, including Distributed Systems \citep{Barwise1} in \S\ref{distributed}, Cognizance Classification \citep{SS1,SS2} in \S\ref{cognizance}, and the flow of information in Ontology Comparison and Alignment \citep{Kalfoglou1,Schorlemmer,Schorlemmer2005} in \S\ref{ontologies}.  The category-theoretic concepts of \emph{cocone} and \emph{colimit} (e.g. \citet{Awodey}) naturally arise in both Chu space and Channel Theory descriptions; we review these concepts in \S\ref{colimits} with illustrative examples.

The second part of the paper presents new results.  We begin in \S\ref{cognitive} with brief reviews of perception, categorization and attention as neurocognitive processes and of multi-layer recurrent network models (e.g. \citet{Friston2,Grossberg13}) of these processes.  In \S\ref{tt-flow}, we re-describe perception and categorization, using the Chu space and Channel Theory tools assembled in the first part, in a way that makes explicit the dualities between dynamic and static properties, individuals and categories, and states and events.  We capture these dualities in a ``cone-cocone diagram'' that formalizes the inferential steps required to link object tokens together to produce a ``history'' of a persistent object.  We then turn our attention, in \S\ref{mereological}, to mereological categorization and to the key mereotopological question of how the \emph{boundaries} between the components of a complex object are defined.  As scenes are themselves mereological complexes, the boundary construction required for object segmentation emerges as the simplest case of inter-object boundary definition.

Category theory is in essence a theory of dualities.  Applying the tools of Chu spaces and Channel Theory to cognition, and in particular to object identification, emphasizes the role of concepts and processes representable as category-theoretic duals in cognitive processing.  The roles of complementary information flows at all scales, from on-center/off-surround networks to the dorsal and ventral attention systems to the interplay of memory and prediction that constructs object histories, exemplify such duality.  By taking object identities and object persistence for granted, AI systems have largely neglected the problem of object re-identification that lies at the heart of the frame problem \citep{Fields13, Fields16}.  Taking this problem and the dual organization required to solve it into account suggests reconceptualizations of learning and memory as overarching dual processes.

%%%%%%%%%%%%%%%%%%%%%%%%%%%%%%%%%%%%%%%%%%%%%%%%%%%%%%%%%%%%%%%%%%%%%

\section*{Part I: Category-theoretic Concepts and Tools}

\section{Chu spaces and Chu transforms}\label{chu}

\subsection{Basic definitions for objects and attributes}\label{chu-defs}

\begin{definition}\label{chu-def-1}
\emph{A (dyadic or two-valued) Chu space} $\sfC$ consists of a triple $(C_{\sfo}, \Vdash_{\sfC}, C_{\sfa})$ where $C_{\sfo}$ is a set of \emph{objects}, $C_{\sfa}$ is a set of \emph{attributes}, along with a \emph{satisfaction relation} (or \emph{evaluation})
$\Vdash_{\sfC} \subseteq C_{\sfo} \times C_{\sfa}$.
\end{definition}

For observational purposes, we may regard the ``attributes'' as providing information about the structural and dynamical configurations of and between the ``objects.''  Two objects can be distinguished if but only if there is at least one attribute that they do not share. Otherwise, objects are said to be equivalent.  This sense of equivalence formalizes Leibniz' principle of ''identity of indiscernibles.''  The ``objects'' and ``attributes'' can equally well be thought of as ``states'' and ``events,'' with ``states'' distinguished by the ``events'' that can occur in them or, as we will see, in terms of ``tokens'' and ``types'' or other similar pairs of concepts.

\begin{definition}\label{chu-def-2}
A \emph{morphism} or \emph{Chu transform} of a Chu space $\sfC= (C_{\sfo}, \Vdash_{\sfC}, C_{\sfa})$ to a
Chu space $\sfD= (D_{\sfo}, \Vdash_{\sfD}, D_{\sfa})$ is a pair of functions $(f_{\sfa}, f_{\sfo})$ with
$f_{\sfo}: C_{\sfo} \lra D_{\sfo}$, and $f_{\sfa}: D_{\sfa} \lra
C_{\sfa}$, such that for any $x \in C_{\sfo}$, and $y \in D_{\sfa}$, we have
$f_{\sfo}(x) \Vdash_{\sfD} y$, if and only if $x \Vdash_{\sfC} f_{\sfa}(y)$.
\end{definition}

If $\sfC = (C_{\sfo}, \Vdash_{\sfC}, C_{\sfa})$ is a Chu space, then $\sfC^{\perp} = (C_{\sfa}, \Vdash_{\sfC}^{\rm{op}}, C_{\sfo})$ is the \emph{dual space} of $\sfC$ in which the roles of objects and attributes are interchanged.  This sense of duality allows us to think, for example, of attributes being distinguished by the objects to which they apply, events being distinguished by the states in which they participate, or types being distinguished by the tokens they include.  Chu-space duality will provide, in \S\ref{tt-flow}, the key to representing recurrent networks in a fully-symmetric way.

\begin{remark}\label{multivalued-1}
More generally, for some set $\sfK$, we could also speak of \emph{a $\sfK$-valued Chu space} $\sfC= (C_{\sfo}, \Vdash_{\sfC}, C_{\sfa})$ with a satisfaction relation (evaluation) $\Vdash_{\sfC}: C_{\sfo} \times C_{\sfa} \lra \sfK$, with relation $\Vdash_{\sfC}(a,b)$ an element of $\sfK$.
\end{remark}

\subsection{Chu flows}\label{chu-flow}

What is the information preserved when switching between Chu spaces that are tied by a Chu transform?  Let a \emph{Chu flow} \citep{VBent}, cf. \citet{Barwise1} be specified by a ``flow formula'' constructed from the elements of the following schema:

\begin{equation}
x \Vdash a ~ \vert~ \neg (x \Vdash a) ~ \vert ~ \wedge ~\vert \vee \vert ~ \exists x ~ \vert ~\forall a.
\end{equation}

Any such formula $\psi(a_1, \ldots, a_k, x_1, \ldots, x_m)$ specifies which objects $x_i$ have which attributes $a_i$ in the Chu space in which it applies.  \citet{VBent} has shown that for finite Chu spaces $\sfC$ and $\sfD$, the existence
of a Chu transform $\sfC \lra \sfD$ is equivalent to every flow formula valid in $\sfC$ being valid in $\sfD$ as well.  The transform $\sfC \lra \sfD$ can, in this case, be viewed as ``transporting'' the information encoded in valid flow formulas from $\sfC$ to $\sfD$; it can thus be thought of informally as a ``channel'' from $\sfC$ to $\sfD$ and as implicitly providing a sense of ``spatial'' and/or ``temporal'' separation between $\sfC$ and $\sfD$.  These informal notions will be made more precise in \S\ref{channel-I}.

\begin{example}
A given flow formula $\psi(a_1, \ldots, a_k, x_1, \ldots, x_m)$ can give rise to useful relations between $k$ objects and $m$ types.
For instance \citep{VBent}:
$$
\begin{aligned}
\forall x (\neg a_1 \in x \wedge a_2 \in x )~~ &\subset~~\text{object~inclusion}, \\
\forall x (\neg a_1 \in x \vee a_2 \in x )~~ &\boxminus ~~\text{object~incompatibility}, \\
\exists a (a \in x_1 \wedge a \in x_2 ) ~~&~\sfo ~~~\text{type~overlap}. \\
\end{aligned}
$$
\end{example}

\subsection{Biextensional collapse}\label{bi-ext}

Following \citet{Pratt2} we define a pair of maps relative to power sets $\cP(\cdot)$ as follows:
\begin{equation}\label{pratt-1}
\begin{aligned}
\hat{\a} &: C_{\sfo} \lra \cP(C_{\sfa}) ~\text{with}~ \hat{\a}(x) = \{a \in C_{\sfa} : x \Vdash_{\sfC} \sfa \} \\
\hat{\omega} &: C_{\sfa} \lra \cP(C_{\sfo}) ~\text{with}~ \hat{\omega}(\sfa) = \{x \in C_{\sfo}: x \Vdash_{\sfC} \sfa \}.
\end{aligned}
\end{equation}
Given $X \subseteq C_{\sfo}$, and $A \subseteq C_{\sfa}$, the above two maps extend to the following maps, respectively \cite{ZS}:
\begin{equation}\label{pratt-2}
\begin{aligned}
\a &: \cP(C_{\sfo}) \lra \cP(C_{\sfa}) ~\text{with}~ \a(x) = \{\sfa: \forall x \in X ~x \Vdash_{\sfC} \sfa \} \\
\omega &: \cP(C_{\sfa}) \lra \cP(C_{\sfo}) ~\text{with}~ \omega(A) = \{x: \forall \sfa \in A ~ x \Vdash_{\sfC} \sfa \}.
\end{aligned}
\end{equation}

A Chu space $\sfC$ is said to be \emph{extensional} if $\hat{\omega}$ is injective, and \emph{separable} if  $\hat{\a}$ is injective. If $\sfC$ is both extensional and separable, then let us say it is \emph{biextensional}. In fact, any Chu space can be turned into a biextensional type, provided the lack of injectivity of $\a$ and $\omega$ can be factored out in a suitable sense.  This creates a \emph{biextensional collapse} of a Chu space
$\sfC= (C_{\sfo}, \Vdash_{\sfC}, C_{\sfa})$, namely the Chu space
\begin{equation}
\what{\sfC} = (\what{C}_{\sfo}, \Vdash_{\what{\sfC}}, \what{C}_{\sfa})
= (\hat{\a}(C_{\sfo}), \Vdash_{\what{\sfC}}, \hat{\omega}(C_{\sfa})),
\end{equation}
where $\hat{\a}(x) \Vdash_{\what{\sfC}} \hat{\omega}(a)$, if and only if $x \Vdash_{\sfC} \sfa$. In essence this means that in the biextensional collapse any repetitions in the rows of objects (tokens) and columns of attributes (types) are factored out, thus removing unnecessary repetitions in the content of information and hence minimizing the amount of processing units in a given algorithm.

%%%%%%%%%%%%%%%%%%%%%%%%%%%%%%%%%%%%%%%%%%%%%%%%%%%%%%%%%%%%%%%%%%%%%%%%%%%%%%%%%%%%%%%%%

\section{Formal Concept Analysis and Computation in Chu spaces}\label{FCA}

Category theory can be viewed as a unified language for handling conceptual complexities in both mathematics and computer science.  Chu spaces and Chu flows provide a natural way of representing both the structure and processing of information and have been used to investigate the semantic foundations and design of data structures and programming languages.  The examples that follow illustrate these applications and introduce concepts that will prove useful later.

\subsection{Concept lattices and approximable concepts}\label{FCA-1}

\emph{Formal Concept Analysis} (FCA) is an approach to the semantics of symbolic data structures that studies the clustering of attributes into partially ordered sets that give rise to a \emph{concept lattice} \citep{Ganter}.  \emph{Domain Theory} (DT) for programming languages is concerned with higher-order relations between concepts that involve partial information and successive approximation, and with the question of when information can be approximated by finitely representable \emph{approximable concepts} \citep{ZS} (cf. \emph{formal contexts} described in \citet{HZ}).  A central idea of FCA is the distinction between the `extension' of a concept as consisting of all objects belonging to that concept, and the `intension' of the concept as consisting of all attributes common to all objects belonging to that concept.  Defining a concept in FCA thus involves identifying a collection of attributes which agrees with the `intension of the extension'.  Note that the idea of `intension' in FCA captures the philosophical notion of an ``essential property'' that all members (here, objects) of a category (here, a concept) must have.

This FCA notion of `concept' has been shown to be intrinsic to a Chu space \citep{KHZ,ZS}; indeed each Chu space $\sfC = (C_{\sfo}, \Vdash_{\sfC}, C_{\sfa})$ has an associated complete lattice $\mathcal{L} \sfC$ of formal concepts associated with $\sfC$.  \citet[Th. 4.1]{ZS} have further shown that for every complete lattice $D$ of formal (in the sense of FCA) concepts, there is a Chu space $\sfC$ such that $D$ is order-isomorphic to $\mathcal{L} \sfC$.  The following definition(s) then characterize the differences between `formal' (in the sense of FCA) and `approximable' (in the sense of DT) concepts.

Let $P,Q$ be sets, and $\mathcal{A} \subseteq \mathcal{P}(P)$, $\mathcal{B} \subseteq \mathcal{P}(Q)$ (recall that $\mathcal{P}(\cdot)$ denotes the power set).  Any pair of functions $s: \mathcal{A} \lra \mathcal{B}$, $t: \mathcal{B} \lra \mathcal{A}$, is called a \emph{Galois connection}, if for each $X \in \mathcal{A}$ and $Y \in \mathcal{B}$, $s(X) \supseteq Y$ if and only if $X \subseteq t(Y)$.  With respect to a Chu space $\sfC = (C_{\sfo}, \Vdash_{\sfC}, C_{\sfa})$, we recall from \eqref{pratt-2} the two associated functions (depending on $\sfC$, so $\alpha = \alpha_{\sfC}$ and $\omega = \omega_{\sfC}$ ):
\begin{equation}
\begin{aligned}
\a &: \cP(C_{\sfo}) \lra \cP(C_{\sfa}) ~\text{with}~ \a(x) = \{\sfa: \forall x \in X ~x \Vdash_{\sfC} \sfa \} \\
\omega &: \cP(C_{\sfa}) \lra \cP(C_{\sfo}) ~\text{with}~ \omega(A) = \{x: \forall \sfa \in A ~ x \Vdash_{\sfC} \sfa \}.
\end{aligned}
\end{equation}
The pair of maps $(\alpha, \omega)$ forms such a Galois connection \citep{Ganter}: i) the set of attribute (object) concepts of $P$ forms a closure system, i.e. a family of subsets closed under intersection \citep{Caspard03}; the attribute (object) concepts of $\sfC$ under set inclusion form a complete lattice; and, iii) the lattice of attribute concepts, and the lattice of object concepts are anti-isomorphic to each other.  We then have:
\begin{definition}
$~$
\begin{itemize}
\item[(1)] A subset $A \subseteq C_{\sfa}$ is called an \emph{(formal) concept (of attributes)}, if it is a fixed point of $\alpha \circ \omega$, i.e. $\alpha(\omega(A)) = A$. Dually, a subset $X \subseteq C_{\sfo}$ is called a \emph{(formal) concept (of objects)} if it is a fixed point of $\omega \circ \alpha$. For each object $x \in C_{\sfo}$, the set of its attributes $\alpha\{x\}$ is a concept.
\item[(2)]
A subset $A \subseteq C_{\sfa}$ is called an \emph{approximable concept}, if for every finite subset $X \subseteq A$, we have $\alpha (\omega(X)) \subseteq  A$.
\end{itemize}
\end{definition}
\noindent
Note that (1) above allows ``single-object concepts''; these will become important in \S\ref{tt-flow} as representations of object tokens \citep{Fields3}, and for a hierarchial iteration of the idea in \S\ref{mereological}.

\begin{definition}
A \emph{complete algebraic lattice} (henceforth, an \emph{algebraic lattice}) is a partial order which is both a complete lattice and a directed complete partial order (dcpo).
\end{definition}
\noindent
We have now the following basic representation theorem for approximable concepts \citep[Th. 6.3]{ZS}:
\begin{theorem}
For any Chu space $\sfC = (C_{\sfo}, \Vdash_{\sfC}, C_{\sfa})$,  the set of its approximable concepts $\mathcal{A}\sfC$ under inclusion forms an algebraic lattice.  Conversely, for every algebraic lattice $D$, there is a Chu space $\sfC = (C_{\sfo}, \Vdash_{\sfC}, C_{\sfa})$ such that $D$ is order-isomorphic to $\mathcal{A}\sfC$.
\end{theorem}

\subsection{Chu spaces as information systems}\label{chu-info-1}

An \emph{information system} with ``states'' consisting of finite subsets of tokens selected from some set $A$ can be defined in terms of an underlying Chu space as follows.  Let $\mathsf{Fin}(A)$ be the set of finite subsets of $A$ and choose a subset $\mathsf{Con} \subset \mathsf{Fin}(A)$ and a relation $\vdash$ (see \citet{Scott1982} for details).  Interpret the information states $x$ (i.e. elements of $\mathsf{Con}$) as objects, the tokens $a \in A$ as attributes, and let $x \Vdash a$ if and only if $a$ is a member of $x$.  In this case, the subset $\mathsf{Con}$ on $A$ is called the \emph{consistency predicate}, and $\vdash$ the \emph{entailment relation}. Following \citet{ZS}, a Chu space $\sfC = (C_{\sfo}, \Vdash_{\sfC}, C_{\sfa})$ gives rise to an information system $(A_{\sfC}, \mathsf{Con}_{\sfC}, \vdash_{\sfC})$ via the assignment $A_{\sfC} = C_{\sfa}$,~$X \vdash_{\sfC} \sfa$, if $\sfa \in \a_{\sfC} \circ \omega_{\sfC}(X)$, and a consistency predicate $\mathsf{Con}_{\sfC}$
for which every subset of $C_{\sfa}$ is consistent. \citet[Th. 4.6]{ZS} have shown, for a given Chu space $\sfC = (C_{\sfo}, \Vdash_{\sfC}, C_{\sfa})$
with $C_{\sfa}$ finite, a state $X \subset C_{\sfa}$, taken to be a \emph{concept}, is equivalent to $X$ being a state of the derived information system $(A_{\sfC}, \mathsf{Con}_{\sfC}, \vdash_{\sfC})$. Intuitively, a Chu morphism in Definition \ref{chu-def-2} correlating the objects and attributes of $\sfC$ to those of some other Chu space $\sfD$ is a correlation between the respective information systems.  Such a morphism similarly maps sequences of flow formulas valid in $\sfC$ to sequences of flow formulas valid in $\sfD$, and hence correlates information \emph{processes} in the respective information systems.

\subsection{Ordered dynamical systems and computation}\label{ordered}

The stage is now set to develop a general notion of computation for arbitrary dynamical systems with discrete states.  As a sequence of \emph{measurements} of any arbitrary dynamical system can itself be considered a dynamical system with discrete states \citep{Fields89}, nothing is lost by assuming discreteness.  Artificial neural networks (ANNs) are such systems \citep{Nauck}, as are Turing machines, cellular automata, etc.

Consider a quadruple $\langle S, \mathsf{ns}, \leq, \T\rangle$, where $S$ is the state space of an information system as characterized above, $\mathsf{ns}$ is the next-state function, $\leq$ is a partial order and $\T: \cL \lra S$ is a mapping where $\cL$ denotes a propositional (`factual') language \citep{Leitgeb1}.  The map $\T$ assigns some proposition $\phi$ of $\cL$ to each state
$s \in S$; hence it represents the (stipulated) semantics of $S$.  The action of $\mathsf{ns}$, in this case, produces a sequence of propositions $\phi_0, \phi_1, \phi_2, ...$ and so can be interpreted as (in general, nonmonotonic) inference.  If this sequence converges to some stable state $\psi$, the action of $\mathsf{ns}$ has ``halted'' and the proposition $\psi$ can be interpreted as the ``result'' of the action of $\mathsf{ns}$ on $\phi_0$.  The design perspective in which $\mathsf{ns}$ is stipulated and the reverse engineering or debugging perspective in which $\mathsf{ns}$ must be discovered are both clearly supported within this picture.

Recasting the above in the language of $\sfK$-valued Chu spaces (\ref{multivalued-1}) provides a representation of computations with imprecise inputs, outputs or both.  Recalling the attribute symbol $\Vdash$, let us define
$$
\begin{cases} s \Vdash^t \phi ~ &\text{iff} ~ \T(\phi) = s ~\text{(precise state information)}, \\
s \Vdash \phi ~ &\text{iff} ~ \T(\phi) \leq s ~\text{(imprecise state information)}. \\
\end{cases}
$$
The computational interpretation is straightforward: $s \Vdash^t \phi$ if and only if $\phi$ completely specifies the system state, whereas $s \Vdash \phi$ if and only if the system state is described by $\phi$ as well as some other propositions in $\cL$.  A computation with an initial state $s \Vdash^t \phi$ and a final state $s^\prime \Vdash \psi$, for example, would provide an ambiguous answer ($\psi$ together with other propositions) to a precise question ($\phi$).

This Chu/information space representation of computation has been adapted to capture Bayesian inference in a connectionist context \citep{Dayan,McClelland1}; we develop this representation further in \S\ref{tt-flow}.   The close relationship between Chu flows and infomorphisms as defined within Channel Theory \citep{Barwise1} and their application to problems such as ontology alignment are considered in \S\ref{channel-I} and \S\ref{channel-II}, respectively. In particular, state space systems will be further described in the context of Channel Theory in \S\ref{states-1} and \S\ref{states-2}

%%%%%%%%%%%%%%%%%%%%%%%%%%%%%%%%%%%%%%%%%%%%%%%%%%%%%%%%%%%%%%%%%%%%%%%%%%%%%%%%%%%%%%%%%%%%%%%%

\section{Topology of information and observation}\label{topology}

Propositions used to describe the world are semantically related; in the limit, all propositions in any language form a connected semantic network \citep{sowa06}.  Observations or, more precisely, finitely-specifiable observational outcomes are similarly related.  Considering an information system to be defined merely over a \emph{set} of propositions provides no means of capturing such relations.  It is, therefore, useful to introduce additional structure, with the addition of topological structure a natural first step.  Doing this allows a structured notion of sampling the information encoded in a Chu space, and particularly the idea of a \emph{finite sample of attributes} (FSA) for an object or collection of objects.

\subsection{The Sorkin perspective}\label{Sorkin}

One approach to developing an information topology is via a notion of \emph{causality}; this has been pursued by \citet{Sorkin1,Sorkin2}
through the development of \emph{causal sets}.  While the motivation in this case has been to model the fundamental structure of spacetime in a way that could produce the continuum of macroscopic geometry as an emergent `classical limit' (see \cite{Raptis1,Sorkin1,Sorkin2} for details of the mathematical physics application domain), the techniques employed are generally applicable to approximating a class of highly structured or idealized spaces by means of taking a certain limit of less complex, more user-friendly spaces. The point is to represent non-spatial information in a spatial form as means of `visualization', and to consider variation in data and observations in the context of such representations.  The use of spatial dimensions as a way of ``displaying'' information in a meaningful way on both the input and output sides of connectionist systems (and more generally, ANNs) is an example of this approach.  In this case, the very complex, essentially causal relations between information computed by a learning algorithm are approximated, on an imposed spatial array of output ``units'' that have no intrinsic spatial relationships, in a way that makes them meaningful to external observers \citep[Ch. 8]{Rogers}.  An input array similarly approximates causal relations in ``the world'' when the array geometry is assigned semantic significance, e.g. in computer vision applications.

Fundamental causality relations between objects $x,y,z$ can be expressed in terms of an order relation `$\prec$':
\begin{itemize}
\item[(1)] $x \prec y \prec z~\Rightarrow~ x \prec z$~(Transitivity: if $y$ is the outcome of $z$, and $x$ is the outcome
of $y$, then $x$ is the outcome of $z$).

\item[(2)] $x \prec y$ and $y \prec x ~ \Rightarrow x = y$~(Symmetry).

\item[(3)] Let $[[x,y]]$ denote the cardinality of the number of elements $z$ between $x$ and $y$ such that $x \prec z \prec y$, then
$[[x,y]] < \infty$~(Discreteness).
\end{itemize}
These relations can also be expressed in terms of a (locally finite) \emph{partially ordered set} as we do below; we then apply the inherent sense of causality to the structure provided by Chu spaces.  This is achieved by approximating
a highly structured space by a spatial model based on simplicial complexes and related posets as developed in \citet{GP1,GP2,GP3}, which we survey in part here, and in \S\ref{simplicial}.

\subsection{The Sorkin poset $P_{\F}$}

Suppose we are given a topological space $X$, viewed as a space of `observables',
and let us observe $X$ from a finite family of open sets (FFOS) $\F$, not necessarily covering $X$. This will represent \emph{a set of observations made on $X$}, where objects are observed in relationship to their attributes.
In this way, the FFOS partitions $X$ into the `attributes', and $X$ can then be regarded as a union of `zones' (see below) in which two points lie in the same zone if they share the very same attributes; in other words, they consist of clusters of points in the same open set of the FFOS, and thus cannot be distinguished by the corresponding set of observations.

We can define an equivalence relation ``$\sim_{\F}$'', by $x \sim_{\F} x'$, if and only if for all $U \in \F, ~ x \in U$ if and only if $x' \in U$. Thus two points are \emph{equivalent}
if all the observations from $\F$ attribute the same positive or negative result on both of them.
This is simply another way of stating the causality relations above. We can factor out by this equivalence relation
to obtain a quotient mapping:
\begin{equation}\label{quotient-1}
\pi_{\F}: X \lra X_{\F} = X/\sim
\end{equation}
where the quotient $X_{\F}$ can be regarded as encoding the observational data on $X$ in a way that organizes that data by ``merging'' equivalent observations.

The space $X_{\F}$ has topological type $T_0$\footnote{A topological space $X$ is said to be a \emph{$T_0$-space} if given distinct points of $X$, there is an open set of $X$ that contains one but not the other.
$T_0$-spaces naturally give rise to a partial order defined on the set of points of $X$, where $x \leq y$, if for each open set $U \subseteq X$,
$y \in U$ implies $x \in U$, and conversely.}
and corresponds to a \emph{partially ordered set} (\emph{poset}) denoted $P_{\F}$ and constructed as follows. We take $[x]_{\F}$
to be the equivalence class of $x \in X$, with $[x]_{\F} \leq [y]_{\F}$ if and only if for every open set $U \in \F$, if $y \in U$, then $x \in U$.
For practical purposes we consider the family $\F$ as finite, and $X_{\F}$ is a finite $T_0$-space. Each point $[x]_{\F}$ is contained in a minimal open set $U_{[x]}$ of $X_{\F}$, and $[x]_{\F} \leq [y]_{\F}$ if and only if $x \in U_{[y]}$. The resulting poset $P_{\F}$ contains much of the essential observational (or causal data) on $X$. Besides organizing that data, this poset will serve as a means of `measurement' (though not point-dependent) for gauging whether `objects' and `attributes' (or, `tokens' and `types') are seen as proximate to each other, or in contrast, are actually very far apart. Its structure is motivated by the ideas of \citet{Sorkin1}, as adopted by \citet{GP1}, describing how certain types of spaces can be approximated by `inverse limits' of more regular spaces.

Observe that the FFOS $\F$ determines a secondary topology $\tau(\F)$ on $X$ which is just the topology generated by $\F$. If $\tau(X)$ denotes
the original topology on $X$, then $\tau(\F) \subseteq \tau(X)$ with the closure with respect to $\tau(\F)$ interpreted as a proximity between
`zones' (see below). Let $\tau(P_{\F})$ denote the quotient topology on $P_{\F}$ such that the map
\begin{equation}\label{quotient-2}
\pi_{\F}: (X, \tau(\F)) \lra (P_{\F}, \tau(P_{\F})),
\end{equation}
is continuous (and then is seen to be an open map). We summarize the nomenclature in the following:

\begin{definition}\label{sorkin-poset}
Given $X$ and a FFOS $\F$, we say that the pair $(P_{\F}, \pi_{\F})$ is a \emph{Sorkin model of $X$ relative to $\F$}, in which case $P_{\F}$ is called the \emph{Sorkin poset} for $(X, \F)$. Given $x \in P_{\F}$, the corresponding subset $\pi_{\F}^{-1}(x) \subseteq X$ is called the \emph{zone determined by x}, which in general will be neither an open nor closed subset of $X$.
\end{definition}

\begin{definition}
Given two FFOSs $\F$ and $\G$ of a topological space $X$, we say that $\F$ is \emph{a Sorkin refinement} of $\G$ if $\G \subseteq \tau(\F)$.
\end{definition}
\noindent
From \citet[Prop. 11]{GP1} we observe that $\F$ is a refinement of $\G$, if and only if there exists a continuous surjective map
$\pi_{\F \G} : P_{\F} \lra P_{\G}$ such that the following diagram commutes
\begin{equation}
\xymatrix@C=4pc{ (X, \tau(X)) \ar[dr]_{\pi_{\G}} \ar[r]^{\pi_{\F}} & (P_{\F}, \tau(P_{\F}))
\ar[d]^{\pi_{\F\G}} \\ & (P_{\G}, \tau(P_{\G}))}
\end{equation}
that is, $\pi_{\G} = \pi_{\F\G} \circ \pi_{\F}$.

\subsection{A Chu FSA}

We start with a basic observation that any space $X$ along with a FFOS $\F$ can be formulated in terms of a Chu space $\sfC = (X, \in, \F)$ (namely, $X$ is the set of objects, $\in = \Vdash$, and $\F$ is the set of attributes). Thus an object $x \in X$ satisfies an attribute $U \in \F$, if $x \in U$.
This type of Chu space is said to be \emph{normal} \citep{Pratt2}. Thinking back to \S\ref{bi-ext}, we see that the quotient map $\pi_{\F}$ in \eqref{quotient-2}
is simply the universal map to the biextensional collapse of $\sfC = (X, \in , \F)$, and that
the Chu space $\sfC_{P_{\F}} = (\F, \in, \tau(P_{\F}))$ is itself biextensional, observing that the poset structure on $P_{\F}$ is given by
\begin{equation}
\hat{\a}(x) \leq \hat{\a}(y) \Longleftrightarrow \forall a \in C_{\sfa} ~(y \Vdash_{\sfC} a \Rightarrow x \Vdash_{\sfC} a) \Longleftrightarrow
\hat{\a}(x) \supseteq \hat{\a}(y).
\end{equation}
To avoid possible complications, we assume, as in \citep{GP1}, that $\F$ is suitably `sampled' and extensional (briefly, $\F$ has no repetitive
columns). Accordingly, we obtain a Chu space $\sfC = (C_{\sfo}, \Vdash_{\sfC}, C_{\sfa})$ consisting of a finite sample of attributes $\F$
resulting in a pair $(\sfC, \F)$, entitled a \emph{Chu FSA}. Given $(\sfC, \F)$, we call $\sfC_{\vert \F} = (C_{\sfo}, \Vdash_{\sfC}, \F)$ the \emph{corestriction} of $(\sfC, \F)$.

\subsection{Putting a topology on a Chu space}

The next step is to put a topology on a Chu space $\sfC$. Thus, we commence by saying that $\sfC$ is \emph{topologically closed} if the attributes $C_{\sfa}$ is a topology
on objects $C_{\sfo}$, meaning that $\sfC$ is normal, and $C_{\sfa}$ includes all unions and finite intersections. Without too much loss of generality, we assume that $\sfC$ is biextensional. Thus given $\sfC$, we have a topologically closed Chu space
\begin{equation}
\tau(\sfC) = (C_{\sfo}, \in, \tau(C_{\sfa})),
\end{equation}
which is naturally a topological closure of $\sfC$. Furthermore, there is a universal Chu morphism
$\tau: \tau(\sfC) \lra \sfC$, with $\tau_{\sfo}: C_{\sfo} \lra C_{\sfo}$ the identity, and $\tau_{\sfa}: C _{\sfa} \lra \tau(C_{\sfa})$ the inclusion. The point here is that $\tau(\sfC)$ contains the same informational (or observational) structure as the original $\sfC$, and in $\tau(\sfC)$
the information has been encoded by means of the propositional operations of geometric logic, and sampled via $\F \subset C_{\sfa}$.
Hence as proposed by \citet{GP1,GP3}, \emph{a Sorkin model $\sfC_{\F}$ for $(\sfC, \F)$ is defined to be the biextensional collapse$\backslash$Sorkin poset of $\sfC_{\vert \F}$}.  \citet[Prop. 18]{GP1} have also shown that any row $x$ in
$\sfC_{\F}$ consists of $n$ entries $0$ or $1$, and hence corresponds to a flow formula (\S\ref{chu-flow}):
\begin{equation}
(x \Vdash a_{i_1}) \wedge \cdots \wedge (x \Vdash a_{i_k}) \wedge \neg (x \Vdash a_{i_{k+1}}) \wedge
\cdots \wedge \neg (x \Vdash a_{i_n}),
\end{equation}
in turn showing that the rows of  $\sfC_{\F}$ can be considered to encode the elementary flow formulae:
\begin{equation}
\exists x (\bigwedge_{i \in \F_1} (x \Vdash a_i) \wedge \bigwedge_{i \in \F_2} \neg (x \Vdash a_i)),
\end{equation}
for given partitions $(\F_1, \F_2)$ of $\F$ .

Given Chu spaces $\sfC = (C_{\sfo}, \Vdash_{\sfC}, C_{\sfa})$ and  $\sfD = (D_{\sfo}, \Vdash_{\sfD}, D_{\sfa})$,
we say that $\sfC$ is a \emph{Sorkin refinement of $\sfD$} if there exists a Chu transform $\phi: \tau(\sfC) \lra \sfD$, which is the identity on objects ($\phi_{\sfo}(x) = x$). Further, any Chu space is a Sorkin refinement of itself, the Sorkin refinement is transitive, and if $\sfC$ is both extensional and a Sorkin refinement of $\sfD$, then the map $\phi$ is uniquely determined \citep[Prop. 20]{GP1}.

%%%%%%%%%%%%%%%%%%%%%%%%%%%%%%%%%%%%%%%%%%%%%%%%%%%%%%%%%%%%%%%%%%%%%%%%%%%%%%%%%%%%%%%%%%%%%%%%%%%%%%%%%%%%%%%%%%%%%%%%%

\section{Introducing simplicial methods on Chu spaces}\label{simplicial}

Given a topology on an information space, algebraic methods can be used to extend the topology into a geometry.  The resulting (discrete) geometry provides a natural representation of sets of observations made at different resolutions or scales, and hence a natural way to represent coarser- to finer-grained approximations of the topology.  Such approximations will provide, in \S\ref{mereotop}, the basis for a mereotopology of ``parts'' of objects.

\subsection{Simplicial complexes: basic definitions}\label{simpl-def}

We first introduce simplicial complexes as representations of observational data, following \citet{Cordier,Friedman,Goerss} and \citet{Spanier}.

\begin{definition}\label{simpl-def-1}
A \emph{simplicial complex} $K$ consists of a set $K_0$ of objects called the \emph{vertices} and a set of finite, non-empty subsets of $K_0$ called
the \emph{simplices}. The latter satisfy the condition that if $\sigma \subset K_0$ is a simplex, and if $\tau \subset \sigma$ (with $\tau \neq
\emptyset$), then $\tau$ is also a simplex. Simplicial complexes are objects in a category denoted $\mathbf{Simpl}$.
\end{definition}
\noindent
Simplicial complexes over an information space provide the structure needed to define an information geometry.  To each simplicial complex $K$ is associated the \emph{polyhedron} or \emph{geometric realization} of $K$, denoted $\vert K \vert$, formed from the set of all functions $K_0 \lra [0,1]$ satisfying:
\begin{itemize}
\item[i)]
if $\alpha \in \vert K \vert$, then the set $\{ v \in K_0: \alpha(v) \neq 0 \}$ is a simplex of $K$;
\item[ii)] $\sum_{v \in K_0} \alpha(v) = 1$.
\end{itemize}
Here each function $\alpha$ can be thought of as ``picking out'' a subset of vertices to be the vertices of some particular polyhedron.  These functions are normalized so that they ``pick out'' each vertex to the same extent.

For any simplex $s \in K$, there is an associated set $\vert s \vert =\{ \alpha \in \vert K \vert: \alpha (v) \neq 0 \Rightarrow v \in s \}$ as well as a set
$\langle s \rangle =\{\alpha \in \vert K \vert: \alpha (v) \neq 0 \Leftrightarrow v \in s \}$.  Often $\alpha(v)$ is called the \emph{$v^{th}$ barycentric coordinate of $\alpha$}, and the mapping $\vert K \vert \lra [0,1]$ defined by $p_v(\alpha) = \alpha(v)$ is the \emph{$v^{th}$ barycentric projection of $\alpha$}. With these coordinates, a metric $d$ can be defined on $K$ as given by
\begin{equation}
d(\alpha,\beta) =\big( \sum_{v \in K_0} (p_v(\alpha) - p_v(\beta))^2\big)^{\frac{1}{2}}.
\end{equation}
This distance $d(\alpha,\beta)$ measures the number of vertices shared between the polyhedra ``picked out'' by $\alpha$ and $\beta$, normalized to account for differences in the numbers of vertices of the two polyhedra.

\begin{definition}\label{simpl-def-2}
If $K$ and $L$ are two simplicial complexes, a simplicial mapping $f: K \lra L$ is a map $f_0: K_0 \lra L_0$ of vertex sets
that preserves simplices, meaning that if $\sigma \subset K_0$ is a simplex of $K$, then its image $f(\sigma) \subset L_0$ is a
simplex of $L$.
\end{definition}
Any simplicial complex $K$ gives rise, in a straightforward way, to a poset, namely the poset of its ``faces.''  The elements of this poset are the simplices of $K$, arranged according to the rule $\sigma \leq \rho$ if $\sigma$ is a face of $\rho$, that is, if $\sigma \subseteq \rho$ as subsets of the vertex set $K_0$.  Note that unions of ``adjoining'' faces are faces with this definition.  We will tacitly employ the (contravariant) functor relating the categories $\mathbf{Simpl} \lra \mathbf{Sets}$ to speak of a \emph{simplicial set}
corresponding to its underlying structure as a simplicial complex.

\subsection{Simplicial homotopy}

Two simplicial maps $f,g: X \lra Y$ of simplicial sets $X,Y$ are said to be \emph{homotopic} if
there exists a simplicial map $H: X \times I \lra Y$ (here $I=[0,1]$ the closed unit interval) such that $H_{\vert X \times \{0\}} = g$, and
$H_{\vert X \times \{1\}} = f$. In other words, we have $g= H \circ i_0$, and $f = H \circ i_1$, with respect to inclusion maps $i_0: X \times \{0\} \hookrightarrow X \times I$, and  $i_1: X \times \{1\} \hookrightarrow X \times I$. This is summarized by the following commutative diagram
\begin{equation}
\xymatrix@C=5pc{X \times \{1\} \ar[d]_{i_1} \ar[dr]^f \\ X \times I  \ar[r]^{H} & Y
\\ X \times \{0\} \ar[u]^{i_0} \ar[ur]_{g} }
\end{equation}
so that we have $H(x,0) = f(x)$ and $H(x,1) = g(x)$.\footnote{We adopt this natural definition of a simplicial homotopy as found in \cite{Friedman,Goerss}.
In \cite{Friedman} it is compared with the traditional, more technically oriented definition as seen in other textbooks on the subject.}

\subsection{The nerve of a relation}

For a space $X$ and open cover $\F$ of $X$, the \emph{\v{C}ech nerve $N(\F)$ of $\F$} is defined as the simplicial complex whose vertices are
the (open) sets in $\F$ and for which $\{U_0, \ldots, U_n \}$ is an $n$-simplex
of $N(\F)$ if and only if $\bigcap^n_{i=0} U_i \neq \emptyset$.  Intuitively, the \v{C}ech nerve is the simplicial complex over $\F$ comprising only \emph{connected} simplices.  As pointed out by \citet{GP1}, the face poset $P_{\F}$ as defined above bears a close relation with
$N(\F)$, but they need not be identified.
In a dual sense, there is the \emph{Vietoris complex $V(\F)$ of $(X, \F)$} in which the vertices are simply the points of $X$, and $\langle x_0, \ldots, x_n\rangle$ is an $n$-simplex if there exists a $U \in \F$ that contains them all, that is, $\{x_0, \ldots, x_n \} \subseteq U$.

\citet{Dowker} provides an abstraction in this setting, given a \emph{relation} $\R \subseteq X \times Y$ from $X$ to $Y$.  A simplicial complex $K_{\R}$, called \emph{the nerve of the relation} can be specified by: i) the vertices of $K_{\R}$ are those elements
$x \in X$ for which there exists a $y$ such that $(x,y) \in \R$, and ii)  the set $\{x_0, \ldots, x_n\} \in X$ is an $n$-simplex if and only if
there exists some $y$ such that $(x_i, y) \in \R$, for $0 \leq i \leq n$. From this it can be deduced that $N(\F)$ and $V(\F)$ each provide the same information about the open cover $\F$ up to homotopy.

\begin{example}
Let us exemplify some of these concepts for the basic case of the circle $S^1$, following \citet{Porter1}. Here we take an open covering $\F =\{U_1, U_2, U_3 \}$, relative to polar coordinates, with
\begin{equation}
\begin{aligned}
U_1 &= \big( - \frac{2\pi}{3}, \frac{2 \pi}{3} \big); \\
U_2 &= ``\big(0, -  \frac{2\pi}{3} \big)\text{''}  ~{\rm{i.e.}} ~ \big( 0, \pi \big] \cup \big( - \pi, - \frac{2\pi}{3} \big);\\
U_3 &= ``\big(\frac{2 \pi}{3}, 0 \big)\text{''} ~{\rm{i.e.}} ~ \big( \frac{2 \pi}{3}, \pi \big] \cup \big( - \pi, 0 \big).
\end{aligned}
\end{equation}
Every point of $S^1$, with the exception of $0, \frac{2 \pi}{3}$ and $-\frac{2 \pi}{3}$, is in exactly two of these, with a total of six equivalence classes. Choosing three representatives for the non-singleton classes gives the following minimal open sets:
$$
U_0 = U_1, ~ ~ U_{\frac{2 \pi}{3}} = U_2, ~ ~ U_{-\frac{2 \pi}{3}} = U_3
$$
$$
U_{\frac{\pi}{3}} = U_1 \cap U_2 := U_{12}, ~ ~ U_{-\frac{\pi}{3}} = U_1 \cap U_3 := U_{13}, ~ ~ U_{\pi} = U_2  \cap U_3 := U_{23}.
$$
Now we have a partially ordered set with associated Hasse diagram
\begin{equation}
\xymatrix{1 \ar@{-}[d] \ar@{-}[drr] & 2 \ar@{-}[dl] \ar@{-}[d]  & \ar@{-}[dl] 3 \ar@{-}[d] \\
12 & 23  &   13
}
\end{equation}
showing that $S^1_{\F}$ has 6 points, and the homotopy type of the former is that of $S^1$.

If we set $X = S^1$ and take the open cover $\F =\{U_1, U_2, U_3 \}$ as above, the vertices of $N(\F) = \langle U_1\rangle, \langle U_2 \rangle, \langle U_3 \rangle$, and the 1-simplices of $N(\F) = \langle U_1, U_2 \rangle, \langle U_1, U_3 \rangle, \langle U_2, U_3 \rangle$. Thus, $N(\F)$ may be represented schematically by the diagram
\begin{equation}
\xymatrix{& 2 & \\
1 \ar@{-}[ur]^{1,2} \ar@{-}[rr]_{1,3} &  & 3 \ar@{-}[ul]_{2,3}
}
\end{equation}
Recalling  that any simplicial complex determines a poset by subset inclusion of simplices, it can be seen that the resulting poset is the opposite of that representing $X_{\F}$.
\end{example}

\subsection{The \v{C}ech and Vietoris nerves of a Chu space}\label{chu-nerve}

In the context of a Chu space $\sfC= (C_{\sfo}, \Vdash_{\sfC}, C_{\sfa})$, the \emph{\v{C}ech nerve} is the simplicial complex denoted
$N(\sfC)$ with vertex set $C_{\sfa}$ and where a (non-empty) subset $\{a_0, \ldots a_p\}$ of $C_{\sfa}$ is a $p$-simplex if there
is an object $x \in C_{\sfo}$ satisfying $x \Vdash_{\sfC} a_i$, for $0 \leq i \leq p$. This is motivated by the fundamental principle
that for simplicial complexes, \emph{the nerve can be viewed as a set of instructions serving to construct (an approximation of) a space by fitting together
the individual geometric simplicies}.\footnote{The nerve specifies, in effect, which simplices adjoin each other by ``sharing an edge.''  \citet{Porter1} provides a number of illustrative examples providing an intuition leading to the discussion in \citet{GP1}.}  At the same time, the associated \emph{Vietoris nerve} $V(\sfC)$ is, in this context, the \v{C}ech nerve of the dual space $\sfC^{\perp}$.  Given that some $\{a_0, \ldots a_p\}$ comprises a simplex, the latter can be symbolized as $\langle a_0, \ldots a_p \rangle$.

As pointed out by \citet[\S3]{GP1} there may be possible complications in dealing with induced Chu morphisms, since the set $C_{\sfa}$ may be infinitely large. To remedy this situation, it is necessary to finitely sample the attributes by restricting consideration to a subset $\F$ of $C_{\sfa}$. Thus following \citet[Prop. 4]{GP1}, if $f= (f_{\sfo}, f_{\sfa}): \cP= (P_{\sfo}, \Vdash_{\cP}, P_{\sfa}) \lra \Q= (Q_{\sfo}, \Vdash_{\Q}, Q_{\sfa})$ is a morphism of Chu spaces, and $\F$ is an open cover (the
\emph{observations})
representing a finite sample of the attributes of $\Q$, i.e we have $\F \subseteq Q_{\sfa}$ and finite, then there is an induced map
\begin{equation}
f= (f_{\sfo}, f_{\sfa}): \cP= (P_{\sfo}, \Vdash_{\cP}, f_{\sfa}(\F)) \lra \Q= (Q_{\sfo}, \Vdash_{\Q}, \F).
\end{equation}
Furthermore, there are also induced simplicial maps given as follows: firstly, with respect to the Vietoris nerve, we have
\begin{equation}\label{v-simpl-1}
V(f): V(P_{\sfo}, \Vdash_{\cP}, f_{\sfa}(\F)) \lra V (Q_{\sfo}, \Vdash_{\Q}, \F),
\end{equation}
given by $V(f) \langle p_0, \ldots, p_n \rangle = \langle f_{\sfo}(p_0), \ldots, f_{\sfo}(p_n) \rangle$, and secondly, for any choice of splitting,
the function $f_{\sfa}: \F \lra f_{\sfa}(\F)$ (recall $\F \subseteq Q_{\sfa}$) induces a simplicial map with respect to the \v{C}ech nerve
\begin{equation}\label{c-simpl-1}
N(f): N(P_{\sfo}, \Vdash_{\cP}, f_{\sfa}(\F)) \lra N (Q_{\sfo}, \Vdash_{\Q},f_{\sfa}(\F)),
\end{equation}
given by $N(f) \langle f_{\sfa}(q_0), \ldots, f_{\sfa}(q_n) \rangle = \langle q_0, \ldots, q_n \rangle$.

\subsection{The Chu FSA and induced morphisms between nerves}\label{Chu-FSA}

The above simplicial procedures show that, for any Chu FSA $(\sfC, \F)$, there are two associated simplicial complexes $N(\sfC_{\vert \F})$ and
$V(\sfC_{\vert \F})$, along with the associated posets of their faces. Recall that we took $\sfC_{\F}$ to denote the
biextensional collapse$\backslash$Sorkin poset of $\sfC_{\vert \F}$. Let $\what{\F}$ denote a corresponding family of attributes. Again assuming $\sfC_{\vert \F}$ is extensional (no repeated columns in $\F$) then \cite[Th. 21]{GP1}, the quotient map
\begin{equation}
\pi_{\F}: \sfC_{\vert \F} \lra \sfC_{\F},
\end{equation}
exists, and there is an induced isomorphism
\begin{equation}
\pi^N_{\F}: N(\sfC, \F) \ovsetl{\cong} N(\sfC_{\vert \F}, \what{\F}),
\end{equation}
of simplicial complexes.

Intuitively, representing a set of observations of an FSA of a Chu space by a simplicial complex renders them a coarse-graining of the underlying Chu space, with the ``grain size'' determined by the number of vertices in the \v{C}ech nerve.  The adjacency relations implicit in the \v{C}ech nerve provide this coarse-graining with a local geometry.

%%%%%%%%%%%%%%%%%%%%%%%%%%%%%%%%%%%%%%%%%%%%%%%%%%%%%%%%%%%%%%%%%%%%%%%%%%%%%%%%%%%%%%%%%%%%%%%%%%%%%%%%%%%%%%

\section{An excursion into Channel Theory I}\label{channel-I}

\subsection{Classifications: Tokens and Types}\label{classifications-1}

Situation Theory and Channel Theory \cite{Barwise1} provide a structure for describing informational relations in the setting of information flow through systems distributed across space and time.  They provide a conceptual and schematic generalization of the ontological notion of information as causal connection introduced by \citet{Dretske1}. An assumption lending realism to the approach is that the channels through which information flows may have implicit or unknown properties that influence the nature of the inferences they implement.

We focus here on Channel Theory, the fundamental concept of which is the idea of a classification relating tokens to the types that encompass them.

\begin{definition}\label{class-def}
A \emph{classification} $\A = \langle \rm{Tok}(\A), \rm{Typ}(\A), \Vdash_{\A} \rangle$ consists of a set $\rm{Tok}(\A)$ consisting of the \emph{tokens of $\A$}, a set $\rm{Typ}(\A)$ consisting of the
\emph{types of $\A$}, and a classification relation
\begin{equation}
\Vdash_{\A} \subseteq \rm{Tok}(\A) \times \rm{Typ}(\A),
\end{equation}
that classifies tokens to types.
\end{definition}

A classification $\A = \langle \rm{Tok}(\A), \rm{Typ}(\A), \Vdash_{\A} \rangle$ has the structure of a Chu space, that is, via the assignment  $(\rm{Tok}(\A), \Vdash_{\A}, \rm{Typ}(\A)) \mapsto (\rm{object}, \Vdash, \rm{attribute})$ or, as is more typical in \citet{Barwise1}, the `dual' form $(\rm{Tok}(\A), \Vdash_{\A}, \rm{Typ}(\A)) \mapsto (\rm{attribute}, \Vdash, \rm{object})$.  As will be seen in \S\ref{flip} below, these interpretations are interchangeable.  Let us also keep in mind that for Chu spaces, ``objects'' and ``attributes'' can be aptly replaced by terms such as
``events'' and ``states'', with $\Vdash$ then interpreted as selecting the events that occur in a given state or, alternatively, the states participating in a given event.

\begin{remark}\label{multivalued-2}
As for a $\sfK$-valued Chu space in \ref{multivalued-1}, we may speak of a \emph{$\sfK$-valued Classification} $\A = \langle \rm{Tok}(\A), \rm{Typ}(\A), \Vdash_{\A} \rangle$, with a classification relation
$\Vdash_{\A} \subseteq \rm{Tok}(\A) \times \rm{Typ}(\A) \lra \sfK$,
where $\Vdash_{\A}(a,b)$ is an element of $\sfK$.
\end{remark}

\begin{example}
Let $\A = \langle Pts, \mathbf{U}, \Vdash_{\A} \rangle$ where $Pts$ denotes points of a topological space, $\mathbf{U}$ denotes the open sets of that space, and $x \Vdash_{\A} U$ if and only if $x \in U$. Thus, points are classified by the open sets in which they are contained. The open sets may be classified by the points within them by reversing $\Vdash_{\A}$.
\end{example}

\begin{example}
 Following \citet{Allwein1}, let
 $$
 \mathbf{FOL} = \langle Models, Sentences, \Vdash_{\mathbf{FOL}} \rangle,
 $$
 where $Sentences$ are sentences in First Order Logic (FOL). $Models$ are models of FOL sentences, and $x \Vdash_{\mathbf{FOL}} S$ if and only if $x$ is a model of the sentence $S$. Here, there are various internal relations holding on both the set of sentences and that of models, but none are imposed as external conditions in this case without further modification. One could also reverse matters, by taking the Types to be Models, and the Tokens as Sentences, so that Sentences in this case would be classified by Models.
\end{example}

\subsection{Infomorphisms}\label{tokens-1}

Here we recall the idea of a Chu morphism in order to link the information between two given classifications
$\A = \langle \rm{Tok}(\A), \rm{Typ}(\A), \Vdash_{\A} \rangle$ and $\B = \langle \rm{Tok}(\B), \rm{Typ}(\B), \Vdash_{\B} \rangle$.  In this case it is useful to define ``switching relations'' $\overrightarrow{f}: \rm{Typ}(\A) \lra \rm{Typ}(\B)$ and $\overleftarrow{f}: \rm{Tok}(\B) \lra
\rm{Tok}(\A)$ that can be specified by introducing the Channel Theory concept of an \emph{infomorphism}.  Specifically:
\begin{definition}\label{infomorph-1}
Given two classifications $\A = \langle \rm{Tok}(\A), \rm{Typ}(\A), \Vdash_{\A} \rangle$ and $\B = \langle \rm{Tok}(\B), \rm{Typ}(\B), \Vdash_{\B} \rangle$, an \emph{infomorphism} $f: \A \rightleftarrows \B$, is a pair of contravariant maps
\begin{itemize}
\item[i)] $\overrightarrow{f}: \rm{Typ}(\A)  \lra  \rm{Typ}(\B)$

\item[ii)] $\overleftarrow{f}: \rm{Tok}(\B) \lra \rm{Tok}(\A)$
\end{itemize}
such that for all $b \in \rm{Tok}(\B)$, and for all $a \in \rm{Typ}(\A)$, we have
\begin{equation}
\overleftarrow{f}(b) \Vdash_{\A} a, ~\text{if and only if}~ b \Vdash_{\B} \overrightarrow{f}(a).
\end{equation}
This last condition may be schematically represented by:
\begin{equation}\label{info-diagram-1}
\xymatrix@!C=3pc{\rm{Typ}(\A) \ar[r]^{\overrightarrow{f}}   & \rm{Typ}(\B) \ar@{-}[d]^{\Vdash_{\B}} \\
\rm{Tok}(\A) \ar@{-}[u]^{\Vdash_{\A}}  & \rm{Tok}(\B) \ar[l]_{\overleftarrow{f}}}
\end{equation}
\end{definition}

Note that this definition, given in \citet{Barwise1}, employs the `dual' interpretation of types as objects and tokens as attributes.  Interpreting tokens as objects and types as attributes yields infomorphisms with the usual Chu-morphism arrow directions.

\begin{remark}
In the context of situations, `attributes' can be interpreted as statements of `situation types'. In the Dretske spirit, to say that ``$x$
is $T_1$'' transmits information that ``$y$ is $T_2$'' can be represented as an informorphism representing these classification statements.  Here the content
of information such as $(T_1,T_2)$ is defined as the `type', and the carrier of the respective types, such as $(x,y)$, is defined
as the `token'.
\end{remark}

\begin{example}
Let $\mathbf{M} = \langle Messages, Contents, \Vdash_{\mathbf{M}} \rangle$ where Messages are classified by their Contents \citep{Allwein1}. Suppose we have another such classification $\mathbf{M}' = \langle Messages', Contents', \Vdash_{\mathbf{M}'} \rangle$. An infomorphism $f: \mathbf{M} \lra \mathbf{M'}$ may represent a function decoding messages from $\mathbf{M}'$ to messages in $\mathbf{M}$, so that whatever can be noted about the translation, may be mapped into something noted in the original message. That is, $m^f \Vdash_{\mathbf{M}} C \Leftrightarrow m \Vdash_{\mathbf{M}'} C^f$.
\end{example}

\begin{example}
Here is an example from decision theory \citep[\S2.3]{Allwein}. Let $\mathbf{s}$ be a classification of propositional logic and its
model states, and let $f$ represent a decision which evaluates a state $s$ and the agent making the decision (e.g. ``either walk home, or take the
bus home''). Let $\mathbf{O}$ be the classification of outcomes, and let $s^f$ represent a particular outcome of the decision of choosing either option (either ``walk home'' or ``take a bus home''). A proposition, denoted $Q$ over outcomes (in accordance with a slogan such as ``Keeping Fit'') characterizes them, and let $Q^f$ be the proposition categorizing all of the states in which $Q$ is satisfied. Thus, with respect to the above scheme of infomorphisms, we set
the classification $\A = \mathbf{O}$, and $\B = \mathbf{S}$, and \eqref{info-diagram-1} thus leads to
\begin{equation}\label{info-diagram-2}
\xymatrix@!C=3pc{{\rm{Typ}}(\mathbf{O}) \ar[r]^{\overrightarrow{f}}   & {\rm{Typ}}(\mathbf{S}) \ar@{-}[d]^{\Vdash_{\mathbf{S}}} \\
{\rm{Tok}}(\mathbf{O})  \ar@{-}[u]^{\Vdash_{\mathbf{O}}}  & {\rm{Tok}}(\mathbf{S}) \ar[l]_{\overleftarrow{f}}}
\end{equation}
in which case the infomorphism condition is expressed by $s \Vdash_{\mathbf{S}} Q^f$ if and only if $s^f \Vdash_{\mathbf{O}} Q$.
\end{example}

\begin{remark}\label{tokens-states}
It is worth noting that in the framework of infomorphisms, there is a natural mapping between tokens $A$ and the set of informational states:
\begin{equation}
A \lra \sfS(A).
\end{equation}
For instance, as pointed out by \citet[\S2.5]{Barwise1}, the truth classification of a first order language $L$ is the classification whose types are the sentences of $L$, and
the tokens are the $L$-structures. In which case, the classification relation is defined by $N \Vdash \vp$, if and only if $\vp$ is true in the structure of $L$ \cite[Example 4.3]{Barwise1}.
\end{remark}

\subsection{Information channels}

An \emph{information channel} $\mathbf{Chan}$ consists of an indexed family $\{f_i: \A_i \rightleftarrows  \mathbf{C}\}_{i \in \mathcal{I}}$ of infomorphisms
having a common codomain $\mathbf{C}$ called the \emph{core of the channel $\mathbf{Chan}$}:
\begin{equation}
\xymatrix{&\mathbf{C} &  \\
\A_1 \ar[ur]^{f_1} & \A_2 \ar[u]_{f_2}  & \ldots ~\A_i \ar[ul]_{f_i}~ \ldots
}
\end{equation}
The core $\mathbf{C}$ is essentially a carrier of information flow between the $f_i$ and hence between the classifications $\A_i$, and is itself a classification in the above sense.
The tokens $\rm{Tok}(\mathbf{C})$ of $\mathbf{C}$ are called \emph{connections}. A connection $c$ is said to \emph{connect} the tokens $f_i(c)$ of the classifications $\A_i$ for $i \in \mathcal{I}$ (note that tokens are mapped from $\mathbf{C}$ to the $\A_i$ in the `dual' interpretation of \citet{Barwise1}).  A channel with index set $\{0, \ldots, n-1\}$ is called an \emph{$n$-ary} channel.  Composing information channels amounts to taking their limit and the channels themselves may be refinable by straightforward
categorical means \citep{Barwise1}.

The above definition extends the intuitive picture of a channel as a wire connecting two agents (i.e. classifiers) to the idea of a blackboard or other shared memory via which multiple classifiers exchange information.  The shared memory $\mathbf{C}$ being itself a classifier provides it with a structure that can affect how information is written to and read from it.  Consistent with the essentially causal notion of information of \citet{Dretske1}, the connections between the tokens of different classifiers are purely functional; no overarching semantics is assumed.  How such a semantics can be constructed, \emph{post hoc}, given a channel is discussed in \S\ref{ontologies} below.

\begin{example}\label{idealization}
For instance, if we have a binary channel $\mathbf{Chan} = \{f: \A \rightleftarrows  \mathbf{C}, g: \B \rightleftarrows  \mathbf{C} \}$, then the local logic (see \S\ref{local} below) on $\B$ induced by $\mathbf{Chan}$, is the logic $\Lg_{\mathbf{Chan}}(\B) = g^{-1}[f [\Lg(\A)]]$ (in \citet[14.1]{Barwise1} classifications $\A$ and $\B$ are interpreted as ``idealization'' and ``reality'',  respectively). This induced logic can be characterized by \citep[Prop. 14.2]{Barwise1}:
\begin{itemize}
\item[(i)] A partition $\langle \Gamma, \Delta \rangle$ of $\rm{Typ}(\B)$ is consistent in $\Lg_{\mathbf{Chan}}(\B)$ if and only if
$\langle f^{-1}[g[\Gamma]],  f^{-1}[g[\Delta]] \rangle$ is the state description of some $a \in \rm{Tok}(\A)$.

\item[(ii)] A token $b \in \rm{Tok}(\B)$ is normal in $\Lg_{\mathbf{Chan}}(\B)$, if and only if it is connected to some token
$a \in \rm{Tok}(\A)$.
\end{itemize}
\end{example}

\subsection{Cocone of infomorphisms}\label{info-cocone}

A network of infomorphisms between classifications admits a limit classification that gathers all of the information in the network into a single classification (a \emph{cone}) with projections back down to the individual classifications \cite{Barwise1}. There is  a dual notion which we will describe as follows.
A channel is an instance of the more general category-theoretic concept of a \emph{cocone} being the core classification.  To motivate the construction, consider any finite directed graph with vertex labels $1, 2, ..., n$ and edge labels $f_{ij}$.  Considering such a graph to represent a network of communicating agents is, from a category-theoretic perspective, invoking a map $G$ (technically, a functor from the category of finite directed graphs to the category of classifications) that constructs a classification $G(i)$ at each vertex and an infomorphism $G(f_{ij})$ at each edge.   A \emph{commuting finite cocone} of infomorphisms (e.g. \citet{Barwise1,Allwein}) is a finite network of classifications $G(i)$ and infomorphisms $G(f_{ij})$, a vertex classification $\mathbf{C}$, and a collection of infomorphisms $g_i: G(i) \lra \mathbf{C}$.
\begin{equation}\label{cocone-diagram}
\xymatrix{& & &\mathbf{C} & \\
&\ar[urr]^{g_1} G(1) \dots &  G(i) \ar[ur]_{g_i}  \ar[rr]_{G(f_{ij})} &  & G(j) \ar[ul]^{g_j} & \ar[ull]_{g_n} \ldots  G(n)
}
\end{equation}
The commutativity condition is that for all $f_{ij}$, we have $g_i = g_j \circ G(f_{ij})$. The base of the cocone consists of the classifications and infomorphisms constructed by $G$; the \emph{cocone vertex classification} $\mathbf{C}$ together with the maps $g_i$, is a channel. Note that in the complementary sense, a \emph{commuting finite cone} of infomorphisms consists of a finite network of classifications $G(i)$ and infomorphisms $G(f_{ji})$, a vertex classification $\mathbf{C}$, and a collection of infomorphisms $g_i: G(i) \lra \mathbf{C}$. For all $f_{ji}$, we have $g_i = G(f_{ji}) \circ g_j$, and all arrows in the above diagram are reversed.

In short, we have this colimit classification into which there are infomorphisms from each constituent classification, and this colimit contains all of the information that is common to the different component parts of the network.
The generalization from channel to cocone will prove useful in the discussion of ``minimal covers'' of distributed systems in \S\ref{distributed}.  We further characterize cocones and relate them to colimits in the descriptive discussion of \S\ref{colimits}. We then apply these concepts to model abstraction-based categorization in \S\ref{tt-flow} and to mereological categorization in \S\ref{mereological}.

\subsection{The flip of a classification}\label{flip}

For any classification $\A$, the \emph{flip} of $\A$, is the classification $\A^{\perp}$ whose tokens are the types of $\A$, whose types are the tokens of $\A$, such that $\a \Vdash_{\A^\perp} a$ if and only if $a \Vdash_{\A} \a$ (see \citet[\S4.4]{Barwise1}). In deciding how to model a classification there may be epistemological questions, e.g. the types in question are given as things or attributes we may know about, and the tokens are those things we wish to have information about. The fact that the flip of a classification is a classification (and both can be treated as Chu spaces) and these behave well under infomorphisms, means that a situation involving types or tokens, can be dualized to tokens or types. For instance, ``the type set of a token'' dualizes to ``the token set of a type''. Effectively, $f: \A \rightleftarrows \B$ is an informorphism if and only if $f^\perp: \B^\perp \rightleftarrows \A^\perp$ is an infomorphism \citep[Prop. 4.19]{Barwise1}. Further, $(\A^\perp)^\perp = \A$ with $(f^\perp)^\perp = f$, and $(fg)^\perp = g^\perp f^\perp$ \citep[Prop. 4.20]{Barwise1}. Thus for $f, f^\perp$ respectively, we have the diagrams

\begin{equation}
\xymatrix@!C=3pc{\rm{Typ}(\A)\ar[r]^{\overrightarrow{f}}   & \rm{Typ}(\B) \ar@{-}[d]^{\Vdash_{\B}} \\
\rm{Tok}(\A) \ar@{-}[u]^{\Vdash_{\A}}  & \rm{Tok}(\B) \ar[l]_{\overleftarrow{f}}} ~ ~  ~
\xymatrix@!C=3pc{\rm{Tok}(\B) \ar[r]^{\overleftarrow{f}}   & \rm{Tok}(\A) \ar@{-}[d]^{\Vdash^{-1}_{\A}} \\
\rm{Typ}(\B) \ar@{-}[u]^{\Vdash^{-1}_{\B}}  & \rm{Typ}(\A) \ar[l]_{\overrightarrow{f}}}
\end{equation}

\subsection{The nerve of a classification}

As any classification is a Chu space, any operation defined for Chu spaces is meaningful for a classification. A finite sample $\F$ of `attributes', for example, becomes a finite sample of `tokens' (or `types').  Simplicial complexes are defined as in \S\ref{simpl-def} and nerves as in \S\ref{chu-nerve}.  The \v{C}ech nerve of a classification $\A = \langle \rm{Tok}(\A), \rm{Typ}(\A), \Vdash_{\A} \rangle$, for example, is the simplicial complex
$N(\A)$ with vertex set $\rm{Tok}(\A)$, where a (non-empty) subset $\{b_0, \ldots b_p\}$ of $\rm{Tok}(\A)$ is a $p$-simplex if there
is a type $v \in \rm{Typ}(\A)$ satisfying $v \Vdash_{\A} b_i$, for $0 \leq i \leq p$. The notion of the Vietoris nerve follows in a similar way as in \S\ref{chu-nerve}.

If $\F = \rm{Tok}(\B)$ is a finite sample of tokens of a classification $\B$, we have the infomorphism \eqref{info-diagram-1}:
\begin{equation}
f=(\overleftarrow{f}, \overrightarrow{f}): (\overleftarrow{f}(\F), \rm{Typ}(\A), \Vdash_{\A}) \lra (\F, \rm{Typ}(\B), \Vdash_{\B}),
\end{equation}
while for a finite sample $\G \subseteq \rm{Typ}(\A)$, we have:
\begin{equation}
f=(\overleftarrow{f}, \overrightarrow{f}): (\rm{Tok}(\A), \G, \Vdash_{\A}) \lra (\rm{Tok}(\B), \overrightarrow{f}(\G), \Vdash_{\B}),
\end{equation}
We will henceforth assume that finite samples of tokens (and types) have been taken, so that we may consider, as in \eqref{c-simpl-1}, well-defined simplicial maps
\begin{equation}\label{class-nerve-1}
N(f): N(\A) \lra N(\B),
\end{equation}
as defined for the \v{C}ech nerve of the corresponding Chu spaces, here with respect to finite samples of tokens and types.

\subsection{Associating a theory with a classification}

Here and in the following sections we collect together some useful definitions from \citet{Barwise1} and \citet{Barwise2}, starting with \emph{sequents} and \emph{theories}:
\begin{definition}
Let $\Sigma$ be an arbitrary set (which may be viewed as set of \emph{types}). A binary relation $\vdash$ between subsets of $\Sigma$ is called a \emph{consequence relation on $\Sigma$}. A \emph{(Gentzen) sequent} is a pair $I= \langle \Gamma, \Delta \rangle$ of subsets of $\Sigma$
(here it is apt to view $\Gamma$ and $\Delta$ as sets of \emph{situation types}). A sequent $I= \langle \Gamma, \Delta \rangle$ is said to hold of a situation $s$ provided that if $s$ supports every type in $\Gamma$, then it supports some type in $\Delta$. A sequent $I$ is said to be \emph{information} about a set $S$ of situations if it holds at each $s \in S$; here again the causal notion of information flow is evident. Finally, a sequent is called a \emph{partition} of a set $\Sigma'$ if
$\Gamma \cup \Delta = \Sigma'$ and $\Gamma \cap \Delta = \emptyset$.
\end{definition}

\begin{definition}
A \emph{theory} is a pair $T= \langle \Sigma, \vdash_T \rangle$, where
$\vdash_T$ is a consequence relation on $\Sigma$. A \emph{constraint} of the theory $T$ is a sequent $\langle \Gamma, \Delta \rangle$
of $\Sigma$ for which $\Gamma \vdash_T \Delta$. A sequent $\langle \Gamma, \Delta \rangle$ is \emph{$T$-consistent} if $\Gamma \nvdash_T \Delta$.
\end{definition}
\noindent
Here again, the idea of some aspects of a situation either causally requiring or merely causally allowing other aspects of a situation makes this definition clear.

Each classification has a theory associated with it in the following way (see also Definition \ref{local-2} below). A \emph{theory} $\Th(\A) = (\Sigma_{\A}, \vdash_{\A})$ generated by a classification $\A$, satisfies for all types $\a$ and all
sets $\Gamma, \Gamma', \Delta, \Delta', \Sigma', \Sigma_0, \Sigma_1$ of types \cite[Prop 9.5]{Barwise1}:
\begin{itemize}
\item[(1)] \emph{Identity}: $\a \vdash \a$.

\item[(2)] \emph{Weakening}: If $\Gamma \vdash \Delta$, then $\Gamma, \Gamma' \vdash \Delta, \Delta'$.

\item[(3)] \emph{Global cut}: If $\Gamma, \Sigma_0 \vdash \Delta, \Sigma_1$, for each partition $\langle \Sigma_0, \Sigma_1 \rangle$
of $\Sigma$, then $\Gamma \vdash \Delta$.
\end{itemize}
More generally, we can say that a theory $T = \langle \Sigma, \vdash_T \rangle$ is \emph{regular} if it satisfies the above three conditions.

\subsection{Local logics}\label{local}

We can specify a classification of a regular theory $T$ as given by:
\begin{itemize}
\item[(1)] $\rm{Typ}\mathnormal{(\Cl(T))} = \rm{Typ}\mathnormal{(T)}$.

\item[(2)] $\rm{Tok}\mathnormal{(\Cl(T)) = \{ \langle \Gamma, \Delta \rangle: \langle \Gamma, \Delta \rangle~
\text{is a $T$ consistent partition of}}~\rm{Typ} \mathnormal{(T) \} }$.

\item[(3)] $\langle \Gamma, \Delta \rangle \vdash_{\Cl(T)} \a $ if and only if $\a \in \Gamma$.
\end{itemize}
Indeed, for any regular theory it can be seen that $\Th(\Cl(T)) = T$.

Since we are mainly considering distributed systems, and information processing entails computation within a logical
framework, the following system of local logics \cite[Def. 12.1]{Barwise1} is one suited to representing various types of state spaces.
\begin{definition}\label{local-1}
\emph{A local logic} consists of a triple $(\cL = \langle \rm{Tok}(\cL), \rm{Typ}(\cL), \Vdash_{\cL} \rangle, \vdash_{\cL}, \sfN_{\cL})$ in which we have:
\begin{itemize}
\item[(1)] a classification $\cL = \langle \rm{Tok}(\cL), \rm{Typ}(\cL), \Vdash_{\cL} \rangle$,

\item[(2)] a regular theory ${\rm{Th}}(\cL) = (\rm{Typ}(\cL), \vdash_{\cL})$, and

\item[(3)] a subset $\sfN_{\cL} \subset \rm{Tok}(\cL)$, called \emph{the normal tokens} of $\cL$, which satisfy all of the constraints
of the theory $\rm{Th}(\cL)$ in (2).
\end{itemize}
\end{definition}

\begin{definition}\label{local-2}
Let $\A$ be a classification. The \emph{local logic generated by $\A$}, denoted ${\Lg}(\A)$, has classification $\A$, a regular theory ${\Th}(\A) = (\rm{Typ}(\A), \vdash_{\A})$, and all its tokens are normal. A logic is said to be \emph{natural} if it is generated by some classification.
\end{definition}
In fact, for any local logic $\mathcal{L}$ on $\A$, we have $\mathcal{L} = {\Lg}(\A)$ by \citet[Prop. 12.7]{Barwise1}. This relationship will be exemplified in Example \ref{building} in the context of ontologies.

Intuitively, a local logic is ``local'' to the classification that generates it.  Infomorphisms allow mapping the local logic of one classification to that of another; hence we can think of channels as supporting the flow of locally-defined logical relations between classifications.  Recalling from \S\ref{Chu-FSA} that any classification can be interpreted as defining a coarse-graining and hence a ``scale'' at which information is being organized and represented, each local logic can be thought of as a ``logic at some level of description.''  This interpretation is made explicit in \S\ref{cccd-logic-1} and \S\ref{mereotop} below.  As any classification can also be interpreted as describing a state space (\S\ref{chu-info-1}), one can further associate a canonical logic $\Lg(\sfS)$ to any state space $\sfS$. Specifically, if $\sfS$ is such a state space with a classification of events $\Evt(\sfS)$, then we can speak of an $\sfS$-logic as a logic $\mathfrak{L}$ on this classification such that $\Lg(\sfS) \subseteq \mathfrak{L}$, with the intuition that this $\sfS$-logic can accommodate the theory that is implicit to the structure of $\sfS$ \citep[\S16]{Barwise1}.

\begin{remark}
In \citet{Barwise2}, $\cL$ is called an \emph{information context} and $\vdash$ is a binary relation relating sets of situation types.  In this case $\sfN_{\cL}$ is said to be a set of \emph{normal situations}. Intuitively, these are the situations that the available information concerns. They may comprise all or only some of the situations satisfying the information. For instance, we may start with some set of normal situations accounting for an individual's experiences to date, and then the information context consists of all the sequents satisfied by, i.e. consistent with, this experience.  Stepping outside of the context generates ``surprise'' in the sense of expectation violation (cf. \citet{Friston2}).
\end{remark}

Next, we look to what extent an infomorphism between classifications will respect the associated local logics. This is given by the following (\citet[12.3]{Barwise1}):
\begin{definition}\label{logic-info}
A \emph{logic infomorphism} $f: \mathcal{L}_1 \leftrightarrows \mathcal{L}_2$, consists of a covariant pair $f= \langle f\sphat, f\spcheck \rangle$ of functions satisfying
\begin{itemize}
\item[(1)] $f: \Cl(\mathcal{L}_1) \leftrightarrows \Cl(\mathcal{L}_2)$ is an infomorphism of classifications.

\item[(2)] $f\sphat : \mathsf{Th}(\mathcal{L}_1) \lra \mathsf{Th}(\mathcal{L}_2)$ is a theory interpretation, and

\item[(3)] $f\spcheck[\sfN_{\mathcal{L}_2}] \lra \sfN_{\mathcal{L}_1}$
\end{itemize}
\end{definition}
For further consequences of this definition, see \citet{Barwise1}.

\subsection{Boolean Classification}

We can exemplify local logics following \citet{Barwise1,Barwise3} in terms of a Boolean classification $\A = \langle S, \Sigma, \Vdash, \wedge, \neg \rangle$. Here we have a set $S$ of \emph{situations} (tokens as objects) and a set $\Sigma$ of \emph{propositions} (types as attributes). This leads to a \emph{Boolean local logic} $\cL = \langle \A, \vdash, N \rangle$ where $N \subseteq S$ consists of the normal situations \citep{Barwise1,Barwise3}. If $s \in N$ is a normal situation, $\Gamma \vdash \Delta$ and $s \Vdash p$,
for all $p \in \Gamma$, then $s \Vdash q$ for some $q \in \Delta$. A partial ordering ``$\subseteq$'' on local logics $\cL_1, \cL_2$ on a fixed classification of $\A$ is defined by $\cL_1 \subseteq \cL_2$, if and only if
\begin{itemize}
\item[(1)] for all sets $\Gamma, \Delta$ of propositions, $\Gamma \vdash_{\cL_1} \Delta$ entails $\Gamma \vdash_{\cL_2} \Delta$, and
\item[(2)] every situation of $\A$ that is normal in $\cL_2$ is also normal in $\cL_1$.
\end{itemize}
Now, if we take any set of sequents $T$, the logic $\Lg(\A^T)$ generated by $T$ on $\A$:
\begin{itemize}
\item[(i)] has as normal situations all of those situations that satisfy the sequents in $T$,

\item[(ii)] has as constraints all sequents satisfied by all situations in $N$, and

\item[(iii)] has as normal situations all of the situations of $\A$ satisfying these constraints.
\end{itemize}
Given a fixed Boolean classification \citep{Barwise3}:
\begin{itemize}
\item[(1)] If $T_0 \subseteq T_1$, then $\Lg(\A^{T_0}) \subseteq \Lg(\A^{T_1})$.

\item[(2)] If $N_0 \supseteq N_1$, then $\Lg(\A_{N_0}) \subseteq \Lg(\A_{N_1})$.
\end{itemize}
\begin{example}
Suppose $\A$ is a classification of bird sightings (observations), and
$N$ consists of the actual sightings to date. Then $\Lg(\A_N)$ has as constraints all sequents satisfied by all those bird sightings to date, and the normal situations consist of all bird sightings that satisfy all of these constraints, a set that clearly contains $N$. This logic may entail the constraint BIRD $\vdash$ FLY, a constraint that holds as long as the situations encountered are meaningfully compatible with the elements of $N$. But now, suppose a penguin is observed. It will lie outside of the normal situations since it violates BIRD $\vdash$ FLY. This observation uncovers a new set $N' \supset N$, and accordingly their logics satisfy $\Lg(\A_{N'}) \subseteq \Lg(\A_N)$. There will be fewer constraints tenable in $\Lg(\A_{N'})$ as we can see, since BIRD $\nvdash$ FLY in this new logic.
\end{example}

%%%%%%%%%%%%%%%%%%%%%%%%%%%%%%%%%%%%%%%%%%%%%%%%%%%%%%%%%%%%%%%%%%%%%%%%%%%%%%%%%%%%%%%%%%%%%%%%%%%%%%%%%%%%

\section{An excursion into Channel Theory II}\label{channel-II}

Classifications and channels have been applied widely in theoretical computer science; we briefly review some of these applications here as motivations for applying these tools to perceptual processing.

\subsection{The information channel in a MLP network}\label{mlp}

One of the earliest ANNs studied was the \emph{Multilayer Perceptron} (MLP) network \cite{Rosenblatt,Rummelhart3}.
As with typical ANNs, it has an input layers ($I_i$), hidden layers ($H_i$), and an output layer ($O_i$) with weighted directional (in an MLP, exclusively feedforward) linkages between subsequent layers.
\citet{Kikuchi} develop a Chu space/Channel Theory representation of a 3-layer MLP, showing how the synaptic weights between layers form a channel; we follow their example closely.

Let $\mathbf{wo}_{ij}$ denote a synaptic weight between the $j$-th neuron in the hidden layer and the $i$-th neuron in the output layer.
Similarly, let $\mathbf{wh}_{jk}$ denote a synaptic weight between the $k$-th neuron in the input layer and the $j$-th neuron in the hidden layer.
Then for a given state function $f(x)$, the layers $O_i, H_i$ and $I_i$ are related in accordance with
\begin{equation}
O_i = f( \sum_j \mathbf{wo}_{ij} H_j) = f(\sum_j \mathbf{wo}_{ij} ~f(\sum_k \mathbf{wh}_{jk} I_k)).
\end{equation}
It is convenient to regard an MLP with a fixed topology as a map $\mathbf{F}: \mathbf{I} \lra \mathbf{O}$, from the input data space
$\mathbf{I} =\{I_i\}$ to the output data space $\mathbf{O} =\{O_i\}$, so that $\mathbf{F}$ is uniquely defined by a point in the parameter
space of weights $\mathbf{\Phi} = \{ \langle \mathbf{wh}, \mathbf{wo} \rangle \}$. In this way, a fixed topology on a MLP can be represented as $\mathbf{F}_{\langle \mathbf{wh}, \mathbf{wo} \rangle}$, once given $\langle \mathbf{wh}, \mathbf{wo} \rangle \in \mathbf{\Phi}$.

Next consider the sub-parameter spaces $\mathbf{\Phi}_h = \{\langle \mathbf{wh} \rangle \}$ and
$\mathbf{\Phi}_o = \{\langle \mathbf{wo} \rangle \}$, and the following three classifications
\begin{itemize}
\item[(1)] $\A = (\rm{Tok}(\A), \rm{Typ}(\A), \Vdash_{\A})$ (the states of ``cognition'' i.e. of $\mathbf{O}$)

\item[(2)] $\B = (\rm{Tok}(\B), \rm{Typ}(\B), \Vdash_{\B})$ (the states of the ``environment'' i.e. of $\mathbf{I}$)

\item[(3)] $\Ce = (\rm{Tok}(\Ce), \rm{Typ}(\Ce), \Vdash_{\Ce})$ (the states of the network)
\end{itemize}
where for the tokens $A = \mathbf{\Phi}_h,~ B = \mathbf{\Phi}_o$ and $C = \mathbf{\Phi}$, we define projections
\begin{equation}
\begin{aligned}
g_h &: \mathbf{\Phi} \lra \mathbf{\Phi}_h, ~
\langle \mathbf{wh}, \mathbf{wo} \rangle \mapsto \langle \mathbf{wh} \rangle \\
g_o &: \mathbf{\Phi} \lra \mathbf{\Phi}_o, ~
\langle \mathbf{wh}, \mathbf{wo} \rangle \mapsto \langle \mathbf{wo} \rangle
\end{aligned}
\end{equation}
as well as the obvious respective inclusions $f_h: \mathbf{\Phi}_h \lra \mathbf{\Phi}, ~f_o: \mathbf{\Phi}_o \lra \mathbf{\Phi}$.
Thus, we obtain a core (and vertex of a cocone) that is in $\Ce$ along with an information channel
\begin{equation}
\xymatrix{&\Ce = \langle \mathbf{wh}, \mathbf{wo} \rangle &  \\
\langle \mathbf{wh} \rangle \ar[ur]^{f_h} &   & \langle \mathbf{wo} \rangle \ar[ul]_{f_o}
}
\end{equation}
where, as shown by \citet{Kikuchi}, an algorithm for modifying $\langle \mathbf{wh}, \mathbf{wo} \rangle$
corresponds to a local logic on $\Ce$. This method can be developed in terms of Distributed Systems, as explained below in \S\ref{distributed}
(see also Remark \ref{gnw-remark}).

\subsection{Distributed Systems}\label{distributed}

Following the development of \citet[Ch. 6]{Barwise1} we provide an example of Dretske's ``Xerox principle'', namely, that information flow is transitive. Consider two information channels sharing common infomorphisms.  Suppose the first channel represents the examination of a map, capturing the notion of a person's perceptual state carrying information about the map being examined. The second channel represents the informational relationship between the map and the region it depicts. These coupled channels can be illustrated:
\begin{equation}
\xymatrix{&\mathbf{B}_1 &  & \mathbf{B}_2 \\
\A_1 \ar[ur]^{f_1} & & \A_2\ar[ul]_{f_2}\ar[ur]^{f_3} &  & \A_3 \ar[ul]_{f_4}
}
\end{equation}
Recall that the elements of $\rm{Tok}(\mathbf{B}_1)$ are `connections'. In this case the connections are spatio-temporal perceptual events involving
persons in $\rm{Tok}(\A_1)$ looking at maps in (i.e. elements of) $\rm{Tok}(\A_2)$. The connections of the second channel, elements of $\rm{Tok}(\mathbf{B}_2)$, are spatio-temporal events involving making the maps in $\rm{Tok}(\A_2)$ to represent various regions in
$\rm{Tok}(\A_3)$. Under certain circumstances, a person's perceptual state carries information about a particular mountain, given that the person is reading a map showing that mountain. In this regard, we may reasonably consider $\A_1$ as the \emph{idealized space} of the physical space $A_3$.

The next step is to construct another channel that fits both $\mathbf{B}_1$ and $\mathbf{B}_2$ together. The process is: i) choose a person, ii) go to a map she is reading, and then iii) proceed to the region shown on that map. Here we will restrict to pairs $c=(b_1, b_2)$ ((perceptual event, map-making)), so that $f_2(b_1) = f_3(b_2)= a_2$ holds, i.e. there is just one map in question. In this way, types $\be_1 = f_2(\a_2)$ and $\be_2 = f_3(\a_2))$ are equivalent since they are both translations of $a_2$. This is built into the channel by identifying $\be_1$ and $\be_2$ (cf. the biextensional collapse of a Chu space) and gives rise to a new classification $\Ce$ having the above tokens and $\rm{Typ}(\Ce) = \rm{Typ}(\mathbf{B}_1 \cup \mathbf{B}_2)$, but identifying types originating from a common type in $\A_2$.
Thus we obtain a new channel with core $\mathbf{C}$, connecting $\mathbf{B}_1$ and $\mathbf{B}_2$, as depicted below:
\begin{equation}
\xymatrix{& & \mathbf{C}  & \\ &\mathbf{B}_1\ar[ur]^{g_1}  &  & \mathbf{B}_2 \ar[ul]_{g_2} \\
\A_1 \ar[ur]^{f_1} & & \A_2\ar[ul]_{f_2}\ar[ur]^{f_3} &  & \A_3 \ar[ul]_{f_4}
}
\end{equation}
The channel infomorphisms are defined via composition $h_1 = g_1 f_1$, and $h_3= g_2 h_4$, so linking $\A_1$ to $\A_3$.

In general, we have
\begin{definition}\label{distributed-def}
A \emph{Distributed System} $\A$ consists of an indexed family $\Cl(\A) = \{\A_i\}_{i \in \mathcal{I}}$ of classifications, together
with a set $\Inf(\A)$ of infomorphisms having both domain and codomain in $\Cl(\A)$.
Each classification may be taken to support a local logic, along with the core of the channel.
\end{definition}

\begin{definition}\label{minimal}
An information channel $\mathbf{Chan} = \{h_i: \A_i \rightleftarrows \mathbf{C} \}$ \emph{covers} a distributed system $\A$ if $\Cl(\A)
= \{\A_i\}_{i \in \mathcal{I}}$, and for each $i,j \in \mathcal{I}$, and for each infomorphism $f: \A_i \rightleftarrows \A_j$ in
$\Inf(\A)$, the following diagram commutes
\begin{equation}
\xymatrix{& \mathbf{C} & \\
\A_i \ar[ur]^{h_i} \ar[rr]_{f} &  & \A_j \ar[ul]_{h_j}
}
\end{equation}
$\mathbf{Chan}$ is said to be a \emph{minimal cover} if it covers $\A$, and for every other channel $\D$ covering $\A$ there is a unique infomorphism from
$\mathbf{C}$ to $\mathbf{D}$.
\end{definition}
A minimal cover of a system $\A$ converts the entire distributed system, consisting of multiple channels, into a single channel. Every distributed system has a minimal cover, and this cover is unique up to isomorphism \citep[Th. 6.5]{Barwise1}.

\begin{remark}\label{gnw-remark}
The constructions of \S\ref{mlp} and \S\ref{distributed} can be extended to parallel distributed processing (PDP) and to more general multi-layer, bidirectional ANNs (see e.g. \citet{Dawson1,Rogers,Rummelhart3}).  They are also applicable for modelling constructs of the massively parallel, competitively based, distributed system of the \emph{Global Neuronal Workspace} (GNW) as studied in \citet{Baars3,Dehaene1,Dehaene04,Wallace2005}. Such modelling can be implemented, for example, by the \emph{Learned Intelligent Distribution Agent} (LIDA) architecture \citep{Baars3,Franklin1,Friedlander}.  As observed by \citet{Maia}, feedforward and feedback projections in connectionist networks can engage selective attention toward more salient inputs, producing yet strong weighting, that can predict which of the competing elements will gain access to the GNW central core (cf. \citet{Friston2,Grossberg13,Shan3}, and the relationship to object-event files in \S\ref{episodic}).
\end{remark}

\subsection{Ontologies}\label{ontologies}

It is often useful, when describing events or processes in some domain, to represent the ontology of the domain explicitly as a type hierarchy.  Following \citet{Schorlemmer2005} (cf. \citet{Kalfoglou2,Kalfoglou1}):

\begin{definition}\label{ontology-1}
An \emph{ontology} is a tuple $\mathcal{O} = (C, \leq, \perp, |)$ where
\begin{itemize}
\item[(1)] $C$ is a finite set of concept symbols;

\item[(2)] $\leq$ is a reflexive, transitive, and anti-symmetric relation on $C$ (a partial order);

\item[(3)] $\perp$ is a symmetric and irreflexive relation on $C$ (disjointness);

\item[(4)] $|$ is  symmetric relation on $C$ (coverage)
\end{itemize}
\end{definition}
\begin{remark}
This is a basic working definition by \citet{Schorlemmer2005}. In the case of \emph{reference ontologies}, \citet{Kalfoglou2} append this definition with i) a finite set $R$ of relations, and ii) a function $\sigma: R \lra C^{+}$ assigning to each relation its \emph{arity}. This corresponds to the functor $(-)^+$ which sends a set $C$ to the set of finite tuples whose elements are in $C$ (see the example below).
\end{remark}
In applications, the concepts in $C$ typically characterize concrete objects in the domain, which are brought into the theory by populating $\mathcal{O}$ with tokens. Let $X$ be a set of objects to be classified in terms of the concept symbols in $C$, via a classification relation $\Vdash_{\A}$; we define a classification $\A = \langle X, C, \Vdash_{\A} \rangle$, where $X = {\rm{Tok}}(\A)$, and $C = {\rm{Typ}}(\A)$.  The relation $\Vdash_{\A}$ will have to be defined so that $\leq$,~$\perp$, and $|$, are respected. This requirements lead to:

\begin{definition}
A \emph{populated ontology} is a tuple $\wti{\mathcal{O}} = (\A, \leq, \perp, |)$ such that $\A = \langle X, C, \Vdash_{\A}\rangle$ is an information flow classification, ${\mathcal{O}} = (\A, \leq, \perp, |)$ is an ontology, and for all $x \in X$, and $c, d \in C$, we have:
\begin{itemize}
\item[(1)] if $x \Vdash_{\A} c$, and $c \leq d$, then $x \Vdash_{\A} d$;

\item[(2)] if $x \Vdash_{\A} c$, and $c \perp d$, then $x \nVdash_{\A} d$;

\item[(3)] if $c | d$, then $x \Vdash_{\A} c$, or $x \Vdash_{\A} d$.
\end{itemize}
\end{definition}
A populated ontology $\wti{\mathcal{O}} = (\A, \leq, \perp, |)$ having $\A = \langle X, C, \Vdash_{\A} \rangle$, determines a local logic $\mathfrak{L} = (\A, \vdash) $, whose theory $(C, \vdash)$, is given by the smallest regular theory (i.e. the smallest theory closed under Identity, Weakening, and Global Cut), such that for all $c, d \in C$, we have:
\begin{equation}
\begin{aligned}
c \vdash d &\Leftrightarrow c \leq d \\
c, d \vdash &\Leftrightarrow c \perp d \\
\vdash c, d &\Leftrightarrow c | d
\end{aligned}
\end{equation}

\begin{example}\label{building}
To get an idea of what these last relations mean, take the case of a reference ontology $\mathcal{O} = (C, R, \leq, \perp, |, \sigma)$ as in \citet[\S 4]{Kalfoglou2}, with a set of concepts $C =\{{\mathsf{building,bird,starling}}\}$, the relation $R=\{\mathsf{isHavenFor} \}$, arities $\sigma(\mathsf{is HavenFor}) = \langle\mathsf{building,bird} \rangle$, where the partial order $\leq$, disjointness $\perp$, and coverage $\vert$, are defined by the following lattice:
\begin{equation}
\xymatrix@C=4pc{& \blacklozenge \ar@{-}[dl]  \ar@{-}[dr] &\\
\mathsf{building}\ar@{-}[ddr] &    & \mathsf{bird} \ar@{-}[d] &\\
 & & \mathsf{starling} \ar@{-}[dl] &\\
 & \lozenge  &
}
\end{equation}
where $\blacklozenge$ is the top and $\lozenge$ is the bottom of the lattice, i.e. $\mathsf{building} \perp \mathsf{bird}$ and $\mathsf{building} \vert \mathsf{bird}$. In this set up, we then have
\begin{equation}
\begin{aligned}
{\mathsf{building,bird}} & \vdash \\
{\mathsf{starling}} & \vdash {\mathsf{bird}}\\
            &\vdash {\mathsf{building,bird}}
\end{aligned}
\end{equation}
where the comma on the left-hand side has conjunctive force, whereas on the right-hand side it has disjunctive force. Thus, with respect to set of concepts $C$, the above constraints declare, respectively: nothing is both a building and a bird, all starlings are birds, and everything is either a building or a bird.

With respect to the theory in question, the corresponding sequents are:
$$
\begin{aligned}
& \langle \{\mathsf{bird,starling} \}, \{ \mathsf{building}\}  \rangle \\
&\langle  \{\mathsf{building}\},  \{\mathsf{bird,starling} \} \rangle \\
&\langle  \{\mathsf{bird}\},  \{\mathsf{building,starling} \} \rangle
\end{aligned}
$$

Then we have a classification in terms of the above sequents, as given by:

\bign

\begin{tabular}{ l | r r r }
$\Vdash_{\A}$  &$\mathsf{building}$  &$\mathsf{bird}$ & $\mathsf{starling}$  \\ \hline
$\langle \{\mathsf{bird,starling} \}, \{ \mathsf{building}\}  \rangle$  & 0 &1 &1 \\
$\langle \{\mathsf{building}\},  \{\mathsf{bird,starling} \}  \rangle $  &1 &0 & 0 \\
 $ \langle \{\mathsf{bird}\},  \{\mathsf{building,starling} \}  \rangle$  &0 & 1  & 0
   \\
\end{tabular}

\bign
Let the above set of sequents be denoted by $X$.
Note how the sequents code the classification $\A = \langle X, C, \Vdash_{\A} \rangle$ whereby the left-hand sides of these indicate which columns contain a `1' entries, and the right-hand sides indicate which columns contain `0' entries.
Assuming that $X$ consists of normal tokens, as in \citet{Kalfoglou2}, we obtain a local logic $\mathcal{L} = (\A, \vdash)$ of the ontology $\mathcal{O}$.
Given that $\mathcal{L}$ is a local logic on $\A$, we have by \citet[Prop. 12.7]{Barwise1}, that $\mathcal{L} = {\Lg}(\A)$; that is, with regards to Definition \ref{local-2}, $\mathcal{L}$ is the local logic generated by the classification $\A$.
Associated to $\mathcal{O}$ is a local, populated ontology, as shown in \citet[\S 4]{Kalfoglou2}, to which we refer for details.
\end{example}

\begin{example}
In order to formalize semantic integration of a collection of agents in Channel Theory, \citet{Kalfoglou1} propose: i) modeling populated ontologies of agents by classifications; ii) defining the channel, its core, and infomorphisms between classifications; iii) defining a logic on the core of the channel; and, iv) distributing the logic to the sum of agent classifications to obtain the required theory for semantic interoperability within the channel.
These steps give rise to a \emph{global ontology} for two candidate agents, $A_1$ and $A_2$, requiring semantic integration. This commences with a distributed logic of a channel $\mathcal{C}$ connecting the classifications $\mathbf{A_1}$ and $\mathbf{A_2}$ that model the agents' populated ontologies
$\wti{\mathcal{O}}_1$ and $\wti{\mathcal{O}}_2$, respectively:
\begin{equation}
\xymatrix{& \mathbf{C} & \\
\mathbf{A}_1 \ar[ur]^{f_1} \ar[rr]_{f} &  & \mathbf{A}_2 \ar[ul]_{f_2}
}
\end{equation}
\noindent
At the core of the channel $\mathcal{C}$, ${\rm{Typ}}(\mathbf{C})$ covers ${\rm{Typ}}(\mathbf{A}_1)$ and ${\rm{Typ}}(\mathbf{A}_2)$, while
elements of ${\rm{Tok}}(\mathbf{C})$ connect tokens from ${\rm{Tok}}(\mathbf{A}_1)$ with those from ${\rm{Tok}}(\mathbf{A}_2)$. Effectively, the global ontology comes about when the logic on the core of the channel is distributed to the sum of classifications $\mathbf{A}_1 + \mathbf{A}_2$, for the total semantic integration of the combined events.

The structure of a typical ontology mapping may thus be seen as follows \cite{Kalfoglou2}:
\begin{equation}
\xymatrix@C=4pc{& \mathcal{O}_{\rm{r}} \ar[dl] \ar[dr] &\\
\mathcal{O}_{\rm{loc}_1} \ar@{.>}[dr]  &   & \mathcal{O}_{\rm{loc}_2} \ar@{.>}[dl] & \\
& \mathcal{O}_{\rm{glob}}
}
\end{equation}
where $\mathcal{O}_{\rm{r}}$ is a reference ontology, $\mathcal{O}_{\rm{loc}_1}, \mathcal{O}_{\rm{loc}_2}$ are local ontologies, and $\mathcal{O}_{\rm{glob}}$ is a global ontology.

\end{example}

\subsection{Quotient channel}

Given an invariant $I = \langle \Sigma, R \rangle$ on a classification $\A$, the \emph{quotient channel of $\A$ by $I$} is the limit of the distributed system depicted by
\begin{equation}
\A \ovsetll{\tau_I} \A/{I} \ovsetl{\tau_I} \A.
\end{equation}
As a refinement of any other such channel, the quotient channel makes the following diagram commute
\begin{equation}
\xymatrix{& \mathbf{C} &  \\
\A \ar[ur]^{g_1} & \ar[l]_{\tau_I} \A/{I} \ar[u]^{h} \ar[r]^{\tau_I} & \A \ar[ul]_{g_2}
}
\end{equation}

\begin{remark}
An example is a hierarchially modular chain, where at each level of abstraction, the tokens can be inherited, and the resulting infomorphisms are created by systematically composing those from the levels below. As seen in \citet{Franklin1} or \citet{Friedlander}, the perceptual memory relationships and actions of a typical LIDA semantic network architecture appear to fit into this pattern.
\end{remark}

\begin{remark}
Ideas such as information flow, formal concepts, conceptual spaces, and local logics can be categorically unified when they are embraced within the abstract axiomatization of an \emph{Institution} \citep{Goguen1,Goguen2}. This consists of a functor from an abstract category of `Signatures' to a category of classifications that involves `contexts' linked
via the `satisfaction relation' ($\Vdash$). As pointed out in \citet{Kent}, information flow is a particular case of FOL (which is thus one Institution), but the classification relation between Tokens and Types abstracts the Institution satisfaction relation between structures and sentences.
\end{remark}

\subsection{State spaces and projections}\label{states-1}

Recall that a state space is a classification $\sfS$ for which each token is of exactly one type, and where the types of the space are simply
the states themselves. Here $a$ is said to be \emph{in state $\sigma$} if $a \Vdash_{\sfS} \sigma$. The space $\sfS$ is \emph{complete} if every state is the state of some token.

\begin{definition}
A \emph{projection} $f: \sfS_1 \rightrightarrows \sfS_2$ from a state space $\sfS_1$, to a state space $\sfS_2$ is given by a covariant pair of functions, such that for each token $a \in \rm{Tok}(\sfS_1)$, we have $f({\rm{state}}_{\sfS_1}(a)) = {\rm{state}}_{\sfS_2}(f(a))$. This amounts to the commutativity of the following diagram:
\begin{equation}\label{proj-1}
\xymatrix@!C=3pc{{\rm{Typ}}(\sfS_1) \ar[r]^{f}   & {\rm{Typ}}(\sfS_2) \\
{\rm{Tok}}(\sfS_1)  \ar[u]^{{\rm{state}}_{\sfS_1}} \ar[r]_{f} & {\rm{Tok}}(\sfS_2) \ar[u]_{{\rm{state}}_{\sfS_2}}}
\end{equation}
\end{definition}
The composition of projections is well-defined, and so too is the Cartesian product $\Pi_{i \in \mathcal{I}}\sfS_i$ of indexed
state spaces with natural projections
\begin{equation}
\pi_{\sfS_i} : \Pi_{i \in \mathcal{I}}\sfS_i \rightrightarrows \sfS_i,
\end{equation}
(see \cite[\S8.2]{Barwise1}).

\subsection{The Event Classification}\label{event-class}

Following the development of ideas in \citet[Ch. 9]{Barwise1}, studies such as \citet{Kakuda1} define the \emph{Event Classification $\Evt(\sfS)$} associated with a state
space $\sfS$ as follows:
\begin{itemize}
\item[(1)] $\rm{Tok}(\Evt(\sfS)) = \rm{Tok}(\sfS)$;

\item[(2)] $\rm{Typ}(\Evt(\sfS)) = \cP(\rm{Typ}(\sfS))$;

\item[(3)] $s \Vdash_{\Evt(\sfS)} \a$ is defined by $\mathsf{state}_s (s) \in \a$, for $s \in \rm{Tok}(\Evt(\sfS))$ and
$\a \in \rm{Typ}(\Evt(\sfS))$;
\end{itemize}
where as before, $\cP(\cdot)$ indicates the power set.

Briefly recapping, this says that the space of events $\rm{Evt}(\sfS)$ associated to $\sfS$ has as its tokens the tokens of $\sfS$ and its types are arbitrary sets of sets of states of $\sfS$. The classification relation $s \Vdash_{\rm{Evt}(\sfS)} \a$ above is equivalent to ${\rm{state}}_{\sfS}(s) \in \a$. Following \citet[Prop. 8.17]{Barwise1}, given state spaces $\sfS_1$ and $\sfS_2$, the following are equivalent:
\begin{itemize}
\item[(1)] $f: \sfS_1 \rightrightarrows \sfS_2$ is a projection;

\item[(2)] ${\rm{Evt}}(f): {\rm{Evt}}(\sfS_2) \rightleftarrows {\rm{Evt}}(\sfS_1)$ is an infomorphism.
\end{itemize}
In fact, for any state space $\sfS$, the classification $\rm{Evt}(\sfS)$ is a Boolean classification in which the operations of taking intersection,
union, and complement are here conjunction, disjunction, and negation, respectively \cite[Prop. 8.18]{Barwise1}.
\begin{definition}\label{local-3}
Let $\sfS$ be a state space. The \emph{local logic generated by $\sfS$}, denoted ${\Lg}(\sfS)$, has classification $\rm{Evt}(\sfS)$ , regular theory ${\Th}(\sfS)$, and all of its tokens are normal.
\end{definition}
For further relationships see \citet[12.1 - 12.2]{Barwise1}.

\begin{remark}
Taking $\sfK = [0,1]$, the evaluation relation
\begin{equation}
\Vdash_{\Evt(\sfS)}: \rm{Tok}(\Evt(\sfS)) \times \rm{Typ}(\Evt(\sfS)) \lra [0,1]
\end{equation}
together with (logic) infomorphisms between event classifications, may be compared with the concept of a \emph{perceptual strategy} as described in \citet{Hoffman1}.
\end{remark}

\subsection{State space systems}\label{states-2}

Considering the above ingredients we now seek a unifying principle that characterizes the state space model
and provides a suitable information channel. This motivates starting with:
\begin{definition}
A \emph{state-space system} consists of an indexed family $\mathcal{S} = \{f_i: \sfS \rightrightarrows \sfS_i\}_{i \in \mathcal{I}}$
of state-space projections with a common domain $\sfS$, called the \emph{core} of $\mathcal{S}$, to state spaces $\sfS_i$ (for $i \in \mathcal{I}$);
$\sfS_i$ is called the $i$th component space of $\mathcal{S}$.
\end{definition}
We now consider an ``event'' $\rm{Evt}$ as a functor that transforms a state-space system into an information channel. Taking a pair of state spaces as an example, we first take projections:
\begin{equation}
\xymatrix{& \sfS \ar[dl]_{f_i} \ar[dr]^{f_j} &\\
\sfS_i &   & \sfS_j
}
\end{equation}
Next, applying the functor $\rm{Evt}$ to this diagram yields a family of infomorphisms with a commom domain $\rm{Evt}(\sfS)$, yielding an information channel:
\begin{equation}
\xymatrix{& \rm{Evt}(\sfS) &  \\
{\rm{Evt}}(\sfS_i) \ar[ur]^{{\rm{Evt}}(f_i)} &   & {\rm{Evt}}(\sfS_j) \ar[ul]_{{\rm{\Evt}}(f_j)}
}
\end{equation}
State space addition produces a further commuting diagram, where for ease of notation, we write $\sigma_i$ for
$\sigma_{{\rm{Evt}}(\sfS_i)}$, and simply $f$ for $\sum_{k \in \mathcal{I}} {\rm{Evt}}(f_k)$:
\begin{equation}
\xymatrix@C=4pc{& \rm{Evt}(\sfS) &  \\
\rm{Evt}(\sfS_i) \ar[ur]^{{\rm{Evt}}(f_i)} \ar[r]_{\sigma_i} & \sum_{k \in \mathcal{I}} \rm{Evt}(\sfS_k) \ar[u]^{f} & \ar[l]^{\sigma_j}  \rm{Evt}(\sfS_j) \ar[ul]_{{\rm{Evt}}(f_j)}
}
\end{equation}

\begin{example}

Following \citet[\S4]{Kakuda1}, let $T$ be a regular theory, and let $\sfS$ be a state space. A \emph{medium system} denoted $D := \langle
\sfD, \sfN, f, p \rangle$ between $T$ and $\sfS$, consists of the following:
\begin{itemize}
\item[(1)] a state space $\sfD$,

\item[(2)] a subset $\sfN$ of $\rm{Tok}(\mathnormal{D})$,

\item[(3)] an infomorphism $f: \Cl(T_{\rm{Typ}(\mathnormal{T})}) \rightleftarrows \Evt(\sfD)$, and

\item[(4)] a projection $p: \sfD \rightrightarrows \sfS$,
\end{itemize}
where $\langle \rm{Tok}(\sfD), \rm{Typ}(\mathnormal{T}), \vdash_{T}, \sfN, \rm{Typ}(\mathnormal{D}), \mathsf{state}_{D}, \overrightarrow{f}, \overleftarrow{f} \rangle$
forms a functional scheme. Here $\sfD$ is called the \emph{medium space of} $D$.  \citet[\S4]{Kakuda1} define an information channel:
\begin{equation}
\xymatrix{& \Evt(\sfD) &  \\
\Cl(T_{\rm{Typ}(T)}) \ar[ur]^{f} &   & \Evt(\sfS) \ar[ul]_{\Evt(p)}
}
\end{equation}
through $\Evt(\sfD)$ to $\Evt(\sfS)$.
\end{example}

\begin{example}\label{cognizance}

\citet{SS1,SS2} employ the concept of the core of a binary channel, when realized as a classification, in order to describe an agent's cognition (cf \cite{Barwise2}).
Consider separate source ($\A$) and target ($\B$) classifications, and represent the agent's knowledge by a regular theory $T = \langle \Sigma, \vdash \rangle$. The idea is to construct a set of
\emph{possible and realizable states} in several steps:
\begin{itemize}
\item[(1)] For source classification $\A$, and target classification $\B$, let $\Omega_{\langle \A, \B \rangle}$ denote the set
of all partitions of $\rm{Typ}(\A) \cup \rm{Typ}(\B)$, called \emph{the set of states generated by $\A$ and $\B$}.

\item[(2)] The set of \emph{realizable states generated by $\A$ and $\B$} is given by
\begin{equation}
\Omega^R_{\langle \A, \B \rangle} :=\{ \langle \Theta, \Lambda \rangle \in \Omega_{\langle \A, \B \rangle}: \exists a \in \A, ~ \rm{Typ}(a) \subseteq \Theta~ \text{and},~ \rm{Typ}^c(a) \subseteq \Lambda \},
\end{equation}
where the notation $\rm{Typ}^c(a)$ indicates the complement, i.e. everything not in $\rm{Typ}(a)$.
\item[(3)] The set of \emph{impossible states under the theory $T$} is given by
\begin{equation}
\Omega^{IP}_{\langle \A, \B\vert T \rangle} :=\{ \langle \Theta, \Lambda \rangle \in \Omega_{\langle \A, \B \rangle}:
\Theta \vdash_T \Lambda \}.
\end{equation}
The ``impossibility'' here is that $\Omega$ constrains a $\Lambda$ with which it is disjoint.
\item[(4)]
\emph{The possible states under the theory $T$} is then
\begin{equation}
\Omega^{P}_{\langle \A, \B\vert T \rangle} = \Omega_{\langle \A, \B \rangle} \backslash \Omega^{IP}_{\langle \A, \B\vert T \rangle}.
\end{equation}

\item[(5)]
\emph{The possible and realizable states under the theory $T$} is thus
\begin{equation}
\Omega^{PR}_{\langle \A, \B\vert T \rangle} = \Omega^P_{\langle \A, \B \vert T \rangle} \cap \Omega^{R}_{\langle \A, \B \rangle}.
\end{equation}
\end{itemize}
The \emph{cognizance classification $\mathfrak{C}_{\langle \A, \B, T \rangle}$} generated by $\A, \B$ and $T$ is then given by
\begin{equation}
\mathfrak{C}_{\langle \A, \B, T \rangle} := \big\langle \Omega^{PR}_{\langle \A, \B\vert T \rangle}, \rm{Typ}(\A) \cup \rm{Typ}(\B), \Vdash_{\mathfrak{C}_{\langle \A, \B, T \rangle}} \big\rangle
\end{equation}
where the relation $\Vdash_{\mathfrak{C}_{\langle \A, \B, T \rangle}}$ is defined as $\langle \Theta, \Lambda \rangle \Vdash_{\mathfrak{C}_{\langle \A, \B, T \rangle}} \a$ if and only if $\a \in \Theta$, i.e. the choice of $\Lambda$ can be arbitrary provided it is disjoint from $\Theta$.

\end{example}

\subsection{On comparing and combining the Shannon Theory of Information with Channel Theory}\label{shannon}

\citet{Barwise2} recalls \emph{Shannon's Inverse Relation} between possibilities and information, basically saying that eliminating possibilities from consideration amounts to increasing one's information and vice-versa. That relationship is fundamental to Dretske's original goal of developing a semantic theory of information based on possibilities \citep{Dretske1,Dretske2}. Though very general as a quantitative theory of communication flow, the original Shannon theory had largely overlooked the question of semantic content. In any Dretske-type theory, the basis of semantic content is in the world, i.e. in the events or situations that signals or states carry information about.  By showing how local logics are connected by information networks, channel theory provides a general qualitative theory of information flow in this context. `Channels' in the theory are more general than the traditional idea of the Shannon communication channels \citep{Cover2006}.  In Shannon's theory, information flow in a channel is defined in terms of reduction of uncertainty about the \emph{type} of event that will occur; it says nothing about the semantics of any specific bit, i.e. about any specific token.  Channel theory specifically concerns particular tokens $x$ and statements of the form ``$x$ is an $A$.''  \citet{Allwein2004,Allwein1} create a synthesis of Shannon's quantitative theory with the Barwise-Seligman qualitative theory to address the question of how specific objects, situations and events carry information about each other.

A classification of possible outcomes of events starts with a probability space $\mathcal{P} = \langle \Omega, \Sigma, \mu \rangle$, where $\Omega$ denotes the set of possible outcomes, $\Sigma$ is a $\sigma$-algebra on $\Omega$ whose members represent events, and $\mu$ is a probability measure on $\sigma$ representing the probability of an event having or being associated with a particular outcome.\footnote{Recall that a \emph{$\sigma$-algebra} over $\Omega$ is a set $\Sigma$ of subsets of $\Omega$, such that $\emptyset \in \Sigma, ~\Omega - e \in \Sigma$, for each $e \in \Sigma$, and $\bigcup E \in \Sigma$, for each countable set $E \subseteq \Sigma$. $\mu$ is a \emph{probability measure} on $\Sigma$, if and only if it satisfies the Kolmogorov axioms: $\mu(\emptyset) = 0,~ \mu(\Omega -e) = 1 - \mu(e)$, and $\mu(\bigcup E) = \sum_{e \in E} \mu(e)$ if $E$ is countable, and $\mu(e_1 \cap e_2) = 0$, for all $e_1 \neq e_2 \in E$.}  We define a classification $\rm{Tok}(\mathcal{P}) = \Omega$,
$\rm{Typ}(\mathcal{P}) = \Sigma$, and let $\omega \Vdash_{\mathcal{P}} e$, if and only if $\omega \in e$. In this context, an infomorphism between probability spaces is topologically a continuous map \citep{Seligman1}.

To see how the basic ontology of the Shannon theory can be conveniently embedded in that of Channel theory, note that the former's basic unit of information is some tuple of a binary relation. The relation is restricted to be of the form $x \Vdash V$, where $\Vdash$ is regarded as a function, and $V$ is the value of the Token $x$. But closer inspection reveals this characterizes a state space in which $V$ is a state and tokens are ignored. In Channel Theory there are states, but tokens are not ignored and types are values, as in \S\ref{classifications-1}. States may be amalgamated to form Events as in \S\ref{event-class} and \S\ref{states-2}. Channel theory permits this by firstly preserving the tokens, and then replacing states with types, whose events are also types: for some event $E$, $x \Vdash E$, just when $x \Vdash s$, for some state $s \in E$ \citep{Allwein1}. This is basically the embedding of the one theory into the other, and the two theories together admit a certain generalization as follows.

The presence of a sequent in an information channel (as outlined in \S\ref{local}) effectively represents a logical gate, and this kind of structure can be seen as more general than a Markov structure (cf. the Kolmogorov axioms in \citet{Allwein2004,Allwein1}), since sequents enable the information flow to simultaneously support a flow of reasoning.  Specifically, probabilities can be assigned to sequents in $\A$ as follows. Suppose we have:
\begin{equation}\label{prob-1}
M \Vdash_{\A} N \qquad \forall x (x \Vdash_{\A} M \Rightarrow x \Vdash_{\A} N),
\end{equation}
then the sequent relation $\vdash$ can be weakened by removing $\forall$, and instead stating that for any $x$, we have a probability $x \Vdash N$, given $x \Vdash M$; that is, the probability that $x$ satisfies $N$ given it satisfies $M$. This is clearly a conditional probability, so one defines:
\begin{equation}\label{prob-2}
M \Vdash_{\A}^{\mathbf{P}} N := \mathbf{P}( M \vert N).
\end{equation}
Thus when the sequent's conditional probability is $p$, say, we have $M \Vdash_{\A}^{p} N$. \textit{A priori}, one must have $x \Vdash_{\A} M$ in order to apply $M \Vdash_{\A} N$ in a argument. The probability of the former holding in $\A$, is $\mathbf{P}(M)$. Then $x \Vdash_{\A} N$ follows from the rule
$\mathbf{P}(M) \cdot  M \Vdash_{\A}^{\mathbf{P}} N$. Probability axioms for a Countable Classical Propositional Logic are developed in \cite{Allwein2004} (cf. \citet{Allwein1}) to which we refer for details. Note that information flow in distributed systems can be interpreted dynamically; this amounts to causation in an informational context, consistent with the Dretskean nature of the theory. In this respect, the relations between information theory and logic are also conducive to understanding certain relations between causation and computation \citep{Collier,Seligman1}.

%%%%%%%%%%%%%%%%%%%%%%%%%%%%%%%%%%%%%%%%%%%%%%%%%%%%%%%%%%%%%%%%%%%%%%%%%%%%%%%%%%%%%%%%%%%%%%%%%%%%%%%%%%%%%%%%%%%%%%%%%%%%%%%%%%%%

\section{Colimits for piecing together local-to-distributed information: Application to information flow}\label{colimits}

\subsection{Cocones and Colimits}\label{cocone-1}

To a category theorist, minimal covers as described in \S\ref{distributed} are familiar as \emph{colimits}, where the channel core $\mathbf{C}$ represents the vertex of a cocone (as was introduced in \S\ref{info-cocone}; more formally, see e.g. \citet[Ch. 5]{Awodey}) and the existence of colimits of Chu spaces (realized here by the classifications) has been verified \citep{Barr1}. Let us briefly recall this concept with some graphic intuition behind the idea following \citet{Baianu2006,Brown2003}.

The ``input data'' for a colimit is a \emph{diagram} $\mathbf{D}$, i.e. a collection of some objects in a category $\mathfrak{C}$,
together with some
arrows between them, as depicted by:
\begin{equation}\label{colim-1}
\xymatrix@R=3pc{ & . \ar[rr]&&.\ar[dl]&\\ **[l] \mathbf{D} = \qquad \cdot \ar[ur]\ar[dr]& & .
\ar[ul]\ar[rr]\ar[dl] & &.\ar[ul]\\ &.&&&}
\end{equation}
This generalizes our use of a directed graph in \S\ref{info-cocone} by allowing the ``vertices'' and ``edges'' of $\mathbf{D}$ to be objects and morphisms of an arbitrary category.  Next we need `functional controls' comprising a \emph{cocone with base $\mathbf{D}$ and vertex an
object $\mathbf{C}$ in $\mathfrak{C}$},

\begin{equation}\label{colim-2}
\xymatrixcolsep{3pc} \xymatrixrowsep{2pc}\xymatrix{
&&\mathbf{C}&&\\&&&&\\ &&&&\\
& . \ar@{-}[r]|(0.35)\hole\ar[uuur]&\ar[r]&.\ar[dl]\ar[uuul]&\\
\mathbf{D} \qquad.\ar[ur]\ar[uuuurr]\ar[dr]& & . \ar[ul]|!{[dl];[uuuu]}\hole\ar[rr]\ar[dl]\ar[uuuu]
& &.\ar[ul]\ar[uuuull]\\
&.\ar[uuuuur]&&&}
\end{equation}
\noindent
such that each of the triangular faces of the cocone is commutative. The ``output'' from
$\mathbf{D}$ will be an object $\mathsf{colim}(\mathbf{D})$ in our category $\mathfrak{C}$ defined by a special \emph{colimit cocone}, such that any cocone on $\mathbf{D}$ factors \emph{uniquely} through
the colimit cocone. Effectively, the commutativity condition on the cocone induces, in the
colimit, an interaction of images from different parts of the diagram $\mathbf{D}$. The
uniqueness condition makes the colimit the optimal solution to this factorisation problem.

Let us set
\begin{equation}
\blacklozenge= \mathsf{colim}(\mathbf{D})
\end{equation}
where the dotted arrows in the diagram below represent new morphisms which
combine to make the colimit cocone:

\begin{equation}\label{colim-3}
\xymatrixcolsep{3pc} \xymatrixrowsep{2pc}\xymatrix{ &&\mathbf{C}&&\\&&&&\\
**[l]\mathsf{colim}(\mathbf{D})=\blacklozenge\ar@{-->}[rruu]^
\Phi &&&&\\
& .\ar@/_/@{.>}[lu] \ar@{-}[r]|(0.35)\hole\ar[uuur]&\ar[r]&.\ar@{.>}[lllu]\ar[dl]\ar[uuul]&\\
**[l]\mathbf{D} \qquad \cdot  \ar@{.>}[uu] \ar[ur]\ar[uuuurr]\ar[dr]& & .
\ar@/^/@{.>}[lluu] \ar[ul]|!{[dl];[uuuu]}\hole \ar[rr]\ar[dl]\ar[uuuu]
& &.\ar[ul]\ar@{.>}[uullll]\ar [uuuull]\\
&.\ar[uuuuur] \ar@{.>}[uuul]&&&}
\end{equation}
and for which the broken arrow $\Phi$ is constructed by requiring commutativity for all of the
triangular faces of the combined diagram. Next, stripping away the `old'
cocone results in a factorisation of the cocone via the colimit:
\begin{equation}\label{colim-4}
\xymatrixcolsep{3pc} \xymatrixrowsep{2pc}\xymatrix{ &&\mathbf{C}&&\\&&&&\\
**[l]\mathsf{colim}(\mathbf{D})=\blacklozenge
{\ar@{-->}[rruu]^\Phi} &&&&\\
& .\ar@/_/@{.>}[lu] \ar@{-}[r]&\ar[r]&.\ar@{.>}[lllu]\ar[dl]&\\
**[l]\mathbf{D} \qquad \cdot  \ar@{.>}[uu] \ar[ur]\ar[dr]& & .
\ar@/^/@{.>}[lluu] \ar[ul]\ar[rr]\ar[dl]
& &.\ar[ul]\ar@{.>}[uullll]\\
&. \ar@{.>}[uuul]&&&}
\end{equation}
Intuitively, the process can be seen as follows. The object $\mathsf{colim}(\mathbf{D})$ is pieced together from the
diagram $\mathbf{D}$ by means of the colimit cocone. From beyond $\mathbf{D}$, an arbitrary object $\mathbf{C}$ in $\mathfrak{C}$ `sees' $\mathbf{D}$ as mediated through its colimit. This means that
if $\mathbf{C}$ is going to interact with all of $\mathbf{D}$, then it does so via $\mathsf{colim}(\mathbf{D})$.
The colimit cocone can be thought of as a kind of program: given any cocone on $\mathbf{D}$ with
vertex $\mathbf{C}$, the output will be a morphism
\begin{equation}
\Phi: \mathsf{colim}(\mathbf{D})\to \mathbf{C}
\end{equation}
as constructed from the other data.
\footnote{The diagrams included in \eqref{colim-1}--\eqref{colim-4} are reproduced from \cite{Baianu2006,Brown2003}, with permission from R. Brown}

\begin{example}
\citet{Brown2003} provide an analogy comparing colimits with how an email message can be relayed. Suppose $E$ denotes some email document. This is to be sent via a server $S$, which decomposes $E$ into numerous parts $E_i$ ($i \in \mathcal{I}$, an indexing set), and labels each part $E_i$, so it becomes $E'_i$. These labelled parts $E'_i$ are then sent to a number of servers $S_i$, which then relay these messages as newly labelled messages $E''_i$ to a server $S_C$, for the receiver $C$. The server $S_C$ then combines the $E''_i$ to produce the recovered message $M_E$ at $C$. Breaking the message down and routing it through the $S_i$ appears arbitrary, but the system is designed such that $M_E$ is independent of all choices made at each step of the process. \end{example}
Many other illustrative examples applying colimits to computer science, social systems, and neuroscience, can be seen in \citet{Baianu2006,EV2007,Healy1,Healy2}.

\subsection{Coordinated channels in ontologies}\label{coordinated}

Suppose we have two prospectively interoperating agents $A_1,A_2$, with each agent $A_i$ ($i=1,2$) having its knowledge represented according to its own conceptualization, as specified in relationship to its ontology $\mathcal{O}_i$, respectively. This means that a concept of $\mathcal{O}_1$ will, \textit{a priori}, be considered semantically distinct from $\mathcal{O}_2$, even if they are equivalent syntactically.  However, the behavior of the agents can provide evidence for a meaning common to $A_1$ and $A_2$. Let us us assume that the agents' ontologies are not open to a third-party inspection. \citet{Kalfoglou1} use a channel to coordinate the populated ontologies (cf. \S\ref{ontologies}) $\wti{\mathcal{O}}_1, \wti{\mathcal{O}}_2$ by capturing the degree of participation of each agent in communicative behaviors.  Specifically, i) agent $A_i$ attempts to ``explain'' a subset of its concepts to other agents, and ii) other agents exchange with $A_i$ some of their own tokens, thus increasing the set of tokens originally available to $A_i$.

To see how this degree of participation can be captured by Channel Theory, \citet{Kalfoglou1} introduce classifications $\mathbf{A}_i =
 \langle \rm{Tok}(\mathbf{A}_i), \rm{Typ}(\mathbf{A}_i), \Vdash_{\mathbf{A}_i} \rangle$, corresponding to the agents $A_i$, respectively, along with subclassifications $\mathbf{A}'_i =
 \langle \rm{Tok}(\mathbf{A}'_i), \rm{Typ}(\mathbf{A}'_i), \Vdash_{\mathbf{A}'_i} \rangle$, and infomorphisms
 $g_i: \mathbf{A}'_i \lra \mathbf{A}_i$, for which functions $\hat{g}_i$ and $\check{g}_i$ are the inclusions $\rm{Typ}(\mathbf{A}'_i) \subseteq
 \rm{Typ}(\mathbf{A}_i)$ and $\rm{Tok}(\mathbf{A}'_i) \subseteq \rm{Tok}(\mathbf{A}_i)$, respectively. It is from the subclassifications $\mathbf{A}'_i$ arising from the interactions that coordination is established. Thus we have the following information channel with core (i.e. cocone) $\mathbf{C}'$:
\begin{equation}
\xymatrix{& &\mathbf{C}' &  \\
\mathbf{A}_1 & \ar[l]^{g_1} \mathbf{A}'_1\ar[ur]^{f_1} &  & \mathbf{A}'_2\ar[ul]_{f_2} \ar[r]_{g_2} & \mathbf{A}_2
}
\end{equation}
The optimal coordinated channel that captures semantic integration achieved by the agents is then represented by the colimit $\mathcal{C}' = \mathsf{colim}\{\mathbf{A}'_1 \leftarrow \mathbf{S} \rightarrow \mathbf{A}'_2 \}$ of the diagram linking the subclassifications that model the agents' participation in the interoperation:
\begin{equation}
\xymatrix{& &\mathbf{C}' &  \\
\mathbf{A}_1 & \ar[l]^{g_1} \mathbf{A}'_1\ar[ur]^{f_1} & \ar[l]^{h_1} \mathbf{S}  \ar[r]_{h_2} & \mathbf{A}'_2\ar[ul]_{f_2} \ar[r]_{g_2} & \mathbf{A}_2
}
\end{equation}
\noindent
``Optimality'' here means that every other channel induces a map to $\mathcal{C}'$ when commutativity is required.

%%%%%%%%%%%%%%%%%%%%%%%%%%%%%%%%%%%%%%%%%%%%%%%%%%%%%%%%%%%%%%%%%%%%%%%%%%%%%%%%%%%%%%%%%%%%%%%%%%%%%

\section*{Part II: Applications to Perception and Cognition}\label{perception-cognition}

\section{Perception, categorization and attention as neurocognitive processes}\label{cognitive}

We now turn from the assembly of category-theoretic tools as presented in Part I, to the application of these tools to modelling object recognition and categorization, particularly the recognition and categorization of mereologically-complex entities.  We first review the basic cognitive neuroscience of object recognition and categorization in vision, currently the best understood of the sensory channels.  We focus on the data structures employed -- what \citet{Marr} called the algorithmic/representational level of analysis -- with some pointers to the relevant implementation-level neuroscience (see \citet{Fields13} for a review of implementation details).

\subsection{Dual-process vision and object files}\label{dual-process}

The primate, and in particular human, visual system comprises two early processing pathways, a dorsal pathway specialized for the rapid processing of location and motion information, and a ventral pathway specialized for static (e.g. shape, size, texture and color) feature information (reviewed by \citet{goodale92,scholl08,Fields11}; see \citet{cloutman} for a discussion of interactions between these pathways and \citet{Alain01} and \citet{Sathian12} for evidence that auditory and haptic perception, respectively, have a similar dual-stream organization).  Perception of a located, featured object requires processing by both pathways followed by fusion of the intermediate partial representations they produce.

Studies of visual object tracking over short time periods consistently show that trajectory information dominates static feature information in determining object identity \citep{scholl08,Fields11}.  \citet{Kahneman} termed the initial, transient representation of a moving object in visual short-term memory the ``object file'' (see also \citet{Treisman06}).  As under ordinary circumstances all objects are effectively moving due to visual saccades, object files are at least typically initiated by dorsal-stream processing.  Static feature information extracted from the relevant part of the visual field by ventral-stream processing is then bound to this initially motion-based representation.  These processing steps require 50 -- 100 ms in humans, much shorter than the time required for reportable visual awareness of the object.

The object file is the fundamental ``token'' representation of a located, bounded, featured entity that is distinguished from the ``background'' of a visual scene.  It represents where the object is, its visually-identifiable features, and its instantaneous trajectory during the time window $\Delta t$ from object-file initiation to feature binding.  All further information about the object is added by downstream processing; in particular, whether the object is novel or something previously encountered, either as a type or as a specific individual, must be computed from information available in memory.

\subsection{Feature-category binding and object tokens}

Object files are implemented in a content-dependent way across the posterior temporal cortex \citep{Martin07,Mahon09,Fields13}; features of entities perceived as agents or non-agents, for example, are encoded in the lateral or medial, respectively, fusiform gyrus (Fig. 1).  This distributed, content-dependent encoding indicates that top-down category information, e.g. agent versus non-agent, is already active in the binding of location and motion information to feature information at the level of the object file.

\centerline{\includegraphics[width=18cm]{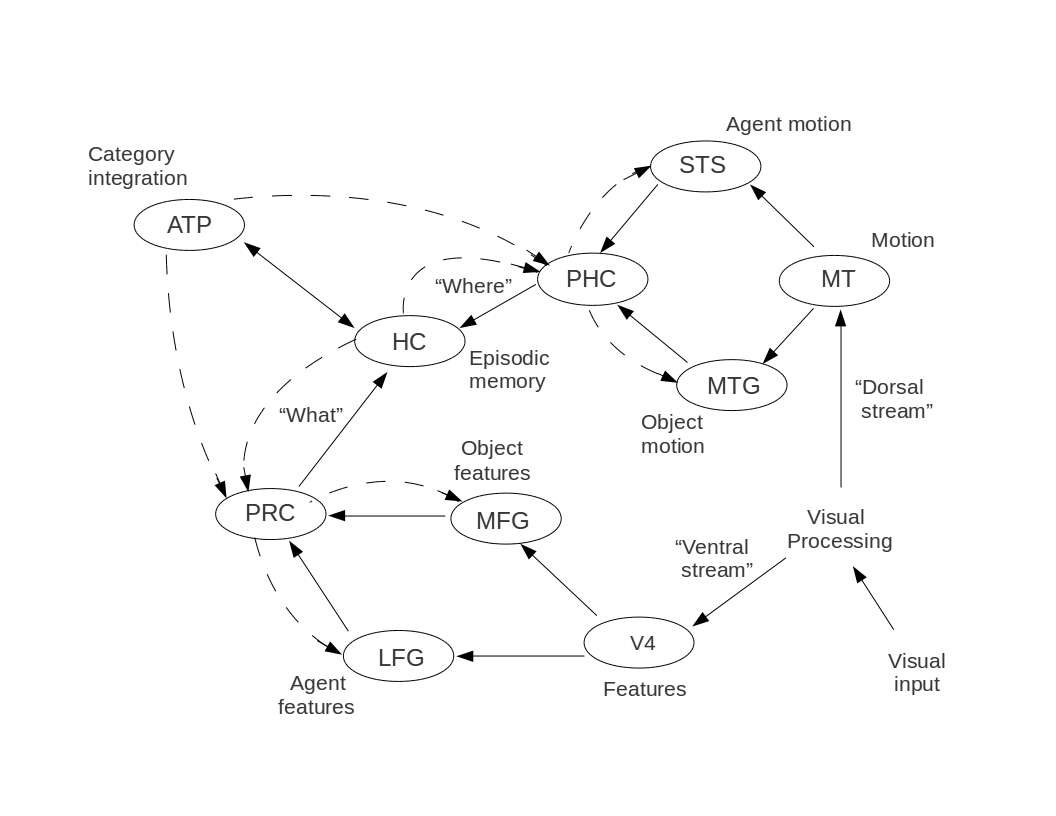}}
\begin{quote}
\textit{Fig. 1}: Simplified functional architecture of visual object perception within the temporal lobe.  Abbreviations are: MT, medial temporal area; STS, superior temporal sulcus; MTG, medial temporal gyrus; PHC, parahippocampal cortex; V4, visual area 4 (occipital cortex); MFG, medial fusiform gyrus; LFG, lateral fusiform gyrus; PRC, perirhinal cortex; HC, hippocampus; ATP, anterior temporal pole.  Solid lines are feedforward; dashed lines feedback.  Adapted from \citet{Fields13}.
\end{quote}

The functional architecture supporting object representation and object-directed attention is already present at birth (see e.g. \citet{Gao15,Huang15} for neuroarchitectural and \citet{Johnson15} for behavioral evidence) and its functionality rapidly matures toward adult levels during the first two years.  Visual feature identification, segregation of co-moving, conjoined objects, and the complementary process of grouping co-moving, non-conjoined objects, for example, are highly dependent on top-down, memory-driven categorization or, in Bayesian terms, expectation confirmation or disconfirmation.  Four- to six-month old human infants, for example, typically do not segregate static or co-moving conjoined objects that older infants, children or adults do segregate, but quickly learn to do so when the objects are separately manipulated \citep{Johnson15}.  Young infants similarly fail to group co-moving, non-conjoined objects (i.e. fail to perform ``object completion'') that older infants, children or adults do group, with the exception of point-light walkers exhibiting biological motion, which infants perceive as single entities from the earliest ages tested (\citet{Johnson15}; see \citet{Schlesinger12} for a replication of a canonical object completion experiment in the iCub robot).

Young infants exhibit robust object memory, particularly for familiar faces, and emotional responses to objects, again from the earliest ages tested.  Feelings of familiarity and their attendant emotions correlate with feature-based object recognition at the level of perirhinal cortex \citep{Eichenbaum07}.  Memory for a particular, re-identifiable object requires a memory-resident representation of the individual object, what \citet{Zimmer} have termed the ``object token.''  Recognizing a novel object as a distinct, individual thing involves encoding a new object token specifically for it.  Recognition or re-identification of the same individual object on a later occassion is then a process of matching the current object file to this previously-encoded object token (Fig. 2).  This process is, in general, not straightforward, as object features, behaviors, and locations may change between encounters; both feature matching and, after about four years of age, the construction of unobserved and hence fictive causal histories (FCHs) of objects are employed to establish individual object identity across observations \citep{Fields3}.  Enabling object recognition across feature, behavior, and context changes requires object tokens to have a ``core'' of essential properties that change only slowly through time.  The distinction between core and variable properties in object tokens is category-dependent and not well understood.

\centerline{\includegraphics[width=18cm]{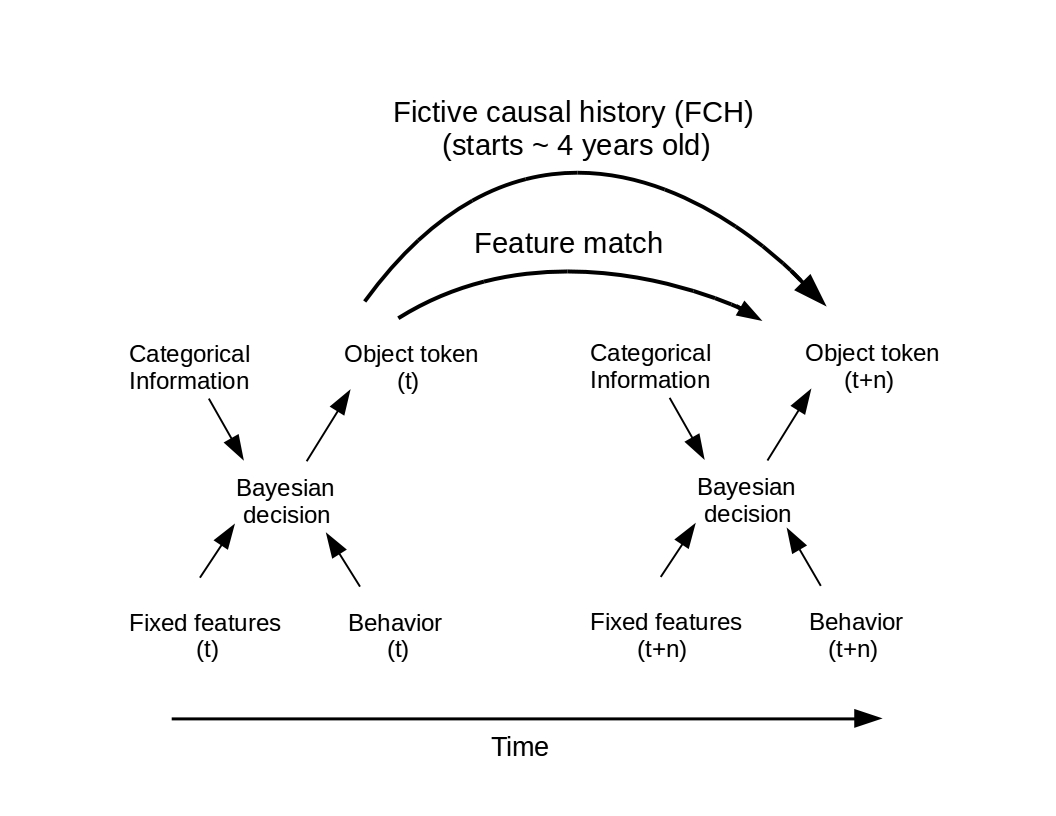}}
\begin{quote}
\textit{Fig. 2}: Identifying an object as the same individual across time requires matching a current object file to a memory-resident object token.  Both feature matching and the construction of unobserved (fictive) causal histories (FCHs) are employed to link object tokens across observations.
\end{quote}

\citet{Eichenbaum07} emphasize that categorization by type precedes object-token matching; e.g. an individual person is typically recognized as a person before they are identified as a particular individual person.  The feeling of familiarity is generated already at the level of type recognition; \citet{Eichenbaum07} consider the example of recognizing a person as a person, as a specific individual person encountered before, and only after some deliberation as a known, named, individual person represented by a memory-resident object token.  How the relationship between categories and object tokens as components of semantic memory is implemented at the neural circuit level, i.e. the details of the circuitry connecting ATP and PRC in Fig. 1, is not well understood.  While both categorization and individual recognition are generally assumed to be implemented by some form of Bayesian predictive coding \citep{Friston2,Maloney10}, how categorization constrains or guides object token matching in particular cases is also not well characterized.  The category - to - object token satisfaction relation $\Vdash$ remains, in other words, an empirical question for each particular type-token pair.

\subsection{Context perception, event files and episodic memories}\label{episodic}

Objects are invariably recognized in some context, typically one involving other recognized or at least categorized objects.  The spatial ``where'' information processed by the dorsal stream (cf. Fig. 1) provides the ``container'' for this context as well as the relative locations and motions of objects within it.  Contexts typically, however, also include ``how'' and ``why'' information, largely derived by processing pathways in parietal cortex \citep{Fields13}, that represent inferences about mechanical and intentional causation, respectively.  As in the case of object categorization and individual object-token encoding, these causal inference capabilities are present in rudimentary form in early infancy, and develop rapidly over the first two years \citep{Baillargeon12,Johnson15}.  Context assembly has been mapped to parahippocampal cortex, with object token to context binding implemented by the hippocampus \citep{Eichenbaum07,Fields13}.  \citet{Hommel} has termed the fully-bound representation of interacting objects in context an ``event file''; these representations mediate event understanding and context-dependent action planning.  Event files are the least complex visual representations that typically enter human awareness; hence they can be considered to be implemented by coherent activity at the level of the GNW (cf. Remark \ref{gnw-remark} ; for specific GNW-based models of awareness, see \citet{Baars3,Dehaene04,Baars13,Dehaene14}; see also \citet{Franklin1,Franklin12} for discussions of LIDA as a robotic architecture based on GNW design principles).

Event files correspond to ``episodes'' in episodic memory, again a hippocampus-centered function \citep{Eichenbaum07,Rugg13}.  As sequences of episodic memories typically contain many of the same ``players'' -- including in particular the self \citep{Renoult12} -- they pose a particular problem for object-token updating.  Each episodic memory must contain at least some episode-specific details -- e.g. what a particular person was wearing -- in addition to the ``core'' identifying information for each included object token.  Linking episodic memories into a temporal sequence requires maintaining this ``core'' -- which it is useful to consider as a ``singular category'' with just one member -- while modifying its ``essential'' identifying information for the represented object as needed, e.g. updating a person's age or personality characteristics (Fig. 3).  This maintenance process is effectively the construction of a history or model of the individual represented by the object token, the FCH discussed above \citep{Fields3}.  Episodic memory recall and reconsolidation can modify the properties associated with or even the presence of the object tokens referenced by the memory, demonstrating the fragility of such FCHs \citep{Schwabe14}.  Infants are capable of episodic recall over short periods -- e.g. the time periods required for experiments assaying causal inference -- but have limited recall and reconsolidation ability over longer periods \citep{Hayne04,Bauer06}.  Hence infants can be expected not to maintain robust object histories until about age four, the age when ``childhood amnesia'' typically ends.

\centerline{\includegraphics[width=16cm]{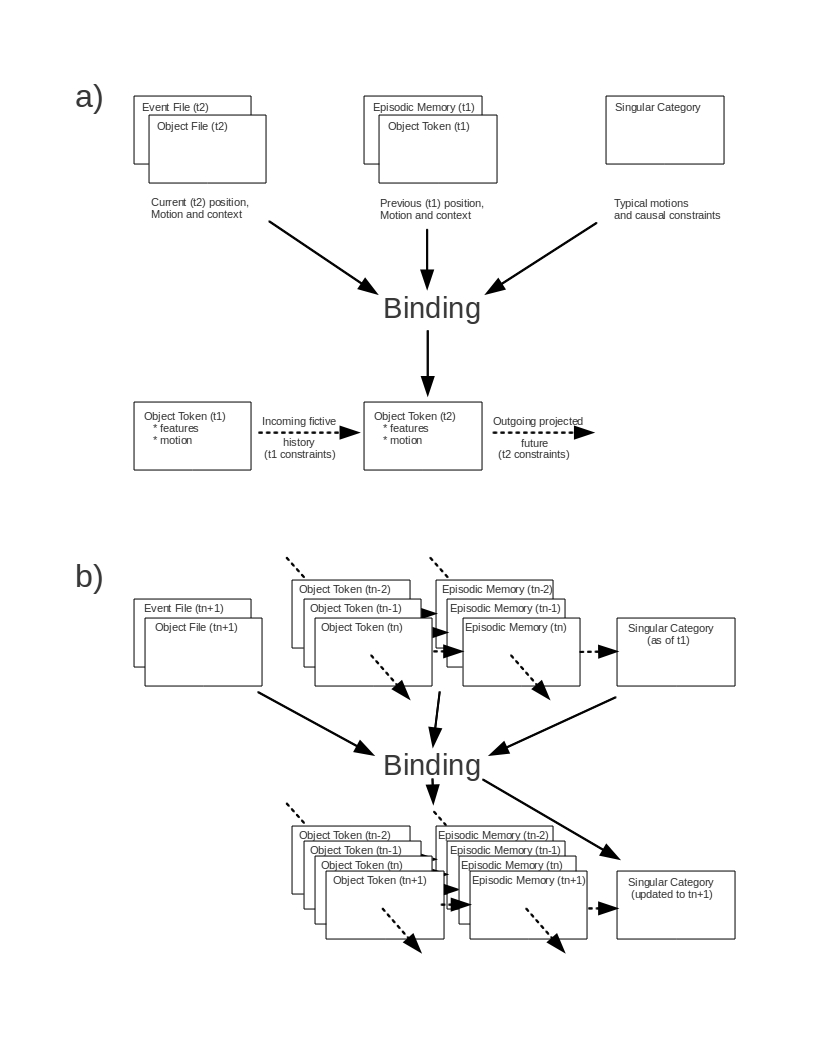}}
\begin{quote}
\textit{Fig. 3}:  a) Object token updating following a new event involving a recognized object.  The singular category specifying the object's ``core'' identifying criteria constrains the recognition and updating processes.  b) Updating the constraint information in a object-specifying singular category given a sequence of episodic memories and a new event.  Such updates must be infrequent compared to object token updates to maintain coherent identification criteria.  Adapted from \citet{Fields3}.
\end{quote}

Episodic memories can also reference object tokens for individuated and categorized but otherwise unidentified objects, e.g. ``some other people'' present at a meeting or ``other cars'' involved in an accident.  These ``other'' objects may appear in no other episodic memories and have no associated histories; they are represented by effectively one-off object tokens that are required by the data structure but play no other role in the system.  The human ability to learn to recognize new individuals indicates that such minimally filled-out object tokens are available for matching new incoming object files; however, their lifetime and the extent and context-dependence of their availability remain poorly understood.

\subsection{Attention, salience and Bayesian precision}\label{bayesian}

Systems with limited cognitive resources must allocate processing to the inputs most likely to be important.  In the current setting, this corresponds to paying attention to some objects and not others.  Attentional control in primates is implemented by competing, cross-modulating dorsal (top-down, goal-driven, proactive) and ventral (bottom-up, percept-driven, reactive) attention systems \citep{Goodale14,Vossel14}.  The ``salience network'' that controls these attention systems develops in concert with the medial-temporal object recognition network, starting from earliest infancy \citep{Gao15,Uddin15}.

Baysian predictive coding has long been employed as a model of perceptual processing from early vision through categorization and individual object identification \citep{Friston2,Maloney10}. \citet{Bastos12} review structural and functional evidence that predictive coding is implemented at the level of local microcircuits comprising cortical minicolumns, the dominant architectural units in mammalian cortex, as well as at the larger scales of functional networks responsible for trajectory recognition, categorization or object-token matching.  This use of the same or similar processing methods at different scales, as well as the overall hierarchical organization of perceptual processing \citep{VanEssen92}, suggests the kind of association between scale and coarse-graining introduced in \S\ref{simplicial}; we will return to this connection below.

In a Bayesian predictive coding system, attentional control can be modelled by varying the precision assigned to inputs and expectations.  In the Bayesian ``active inference'' framework of \citet{Friston2}, relatively high-precision inputs drive the revision of expectations and model reactive, ventral attention, while relatively high-precision expectations drive input-changing behavior and model proactive dorsal attention.  The framework also allows direct alterations of precision assignments as inferential outcomes.  The adaptive resonance (ART) framework of \citet{Grossberg13} provides a functionally similar model of attentional control, although its motivation and underlying ideas are distinct from those of \citet{Friston2} and its state-updating rules are not Bayesian.  When viewed as implementations of constraint hierarchies, however, both Bayesian active inference and ART exhibit the deep duality discussed in \S\ref{CCCD} below.  It is this duality, we will argue, that makes them useful models of attentional control.

%%%%%%%%%%%%%%%%%%%%%%%%%%%%%%%%%%%%%%%%%%%%%%%%%%%%%%%%%%%%%%%%%%%%%%%%%%%%%%%%%%%%%%%%%%%%%%%%%%%%%5

\section{Tokens, types and information flow in perception and categorization}\label{tt-flow}

While the human classification of perceived objects into cognitive categories corresponding to verbally-expressible concepts like ``person'' or ``house'' forms part of the motivation for the work of \citet{Dretske1,Barwise4,Barwise1} and others reviewed in Part I, category-theoretic methods have yet to be applied to the analysis of these processes at the level of detail reviewed in \S\ref{cognitive}.  This formal treatment of ontologies discussed in \S\ref{ontologies}, for example, does not explicitly address the question of how ontologies are constructed or maintained through time in the face of new observations.  We begin in this section to develop the constructs needed for more complete models.  We introduce the idea of a Cone-Cocone Diagram (CCCD) to capture the deep duality evident in the bidirectional flow of constraints between perception and categorization.

\subsection{Representing object files in a Chu space}\label{object-file}

The fundamental perceptual token, the initial representation of a discrete perceptual entity, is the object file.  As outlined in \S\ref{dual-process} above, an object file binds a collection of static features such as size, shape, texture and color extracted by ventral-stream processing to ``instantaneous'' (i.e. within $\Delta t$) location and trajectory information extracted by dorsal stream processing.  Let $F_1 \dots F_n$ be a finite tuple of static features, each of which can have any one of $m$ distinct values; e.g. if $F_i$ is `color' its distinct values are the colors distinguishable by the visual system of interest.  We can then consider a finite binary array $F = [f_{ij}]$, where $f_{ij} = 1$ for some object if and only if feature $F_i$ of that object has its $j^{th}$ possible value.  We can similarly consider finite binary arrays $X = [x_{ijk}]$ of discrete instantaneous three-dimensional locations and $V = [v_{ijk}]$ of discrete instantaneous three-dimensional velocities.  We will restrict attention to the case in which every object has some value for every perceptible feature, a single instantaneous location, and a single instantaneous velocity.  In this case, we can characterize an \emph{object file} as an instance of the finite array $[F,X,V]$.  These instances form a finite set $\{O_i \}$.

The most fundamental abstraction implemented by the visual system is \emph{object permanence}, i.e. the maintenance of object identity over time \citep{Fields17}.  At the level of the object file, the relevant timeframe for object permanence is a ``view'' lasting between half a second and a few seconds.  Objects that remain fully or even partially visible during a view are considered to remain ``the same thing'' while seen.  Whether an object that does not remain visible is perceived as remaining ``the same thing'' during a view depends on the age of the perceiver (less or more than 1 year) and the details of its occluded trajectory.  Objects moving sufficiently fast are ``seen'' as persistent even if their features, e.g. size, shape or color, vary over considerable ranges \citep{scholl08}.

Let $\{C_i \}$ be the finite set of finite (indeed short) sequences of object files that are treated by the cognitive system of interest as indicating object permanence during the course of a single view.  The elements of $\{C_i \}$ are then natural ``types'' relative to the ``tokens'' in the set $\{O_i \}$ of possible object files; an element $C_i \in \{C_i \}$ can be though of as ``associating'' a sequence of object files into a single abstracted representation.  Hence we can consider
\begin{equation}\label{chu-object}
\mathcal{C}_i = ( \{O_i \},\Vdash_{P},\{C_i \} )
\end{equation}
to be a Chu space, where here $\Vdash_{P}$ is the empirically-determined relation ``consistent with object permanence'' defined on sequences of object files.  This Chu space clearly describes a Classification in the sense of \citet{Barwise1} as stated in
\S\ref{classifications-1}.  Note that an element of $\{C_i \}$ may not be a concept in the sense of \S\ref{FCA-1}, as the value of every feature as well as the position and velocity can, at least in principle, change between every object token contributing to the perception of a persistent object.

The association of object files into an element of $\{C_i \}$ adds top-down, expectation-based information about identity over time to the ``raw'' information of perception.  This added information may, in fact, be incorrect; trajectories that appear to preserve object identity may involve distinct objects, while those that appear not to preserve object identity may involve a single object \citep{Fields11}.  More subtly, association into an element of $\{C_i \}$ also subtracts information by suppressing motion information relative to feature information.  While trajectory information dominates feature information in determining which sequences of object files to associate, persistent objects are required to have persistent features, at least during the course of short, single-view interactions \citep{Baillargeon08}.  Conferring persistence on an object converts its motion into a categorizable ``behavior'' that the object may or may not execute on other occasions.

\subsection{From object files to object tokens and object histories}\label{object-token}

Persistent objects are the ``entities'' in the common-sense ontology humans typically develop in late infancy.  These entities participate in episodic memories and are represented by object tokens and, if they recur sufficiently often to be recognized as persistent, by singular categories and (largely fictive) histories.  As with the abstraction of persistence, these successive levels of abstraction both add and subtract information.  Types at one level of abstraction, in particular, become tokens at the next.

As representations of persistent objects, \emph{object tokens} can be identified with elements of the set $\{C_i \}$.  Such tokens are, at their construction, already instances of multiple types (Fig. 4).  Persistent objects are instances, first, of the types representing their visually-identified features.  They are, second, classified automatically by threat detection, agency detection and animacy detection systems active beginning in early infancy \citep{Fields14}; the presence of a face alone indicates agency to human infants.  They are also classified, when possible, into entry-level and then more abstract cognitive categories, an ability also developed in infancy \citep{Rakison10}.  These token - type relationships can be represented as Classifications, as is standard in the literature (e.g. \citet{Barwise1}), and as we have surveyed here.  Recognition of an object by type generates a feeling of familiarity with the \emph{type}; e.g. seeing a cat generates a feeling of familiarity with cats.

\centerline{\includegraphics[width=16cm]{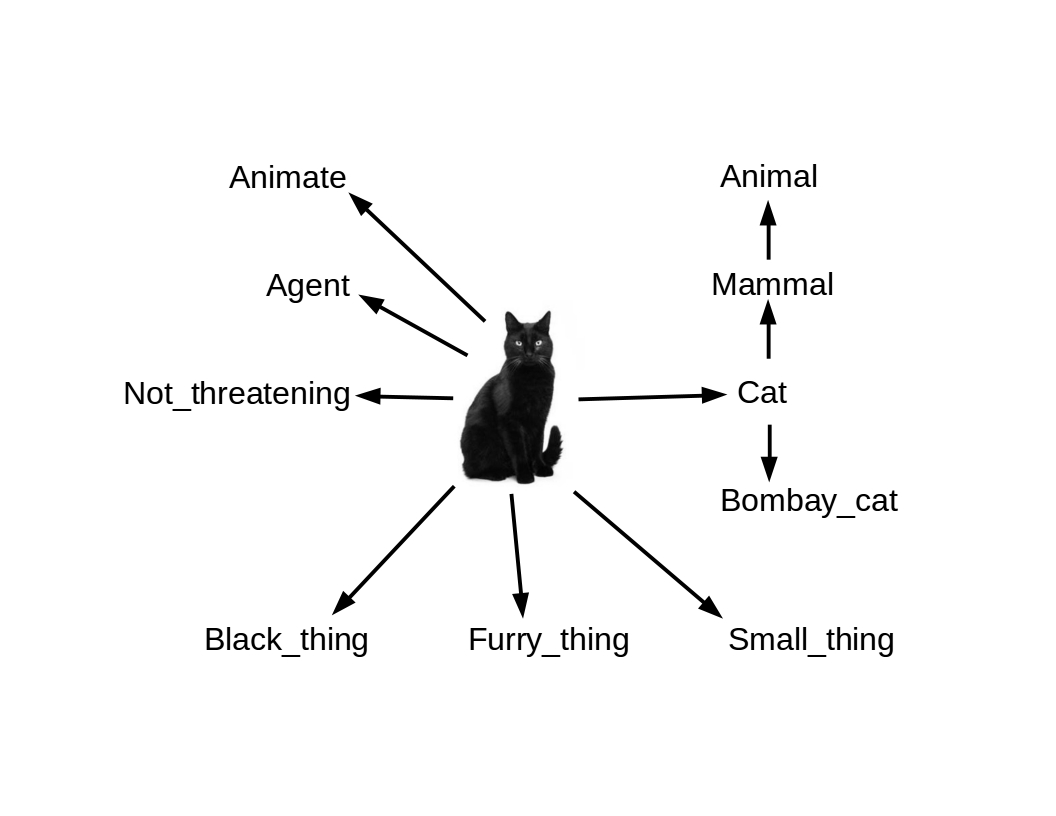}}
\begin{quote}
\textit{Fig. 4}:  An object token is classified at construction into multiple types by distinct but cross-modulating processes.  These include animacy and agency detection, emotion-mediated threat detection, and entry-level followed by superordinate and subordinate categorization into ``type'' of object.
\end{quote}

Here, however, we are primarily interested in the \emph{re-identification} of individual objects, i.e. the creation of an association indicating identity, and hence persistence over time, between an object token constructed now and one constructed previously.  At the object token level, the relevant timeframes for persistence range from the few seconds separating views to the decades separating a high-school graduation from a 50$^{th}$ reunion.

Let $C_i(t_1), C_j(t_2), ... C_k(t_n)$ be a sequence of $n$ object tokens encountered at successive times.  Recognizing successive object tokens as tokens of the very same individual thing involves at least the two processes discussed in \S\ref{episodic} above, i.e. matching to a set of core features composing a singular category and linking via FCH construction, with the uncertainty associated with both increasing with the time between perceptual encounters.
Let $D_l[t_1,t_n]$ comprise both the singular category and the FCH that together confer persistence on the object-token sequence $C_i(t_1), C_j(t_2), ... C_k(t_n)$, and let $\{D_l[t_1,t_n] \}$ be the set of all such representations over sequences of elements of $\{C_i \}$ indexed by observation times in the closed interval $[t_1, t_n]$.  The elements of $\{D_l[t_1,t_n] \}$ are, once again, natural ``types'' for the object tokens over which they are defined.  Hence we can consider a Chu space or Classification (in the sense of \S\ref{classifications-1})  $\A_i[t_1,t_n]$ as given by:
\begin{equation}\label{fch-chu}
\A_i[t_1,t_n] : = \langle \{C_i[t_1,t_n] \},\{D_i[t_1,t_n] \}, \Vdash_{P}[t_1,t_n]  \rangle
\end{equation}
where $\Vdash_{P}[t_1,t_n]$ is the empirically-determined relation ``consistent with object permanence'' defined on sequences of object tokens between $t_1$ and $t_n$.

The set of time-indexed representations $\{D_l \}[t_1,t_n]$ can be conceptualized more abstractly by noting that at each $t_j$, the set of possible object tokens $\{C_i(t_j) \}$ is also the set of types of a classification.  For each single time step $t_j \longrightarrow t_k$, the persistence criterion $\Vdash_{P}[t_j,t_k]$ induces maps -- what we have called FCHs -- between pairs of object tokens that can be consistently considered to be tokens of the same individual object (Fig. 5a).  These FCHs, together with the maps (here assumed to be identities) linking the singular categories for persistent objects, can be considered infomorphisms between the underlying classifications at $t_j$ and $t_k$.  It is then natural to interpret the set $\{D_l \}[t_j,t_k]$ as a channel between the underlying classifications; this channel comprises, intuitively, the (assumed constant) singular categories and the constructed FCHs (Fig. 5b).  Extending the process of linking object tokens by FCHs forward in time results in a hierarchy of channels, with the most temporally-extended channel as the colimit (Fig. 5c).  The colimit cocone $\{D_i\}[t_1, t_n]$ admits a vertex classification, which we denote $\mathbf{C}_i$ (with time interval $[t_1, t_n]$ understood).  Recall that this $\mathbf{C}_i$ is induced by a complex of infomorphisms:
\begin{equation}\label{info-complex}
 \cdots ~\lra \A_i[t_1, t_n] \lra \A_{i+1}[t_1,t_n] \lra ~  \cdots
\end{equation}
as depicted in \eqref{cocone-diagram}.  We will refer to such diagrams $\{D_i\}[t_1, t_n]$ as ``Cocone Diagrams'' or CCDs extending for a specified time interval, e.g. $t_1 ... t_n$ in Fig. 5c.

%\centerline{\includegraphics[width=16cm]{FCH-cocone.jpg}}
\bign
a)
$$
\xymatrix@!C=6pc{\rm{SC}_i(t_j) \ar[r]^{\rm{Id}}  & \rm{SC}_i(t_k) \\
\rm{OT}_i(t_j) \ar[u]^{\rm{Inst}}  & \ar[l]^{\rm{FCH}} \rm{OT}_i(t_k)  \ar[u]_{\rm{Inst}}}
$$

\bign
b)
$$
\xymatrix{& \{D_i\}[t_1, t_2] & \\
\{C_i\}(t_1) \ar[ur]  &  & \{C_i\}(t_2) \ar[ll]^{\rm{FCH}} \ar[ul]}
\xymatrix{& \{D_i\}[t_3, t_4] & ~ ~\\
\{C_i\}(t_3) \ar[ur]  &  & \{C_i\}(t_4) \ar[ll]^{\rm{FCH}} \ar[ul]}
$$

\bign
c)
$$
\xymatrix@!C=3pc{& & & \{D_i\}[t_1,t_n]  & & \\ & & \{D_i\}[t_1, t_4] \ar@{.>}[ur] & .... &   \{D_i\}[t_{n-3}, t_n]\ar@{.>}[ul] & & & &\\
& \{D_i\}[t_1,t_2]\ar@{.>}[ur] \ar@{.>}[r]  & & \{D_i\}[t_3,t_4]  \ar@{.>}[ul]  \ar@{.>}[r]   & & \{D_i\}[t_{n-1}, t_n] \ar@{.>}[ul]  & & & & &\\
\{C_i\}(t_1)\ar[ur] & \ar[l]^{\rm{FCH}} \{C_i\}(t_2)\ar[u]  ...  & \ar@{.>}[l]  \{C_i\}(t_3) \ar[ur] & \ar[l]^{\rm{FCH}} \{C_i\}(t_4) \ar[u]  & & \{C_i\}(t_{n-1}) \ar@{.>}[l]  \ar[u] & \ar[l]^{\rm{FCH}} \{C_i\}(t_n) \ar[ul] & & & &}
$$

$~$
\begin{quote}
\textit{Fig. 5}:  a) Interpreting sequential object tokens as representing the same persistent individual constructs an FCH to link them.  The FCH is depicted as acting backwards in time as it is built from the new observation to the old one.  Here $\rm{SC}_i(t_j)$ denotes Singular Category $i$ at $t_j$, ~$\rm{OT}_i(t_j)$ denotes Object Token $i$ at $t_j$, etc. $\rm{Id}$ = Identity, and $\rm{Inst}$ = Instance.  b) Families of FCHs link sets $\{C_i \}(t_j)$ of object tokens instantiated at different times.  The set $\{D_i \}[t_j,t_k]$ of abstracted singular category plus FCH pairs representing objects persistent from $t_j$ to $t_k$ can be viewed as a channel between sets of linked object tokens.  c) A CCD representing an object history during a time interval $t_1 ... t_n$.  The set $\{D_i \}[t_1,t_n]$ is a colimit cocone for sequences of object tokens consistent with persistence from $t_1$ to $t_n$.
\end{quote}

The ``essential'' identifying properties of objects can change over time, though they cannot all change together without causing identification failure.  In the case of human beings, for example, both (approximate) age and core personality characteristics are identifying properties; hence a child with an adult friend's personality is not identified as one's adult friend.  Slow, asynchronous changes in the composition of singular categories and hence small departures from identity of the linking maps between them do not alter the structures of the above diagrams.  Such changes do, however, render FCH construction more difficult.

\subsection{Contexts, event files and episodic memories}

Objects are never encountered in complete isolation; even the most austere psychophysics experiments have a computer screen and the surrounding laboratory as a context.  In real-life settings, objects are typically encountered in interacting groups.  Object tokens have, therefore, lateral synchronic associations as well as the diachronic links implemented by FCHs.  The event-file construct of \citet{Hommel} provides a ``snapshot'' of such associations over the few-second to few-minute timeframes intuitively regarded as single ``events.''  Event files capture interactions between objects as well as their significance and affective consequences for the observer.  These kinds of information provide crucial input into the FCH construction processes that allow the objects participating in the event to be identified \citep{Eichenbaum07,Zimmer,Fields3}.

Event files as defined by \citet{Hommel} are effectively tokens; each represents a discrete event that can be encoded and then retrieved as a discrete episodic memory.  Single events are by definition localized in time and hence cannot be repeated; recalling an event and hence (partially and perhaps inaccurately) reconstructing an event file, in particular, occurs in a current context and itself constitutes a distinct event.  The recognition of event \emph{types} as such is not well characterized experimentally.  It seems reasonable to expect, however, that events tokens (i.e. events files) and event types form classifications under the action of a satisfaction relation that maps tokens to types.  We consider this further in \S\ref{mereological} below in the broader context of mereological complexity and reasoning.

\subsection{Learning new categories and Cone-Cocone Diagrams}\label{CCCD}

With high frequency in infancy and childhood but typically reduced frequency thereafter, humans encounter not just individual objects, but object types that they have never encountered before.  Humans often learn to recognize such novelties from just one ``training'' encounter.  Understanding how humans achieve such one-shot learning is a major challenge for cognitive neuroscience, just as replicating this ability is a major challenge for machine learning.  Besides achieving efficient, preferably one-shot learning from exemplars, the problem has (at least) two additional components: recognizing novelty and switching from classification mode to learning mode.  As \citet{Oudeyer} have emphasized, it is \emph{learnable} novelty that must be recognized; otherwise precious resources are wasted on attempts to learn the unlearnable.

Consider a novel object that is easily classified as an instance of a familiar entry-level category: a novel cat, for example.  The object is recognized as novel because its object token does not match any existing singular category, cannot be linking to any existing object token by a plausible FCH, or both.  Interacting with the object over an extended period (several views, a few minutes) or encountering it again after a short delay allows certain of its properties to be identified as unchanging; in the case of a cat, these may include size, shape, color pattern, face and voice but not location or behavior.  The (short) sequence of distinct object tokens recorded during such interactions serves, in other words, to associate some properties of the object into a provisional singular category.  The principle of association here is, once again, persistence: each successive object token indicates the object as persistent, and the features encoded by object tokens in the sequence are similar enough to be treated as identical.

We can, in this case, consider the distinct feature instances encoded by the distinct object tokens in the sequence to be feature ``tokens'' and consider the object tokens themselves, which the criterion of persistence identifies as representing one individual object, as jointly defining a ``type'' that organizes those tokens.  The construction employed in \S\ref{object-file} above can then be employed to construct a classification of these tokens into these types.  This classification is the Chu-space dual of the classification of object tokens by singular categories shown in Fig. 5a.

The requirement of a familiar entry-level category can now be relaxed: suppose that what is encountered is not a novel cat, but an entirely novel animal, perhaps a pangolin or a platypus.  In this case an entirely category must be learned.  However novel the object encountered is, it must have \emph{some} familiar features, e.g. its approximate size and shape, whether it has a face, perhaps some aspect of its behavior.  Even very young infants can use features of these kinds to initiate classification and identify novelty \citep{Rakison10}.  Placement in any familiar category allows the construction of a singular category as outlined above.  Construction of a non-singular category -- e.g. [pangolin] -- merely requires abstraction, i.e. allowance of inexact matches.

The problem of maintaining a singular category across changes in essential features introduced in \S\ref{object-token} above can now be seen as a special case of category learning.  A singular category is robust against feature changes if the FCHs linking its instances are strong enough that persistence at the object token level can induce persistence at the singular category level.  The ``flow of association'' in this case is the reverse of that depicted in Fig. 5c; the properties composing the singular category are in this case the ``tokens'' that are held together by the persistent object history as a ``type.''

Reversing the arrows in a CCD (e.g. Fig. 5c) yields a cone, the dual of a cocone.  A system capable of both object history construction and its dual, category learning with singular category maintenance as a special case, is thus characterized by a \emph{cone-cocone diagram} (CCCD); such a diagram can be represented by making all of the arrows in a CCD such as Fig. 5c double-headed. Continuing the notation used in Fig. 5c, we denote the corresponding CCCD by $\mathbf{Dg}_i[t_1, t_n]$.

A CCCD captures the simultaneous upward and downward flow of constraints that characterize human vision and, it is reasonable to suppose, other sensory modalities both functionally and neuro-architecturally \citep{Hochstein}.  The duality expressed by a CCCD is the duality found between dorsal and ventral attention systems, or between high-precision expectations and high-precision inputs in an active inference system.  It resolves the central paradox of familiarity: that familiarity can confer either high or low salience in a context-dependent way.  The ``switch'' between these dual constraint flows appears to be implemented, in humans, by the amygdala - insula - cingulate axis at the core of the salience network (e.g. \citet{Uddin15}).

\subsection{Local logics embedded in CCCDs}\label{cccd-logic-1}

Recall that any classification generates a natural local logic in accordance with Definition \ref{local-2}, and that \citet[Prop. 12.7]{Barwise1} ensures that any local logic defined on a classification can be identified with the local logic generated by the classification (cf. Example \ref{building}).  These ideas can now be applied to the classifications defined above to characterize the categorization and identity maintenance processes in terms of the actions of local logics.

To begin, we can immediately apply the principle of Definition \ref{local-2} to \eqref{chu-object} relating ``instantaneous'' object files (tokens) to short sequences of object files (types) indicating object permanence, to obtain a local logic $\Lg(\mathcal{C}_i)$ with regular theory $\Th(\mathcal{C}_i) = (\{ C_i \}, \vdash)$.  This $\Th(\mathcal{C}_i) = (\{ C_i \}, \vdash)$ expresses the effective criteria for short-term object permanence and hence captures, albeit implicitly, an important part of the semantics of ``object'' for the system it describes.   Likewise, in \eqref{fch-chu}, we have a local logic $\Lg(\A_i[t_1, t_n])$ with regular theory $\Th(\A_i[t_1, t_n]) = (\{D_i\}[t_1, t_n], \vdash)$ (for each $i$) that captures the effective criteria for longer-term object permanence and hence additional components of the semantics of ``object.''  In both cases all tokens are normal (see Definition \ref{local-2}). On recalling Definition \ref{logic-info}, we can take as a working principle that sequences such as in \eqref{info-complex} are logic infomorphisms satisfying the properties of \citet[12.3]{Barwise1}. Accordingly, an underlying semantic structure is built into Fig. 5c, and hence to the ensuing CCCD diagram $\mathbf{Dg}_i[t_1, t_n]$.

Sequents serve purposes in this development.  On the one hand, they are implicitly assumed in the preceding discussion (see also \S\ref{local} and \S\ref{ontologies}).  On the other hand, we recall from \S\ref{shannon} that on relaxing the sequent relation $\vdash$ to a conditional probability (see \eqref{prob-2}), we see how a sequence of logic infomorphisms may function as a chain of Bayesian inferences.  This is consistent with the use of Bayesian methods reviewed in \S\ref{bayesian}, and suggests that the diagrams $\mathbf{Dg}_i[t_1,t_n]$ may be considered as effective carriers of Bayesian inference through sequences of episodic memories, and idea that is extended in \S\ref{mereological} below.

As regards ontologies, we recall that both types $\{C_i\}$ and $\{D_i\}[t_1, t_n]$ are not strictly sets of (formal) concept symbols. We suggest that a weaker sense of ontology is obtainable on assuming the relations $\leq, ~\perp, |$ in Definition \ref{ontology-1}. Generally, however, we can acknowledge the viewpoint of \citet{Kalfoglou2} which sees the local logics themselves as characterizing ontologies.  This derivative sense of ontology is useful for our purposes since ontological partitions into ``entities'' tend to induce ``spatial'' boundaries around conceptual and/or perceptual partitions \citet{Smith1996}.  Such induced boundaries can be identified with coarse-grainings and hence induced geometries, as will be discussed in \S\ref{mereotop} below.

%%%%%%%%%%%%%%%%%%%%%%%%%%%%%%%%%%%%%%%%%%%%%%%%%%%%%%%%%%%%%%%%%%%%%%%%%%%%%%%%%%%%%%%%%%%%%%%%%%%%%5

\section{Parts and wholes: Using Chu spaces and information channels to represent mereological complexity}\label{mereological}

The Formal Ontology introduced by \citet{Husserl} developed a theory of ``parts'' and ``wholes'' towards a foundation for mereological reasoning, a methodology that also has roots in the works of Aristotle, Brentano, Whitehead,  and others (as reviewed and developed in \citet{Casati1,Lando,Les,Simons,Smith1996}).  Formalizations of mereological reasoning (e.g. \citet{Casati1,Smith1996}) have found wide application in geographic information systems (GIS) and formal ontologies.  The implementation of mereological reasoning in humans is not, however, well understood; indeed we have been able to find only a single neuroimaging study explicitly comparing mereological and functional classifications \citep{muehlhaus14}.  As mereological reasoning appears specifically to fail in the ``weak central coherence'' phenotype of autism spectrum conditions \citep{Happe06}, understanding its implementation is potentially of clinical relevance.

\subsection{Perceptual identification of mereologically-complex objects}\label{mcomplex-1}

The macroscopic objects perceptible by humans are by definition mereologically complex: they have multiple perceptible parts, each of which has further parts, etc.  Such objects can, moreover, be assembled into larger complexes, with perceptual scenes being ubiquitous, transient examples.  Many such larger complexes are, however, not transient but rather meaningful, persistent objects in their own right.  A fundamental challenge posed by human object perception is to understand what mereological complexes are perceived as ``whole'' objects, how object tokens representing such complexes are constructed, and how such object tokens are linked into persistent histories despite changes in the properties and even identities of the ``parts'' making up the complex.

A specific example of a mereological hierarchy is shown in Fig. 6.  Individual human beings, such as author CF, are entry-level (EL) objects and hence are represented by EL object tokens.  Human beings are inevitably members of larger complexes, including families, extended families, tribes, ethnic groups, nations, etc.  The smaller instances of such complexes (e.g. human nuclear families) can be directly perceived; larger instances may not be perceptible but can be referred to using language, images, and abstract graphics.  Hence object tokens can be constructed for such complexes.  Object tokens representing ``parts'' such as CF are naturally linked to object tokens representing complexes, such as CF's family, by ``$\mathrm{part\_of}$'' relations.  Such relations similarly link parts of CF to CF.  Entry-level objects appear to play a special role in such hierarchies; ``$\mathrm{part\_of}$'' links are transitive both above and below EL objects, but not across EL objects.  A part of CF is not a part of CF's family, just as a part of a car is not a part of a fleet of cars.

\centerline{\includegraphics[width=20cm]{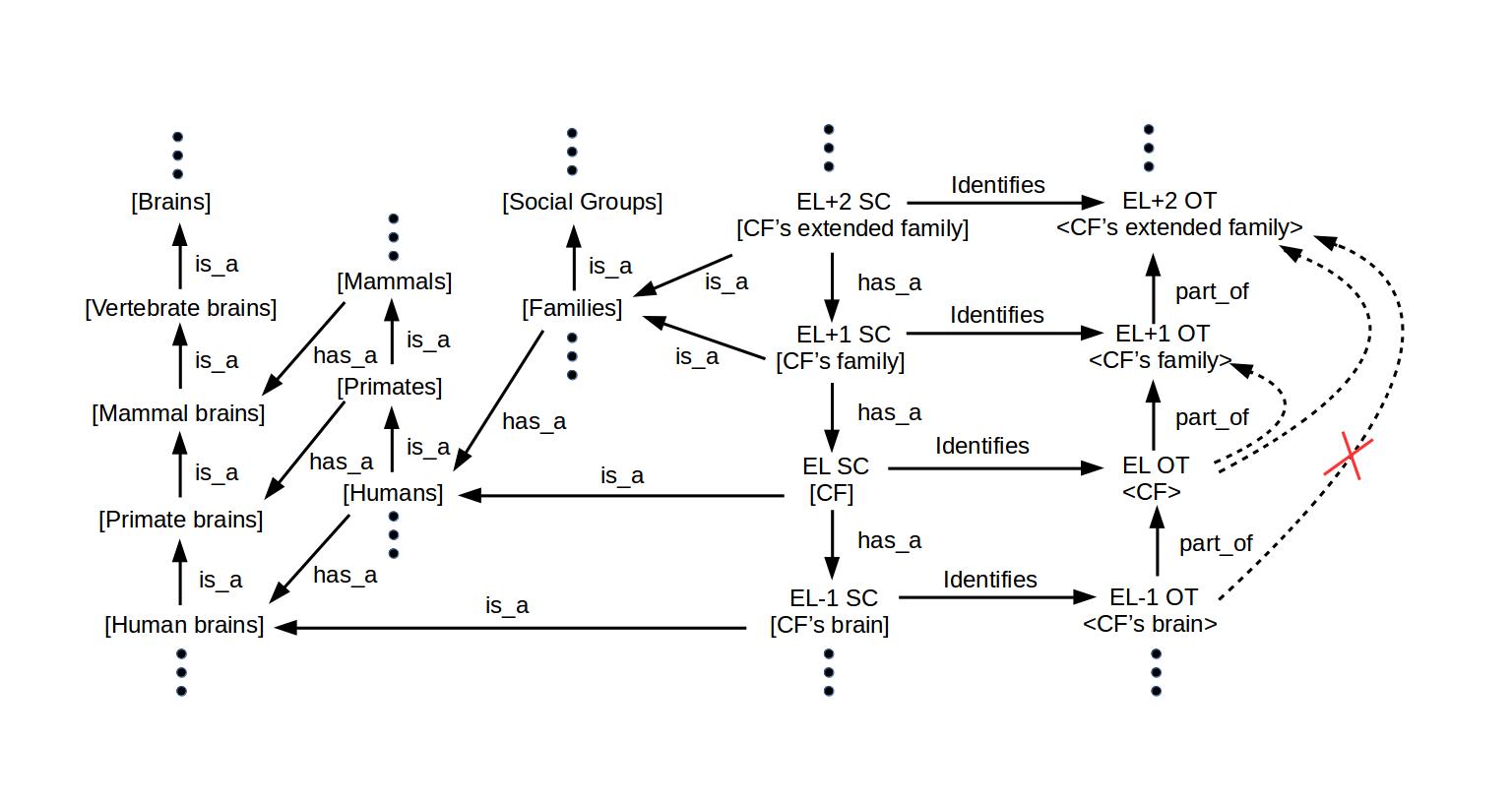}}
\begin{quote}
\textit{Fig. 6}: Example of an object token (OT) hierarchy extending both above and below a mereologically-complex entry-level (EL) object, one of the present authors (CF).  Each OT has an associated singular category (SC) specifying identifying static and behavioral features.  These SCs are in turn associated with general categories, some of which are shown here.  Solid arrows show typical ``$\mathrm{part\_of}$'', ``$\mathrm{has\_a}$'' and ``$\mathrm{is\_a}$'' links.  Dashed arrows show induced $\mathrm{part\_of}$ links; red ``X'' indicated the failure of ``$\mathrm{part\_of}$'' transitivity across the EL OT.
\end{quote}

Hierarchies of tokens linked by ``$\mathrm{part\_of}$'' relations are commonplace in AI systems.  The existence of such ``$\mathrm{part\_of}$'' hierarchies raises, however, the questions of what the ``$\mathrm{part\_of}$'' relation is, how it is established, and how it is maintained over time.  The correspondence between object tokens and singular categories provides a partial answer: ``$\mathrm{part\_of}$'' relations between object tokens correspond to ``$\mathrm{has\_a}$'' relations between singular categories, which in turn correspond to ``$\mathrm{has\_a}$'' relations between general categories (Fig. 6).  While the object token is the locus of learning for the first exemplars of EL objects encountered in infancy and childhood, once a general category has been learned the ``$\mathrm{part\_of}$'' links between new object tokens can be induced by inter-category ``$\mathrm{has\_a}$'' links.  The mechanisms by which such link induction is implemented remain to be elucidated; we consider formal structures supporting this process below.

\subsection{Mereological hierarchies as hierarchies of CCCDs}\label{mcomplex-2}

Let us first consider the Chu space $\sfC = (C_{\sfo}, \Vdash_{\sfC}, C_{\sfa})$, where $C_{\sfo}$ and $C_{\sfa}$ are sets of object tokens and their corresponding singular categories and $\Vdash_{\sfC}$ is the ``Identifies'' relation in Fig. 6.  Recall from \S\ref{FCA-1} the pair of maps $(\alpha,\omega)$ (there considered as a Galois connection) given by:
\begin{equation}\label{mereo-1}
\begin{aligned}
\alpha &: \cP(C_{\sfo}) \lra \cP(C_{\sfa}) ~\text{with}~ \a(x) = \{\sfa: \forall x \in X, ~x \Vdash_{\sfC} \sfa \} \\
\omega &: \cP(C_{\sfa}) \lra \cP(C_{\sfo}) ~\text{with}~ \omega(A) = \{x: \forall \sfa \in A,~ x \Vdash_{\sfC} \sfa \}.
\end{aligned}
\end{equation}
Here $\alpha$ clearly maps an object token to its (unique) singular category and $\omega$ maps a singular category to its (unique) object token.

\begin{definition}\label{mereo-2}
$~$
\begin{itemize}

\item[(1)] Suppose $X$ is a set of subsets consisting of \emph{parts of objects}. Then we define $\omega \circ \alpha (X)$ to be the set of subsets of \emph{whole parts of objects as obtained from $X$}.

\item[(2)] Suppose $Y$ is a set of subsets consisting of \emph{parts of attributes}. Then we define $\alpha \circ \omega(Y)$ to be the set of subsets of \emph{whole parts of attributes as obtained from $Y$}.
\end{itemize}
\end{definition}
\noindent
The usage ``whole parts'' is employed here to emphasize that ``wholes'' on one level may be ``parts'' at the level(s) above.

Likewise, for a given classification we have
$\A = \langle \rm{Tok}(\A), \rm{Typ}(\A), \Vdash_{\A} \rangle$ , and for
$a \in X \subseteq \rm{Typ}(\A), b \in A \subseteq \rm{Tok}(\A)$, we have:
\begin{equation}\label{mereo-3}
\begin{aligned}
\alpha^* &: \cP(\rm{Typ}(\A)) \lra \cP(\rm{Tok}(\A)) ~\text{with}~ \alpha^*(x) = \{b: \forall a \in X, ~x \Vdash_{\A} b \} \\
\omega^* &: \cP(\rm{Tok}(\A)) \lra \cP(\rm{Typ}(\A)) ~\text{with}~ \omega^*(A) = \{a: \forall b \in A,~ a \Vdash_{\A} b \}.
\end{aligned}
\end{equation}
Iterating these conditions allows us to move one rung at a time through the mereological hierarchy when incorporating information channels.

To see how this mereological hierarchy can be constructed, we first of all construct a (quasi-hierarchial) complex of CCCDs following
\S\ref{object-token}, \S\ref{CCCD} and \S\ref{cccd-logic-1}. Recalling \eqref{fch-chu}, we may view the index $i$ as indicating a `level' in a complex of CCCDs constructed from connected sequences. In such sequences, the time intervals will generally be distinct. Thus we commence with families of time dependent logic infomorphisms arising from such morphisms between different classifications as specified in \eqref{fch-chu}:
\begin{equation}\label{cccd-1}
\A_i[t_{i_1},t_{i_n}] \lra \B_j[t_{j_1},t_{j_n}]
\end{equation}
at `levels' $i,j$ (possibly $j=i$), each respecting the `parts' to `wholes' condition of \eqref{mereo-3}.
Both classifications lead to their corresponding diagrams as in Fig. 5c, as explained in \S\ref{object-token}, with an induced (logic) infomorphism
$\mathbf{C}_i \lra \mathbf{C}_j$ between the cocone vertex classifications of the corresponding CCCDs derived from Fig. 5c (again, the time intervals are understood). Following the formulism of \S\ref{CCCD}, we thus obtain induced (logic) infomorphisms:
\begin{equation}\label{cccd-2}
\begin{aligned}
\mathbf{Dg}_i[t_{i_1}, t_{i_n}] &\lra \mathbf{Dg}_j[t_{j_1}, t_{j_n}] \\
\mathbf{C}_i &\mapsto \mathbf{C}_j
\end{aligned}
\end{equation}
Schematically, this leads to a typical quasi-hierarchial configuration as depicted in Fig. 7 below. Note that the assumption of taking logic infomorphisms provides an underlying semantic structure to the various mechanisms as discussed in \S\ref{cognitive} and \S\ref{tt-flow}.

$$
\xymatrix@C=4pc{... & ... & ... & ... & & \\& \mathbf{Dg}_j[t_{j_1}, t_{j_n}]\ar@{.>}[d]  \ar@{.>}[dl]   & &  \ar@{.>}[d] & \\
 \mathbf{Dg}_w[t_{w_1}, t_{w_n}] \ar@{.>}[u]  & \mathbf{Dg}_k[t_{k_1}, t_{k_n}]  \ar@{.>}[dl] \ar@{.>}[d] &  \mathbf{Dg}_{\ell}[t_{\ell_1}, t_{\ell_n}] \ar@{.>}[ul] \ar@{.>}[u]  & \mathbf{Dg}_p[t_{p_1}, t_{p_n}] \ar@{.>}[l]\ar@{.>}[d] \ar@{.>}[r] & \\
\mathbf{Dg}_v[t_{v_1}, t_{v_n}]\ar@{.>}[u] & & \mathbf{Dg}_m[t_{m_1}, t_{m_n}]\ar@{.>}[u] \ar@{.>}[dl] \ar@{.>}[ul] & \mathbf{Dg}_q[t_{q_1}, t_{q_n}]  \ar@{.>}[dl] \ar@{.>}[l] \ar@{.>}[d] \ar@{.>}[r] &  &\\
 \ar@{.>}[u]& \ar@{.>}[ul] \ar@{.>}[dl]  \mathbf{Dg}_s[t_{s_1}, t_{s_n}]    &  \ar@{.>}[l] \mathbf{Dg}_u[t_{u_1}, t_{u_n}]\ar@{.>}[dl]\ar@{.>}[u] & \mathbf{Dg}_r[t_{r_1}, t_{r_n}] \ar@{.>}[dl] \ar@{.>}[l] \ar@{.>}[d] \ar@{.>}[r] &   \\
... & ... & ... & ...& &
}
$$
$~$
\begin{quote}
\textit{Fig. 7}:  A typical complex of interactive CCCDs corresponding to a mereological object-token hierarchy that is maintained over time.
\end{quote}

Observe that the configurations depicted in Fig. 7 are not strictly hierarchial, even though the corresponding colimits are iterated.  Why it is not strictly hierarchial is plain to see: a part can be a part of many wholes, and even of wholes at different levels (for instance, an employee can be part of a division, but also part of a company). As will be seen later, it is in this respect that a mereotopological complex differs from a standard category-theory hierarchy in the sense of e.g. \citet{Baas2004,EV2007}. Here, in particular, because of the existence of co-planar complexes and bi-directional arrows (Fig. 8), it is not always the case that that a relevant object of level $n+1$, say, is the colimit of at least one diagram at level $n$. However, somewhat in line with \citet{Baas2004}, we may also interpret an ``observer'' as an entity that is external or internal to the system, or the system's environment itself, through which ``selection'' induces the further levels of structure.

\centerline{\includegraphics[width=20cm]{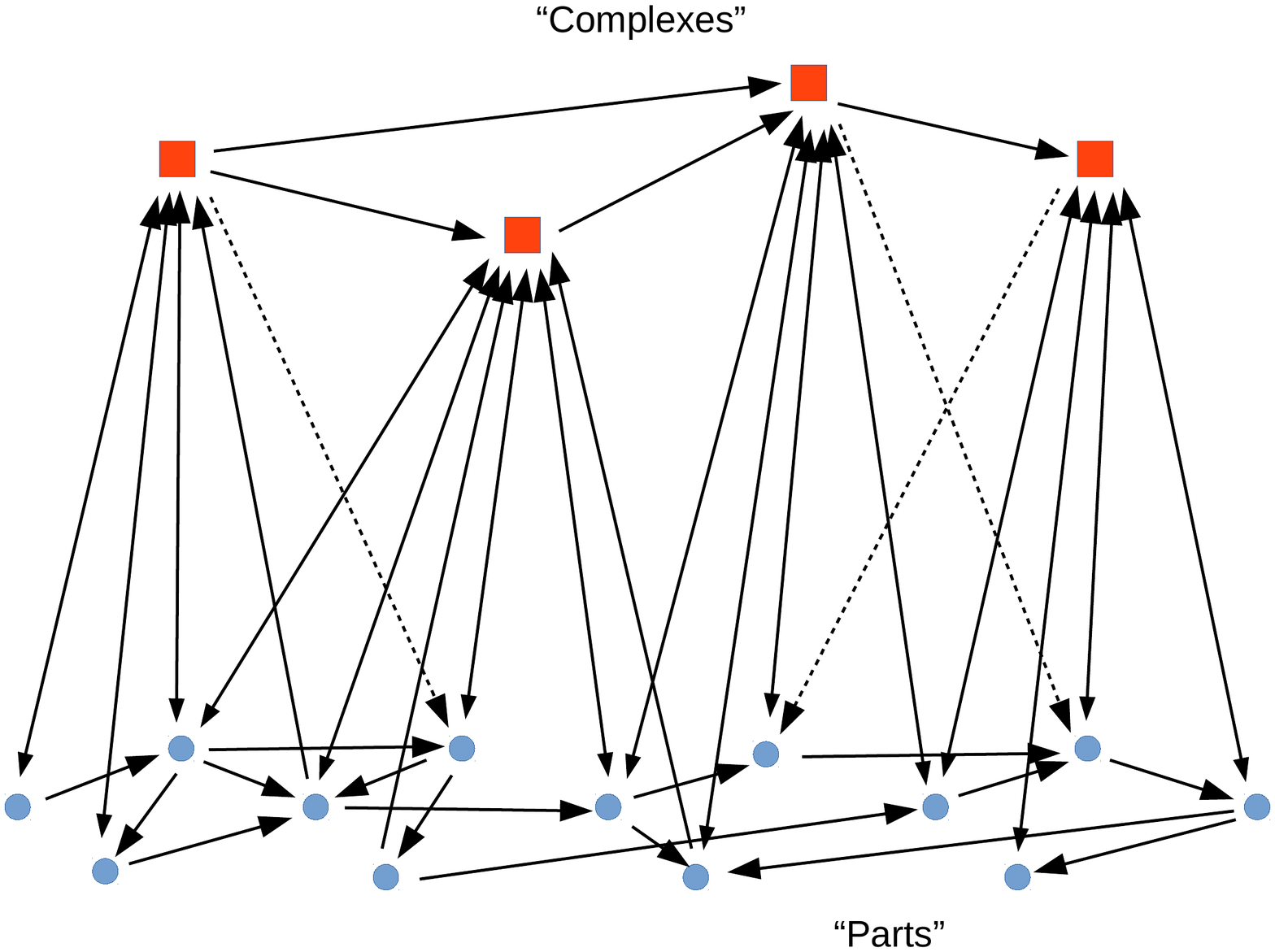}}
\begin{quote}
\textit{Fig. 8}: A two-layer mereological network with ``parts'' at the lower level, and ``complexes'' at the upper level. Each level comprises both tokens and types. For example, the ``complex'' level could consist of EL types such as [cat], [dog], [chair], etc. Each of these has many tokens that are specific individuals. The ``part'' level can include visually-identifiable, but non-essential features of these types, such as ``has four legs'', ``has fur'', ``walks with a gait $X$'', ``is white with brown spots'', etc. Here again there are tokens: e.g. four particular legs, a particular pattern of white-with-brown spots, etc.
\end{quote}

Lateral connections at each level indicate, e.g. the probable co-occurrence in a scene. Vertical arrows indicate mereological inclusions going upwards and top-down prediction from  current ``understanding'' or ``prior probabilities'' going downwards. These do not necessarily select the same relations, as they often do not in real-life situations.

Cocones exist in both directions in this mereological structure: linked (time dependent) colimit cocones as underlying a typical CCCD, and CCCDs as linked by a network of logic infomorphims $ ... \lra \mathbf{C}_i  \lra  \mathbf{C}_{i+1} \lra \mathbf{C}_{i+2} \lra ... $ between the cocone vertex classifications.

``Learning'' in this system would consist of: 1) associating altogether new lower-and upper-level types into the network; 2) distinguishing new high-level individuals as clusters of low-level tokens; 3) Convergence toward better predictions (i.e. all vertical arrows are bidirectional).  ``Building'' in this system is associating a bunch of upward arrows with an upper-level token. ``Abstraction'' would be the grouping of upper-level tokens into an upper-level type while preserving all arrows.  Previous work demonstrating general models of ANNs (\S\ref{mlp}) shows that these processes are allowed in principle; different specific choices of algorithms for these processes would be expected to produce different hierarchical structures.

\subsection{From mereology to mereotopology: Distinguishing objects by boundaries}\label{mereotop}

The detection of edges and their extension into contours that segment an image into bounded, non-overlapping regions is one of the earliest stages of visual processing (reviewed by \citet{wagemans1}).  What, however, distinguishes a two- or even three-dimensional array of bounded, non-overlapping pattern elements -- e.g. an array of color or texture patches -- from an array of bounded, non-overlapping objects?  As discussed above, animacy, agency and independent manipulability are important indicators of bounded objecthood during infancy and early childhood when object categories are first being learned and populated with exemplars.  What, however, are the inferences that enforce boundedness for objects, and how does the constraint of having a boundary affect the informational relations outlined above?

The key idea of mereotopology is that the parts of an object must be \emph{inside} the object, i.e. contained within its boundary \citep{Casati1,Smith1996}.  This constraint is, clearly, more easily satisfied for boundaries that are (at least approximately) smooth and convex.  As simplicity and hence resource efficiency appear to be general principles of perceptual system organization \citep{wagemans2}, one can expect perceivers to ``see'' smooth, convex boundaries -- e.g. convex hulls of geometrically more complex objects -- more easily.  Imposing smoothness and convexity -- e.g. by constructing the convex hull of a geometrically more complex object -- is a form of coarse-graining.  We can, therefore, suggest that constructing an ``exterior'' boundary around a collection of parts that then serves as a boundary for the whole simply is a coarse-graining operation.  In this case, the simplicial methods introduced in \S\ref{simplicial} immediately become relevant, and indeed provide a general method of constructing object boundaries from the bottom up in any mereological hierarchy representable in the CCCD form as in Fig. 7.

As noted earlier, a scene is a mereological complex; segmenting a scene by adding boundaries makes it a mereotopological complex.  At the ``top'' of the mereotopological hierarchy, a whole scene can be considered a multilayer complex of simplicial complexes (i.e. identified ``whole'' objects) of simplices (identifiable ``part'' objects).  Recall from \S\ref{simpl-def} that any such complex, at any level of the hierarchy, has associated barycentric coordinates and a natural metric.  Distances \emph{within} a simplicial complex at level $n$ of the hierarchy, however, can also be viewed as distances \emph{between} its component simplices at level $n-1$.  Boundaries, therefore, induce geometric relations between the bounded objects.  In this sense, spatial relations can be viewed as ``emergent'' from the distinctions between objects, a view with striking similarities to recent proposals within fundamental physics (cf. \citet{Fields17b}).

\subsection{Channels, inter-object boundaries and interactions}

The spatial separation between objects induced by their boundaries -- and hence by their distinguishability -- generates a time-dependent exchange of information and hence an \emph{interaction} as this term is traditionally understood.  Again working from the top down in a mereotopological hierarchy of simplicial complexes indentified with local classifications, every channel between classifications at level $n$ can also be viewed as a channel between the corresponding ``objects'', i.e. simplicial complexes.  This channel corresponds to the boundary between the ``objects'' if they are adjacent, i.e. if the corresponding simplicial complexes share $(n-1)$-level faces.  It is natural to think of the information transmitted along the channel as ``encoded on'' this boundary, i.e. as encoded holographically as this term is used in physics \citet{Fields17b}.  If the objects are not adjacent,  i.e. if the connecting channel is a composition at level $n$, the channel can be thought of as passing through a shared ``environment'' interposed between the objects.  The components of the composed channel cannot in general be expected to be isomorphisms; hence the structure of this interposed environment affects the interaction between the objects.

Perceiving object motion requires tracking the identity of the ``moving'' object through time; hence it involves a temporal sequence of mereological hierarchies along the lines of Fig. 7.  The structure of the top-level scene is different at each time increment; hence the metric relations between component simplicial complexes is time-dependent.  Changing the relative positions of objects re-shapes their shared environment, in general altering the interaction between them.  The distance and material, e.g. transparency or electrical permittivity, dependence of physical interactions can, therefore, be viewed as qualitatively ``emergent'' from the simplicial structure of classifications.

\section{Conclusion}

Category theory provides a rich set of tools for investigating relationships between structures, and hence for representing information flow between structures.  These methods have been applied widely in computer science, and are seeing increasing applications in physics.  As we have reviewed in Part I above, applications of category theoretic methods in the cognitive sciences -- mainly in the investigation of ontologies and ontology convergence -- have mainly been carried out at a high level of abstraction.  In Part II, we have begun the process of characterizing object perception in category-theoretic terms, particularly in terms of Chu spaces, classifications, simplicial complexes, and local logics.  These formal methods provide a natural and intuitively-clear representation perception as a multi-stage process in which types at one level may serve as tokens at a higher level.  The co-cone -- cone duality captured in CCCDs is particularly useful as a representation of the bidirectional information flow employed by the Bayesian predictive coding systems that the brain appears to implement at multiple scales.  The analysis we present in \S\ref{tt-flow} extends previous work on the representation of ANNs by making this duality explicit.  It also makes explicit the essential role of inferences -- here captured as infomorphisms -- in tracking object identity through time.

Human beings, and presumably other animals with relative complex cognitive systems, employ both abstraction and mereological hierarchies to categorize objects.  We show in \S\ref{mereological} that networks of time-indexed CCCDs provide a natural representation of the mutually-constraining relationship between these two categorization methods, particularly as they are employed in object-identity tracking.  We then explore briefly the emergence of spatial relationships and interactions between objects from their description as simplicial complexes embedded in the larger simplicial complex that constitutes a perceptual scene.  This top-down, emergence-based approach to mereotopology differs significantly from previous approaches that are geared to different priorities, and different applications \citep{Le,Smith1996}.

This initial foray into the categorical representation of cognitive processes raises a number of questions and illuminates several open problems.  One of the deepest is whether the satisfaction relations $\Vdash$ operating between tokens and types at any of the processing levels considered here are well-defined.  While it must be assumed that they are to develop models, and associations, it remains possible that ``$\Vdash$'' is token, type, context, or time dependent, as studies of the dependence of language on unspecified ``background knowledge'' \citep{Searle83} or of cognition generally on ``embodiment'' \citep{Anderson03,Chemero13} suggest.  Problems that require further work include the implications of the present results for the representation of ontologies and particularly for ontology convergence between agents that have encountered non- or only partially-overlapping collections of individual objects, the extent to which local logics define or constrain semantic relations between either tokens or types, the extent to which cognitively-significant differences in spatial scale can be captured by coarse-graining, and the question of why the geometry emergent from human visual perception should be three-dimensional.  There also remains the open question of how cognition is affected when one or the other of the categorization systems breaks down, as appears to be the case with the mereological system in autism where a sense of context, and an overall gestalt of certain situations may be adversely impacted \citep{Happe06}.  It is, finally, possible that the conceptual ground which we have covered could be further supplemented by certain related techniques of higher dimensional algebra \citep{Brown2011} and those of $n$-categories \citep{Leinster2004}, topics which remain for further investigation.

%%%%%%%%%%%%%%%%%%%%%%%%%%%%%%%%%%%%%%%%%%%%%%%%%%%%%%%%%%%%%%%%%%%%%%%%%%%%%%%%%%%%%%%%%%%%%%%%%%%%

\end{document}